\documentclass[10pt,twocolumn,letterpaper]{article}

\usepackage{iccv}
\usepackage{times}
\usepackage{graphicx}
\usepackage{epstopdf}
\usepackage{amsmath}
\usepackage{amssymb}
\usepackage{adjustbox}
\usepackage{caption}
\usepackage{subcaption}
\usepackage{color,soul}
\usepackage{tabularx}
\usepackage{multirow,bigdelim}
\usepackage{booktabs}
\usepackage{algorithmicx}
\usepackage{algorithm,setspace}
\usepackage{algpseudocode}
\usepackage{bbm}
\usepackage{xcolor}
\usepackage{colortbl}
\usepackage{array}

\usepackage[pagebackref=true,breaklinks=true,letterpaper=true,colorlinks,bookmarks=false]{hyperref}

\newcommand{\ignore}[1]{}

\DeclareMathOperator*{\argmax}{arg\,max}
\DeclareMathOperator*{\argmin}{arg\,min}
\let\emptyset\varnothing

\iccvfinalcopy 

\def\iccvPaperID{2178} 

\ificcvfinal\fi
\begin{document}

\title{Towards the Success Rate of One:\\ Real-time Unconstrained Salient Object Detection}
\author{
Mahyar Najibi$^1$, Fan Yang$^2$, Qiaosong Wang$^2$, and Robinson Piramuthu$^2$ \\
$^1$University of Maryland College Park \hspace{8mm} $^2$eBay Inc. \\
\tt\small najibi@cs.umd.edu, \{fyang4, qiaowang, rpiramuthu\}@ebay.com
}


\maketitle

\thispagestyle{empty}
\begin{abstract}
In this work, we propose an efficient and effective approach for unconstrained salient object detection in images using deep convolutional neural networks. Instead of generating thousands of candidate bounding boxes and refining them, our network directly learns to generate the saliency map containing the exact number of salient objects. During training, we convert the ground-truth rectangular boxes to Gaussian distributions that better capture the ROI regarding individual salient objects. During inference, the network predicts Gaussian distributions centered at salient objects with an appropriate covariance, from which bounding boxes are easily inferred. Notably, our network performs saliency map prediction without pixel-level annotations, salient object detection without object proposals, and salient object subitizing simultaneously, all in a single pass within a unified framework. Extensive experiments show that our approach outperforms existing methods on various datasets by a large margin, and achieves more than 100 fps with VGG16 network on a single GPU during inference.
\end{abstract}

\vspace{-2mm}
\section{Introduction}
Saliency detection is the problem of finding the most distinct regions from a visual scene. It attracts a great amount of attention due to its importance in object detection~\cite{ren2014region}, image segmentation~\cite{donoser2009saliency}, image thumb-nailing~\cite{marchesotti2009framework}, video summarization~\cite{simakov2008summarizing}, \etc.
\begin{figure}[t]
\centering
\begin{minipage}{0.23\linewidth}
\begin{tabular}{c}
\includegraphics[clip,trim=5cm 3cm 5cm 2cm, width=\linewidth]{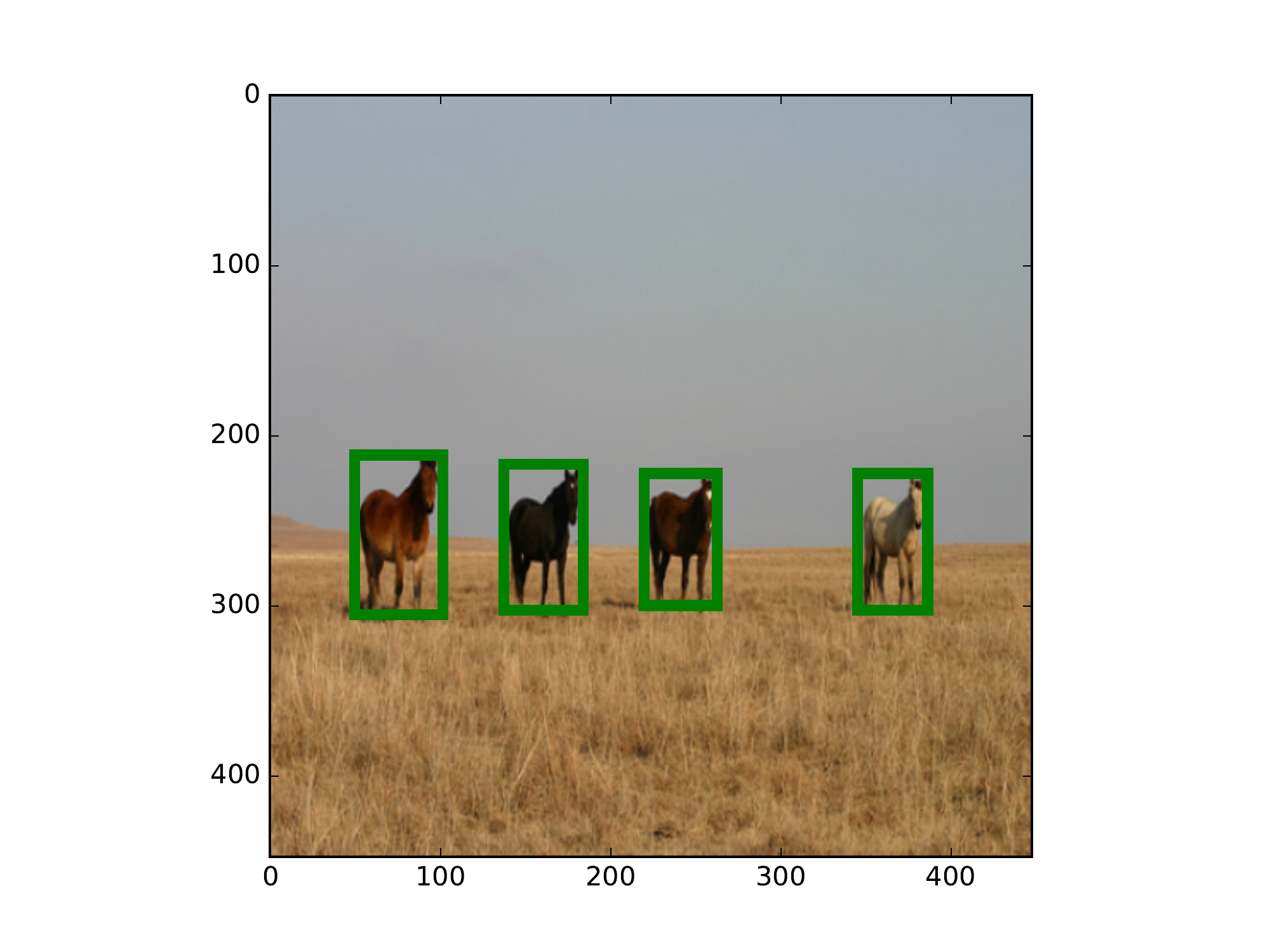}
\\
{\scriptsize (a)}
\end{tabular}
\end{minipage}
\hspace{3mm}
\begin{minipage}{0.5\linewidth}
\begin{tabular}{c@{}c}
\includegraphics[clip,trim=5cm 3cm 5cm 2cm,width=.45\linewidth]{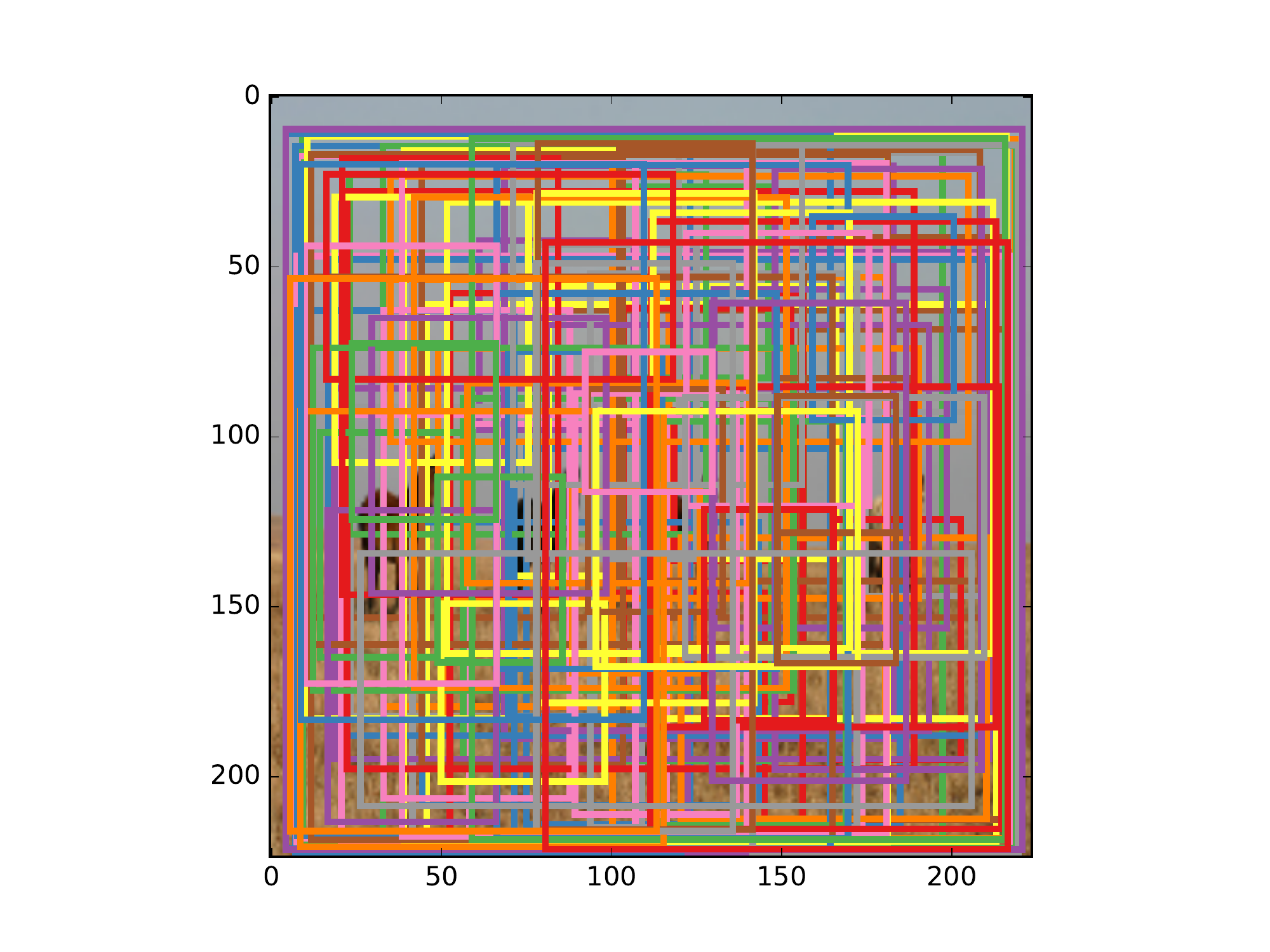}
\ &
\includegraphics[clip,trim=5cm 3cm 5cm 2cm,width=.45\linewidth]{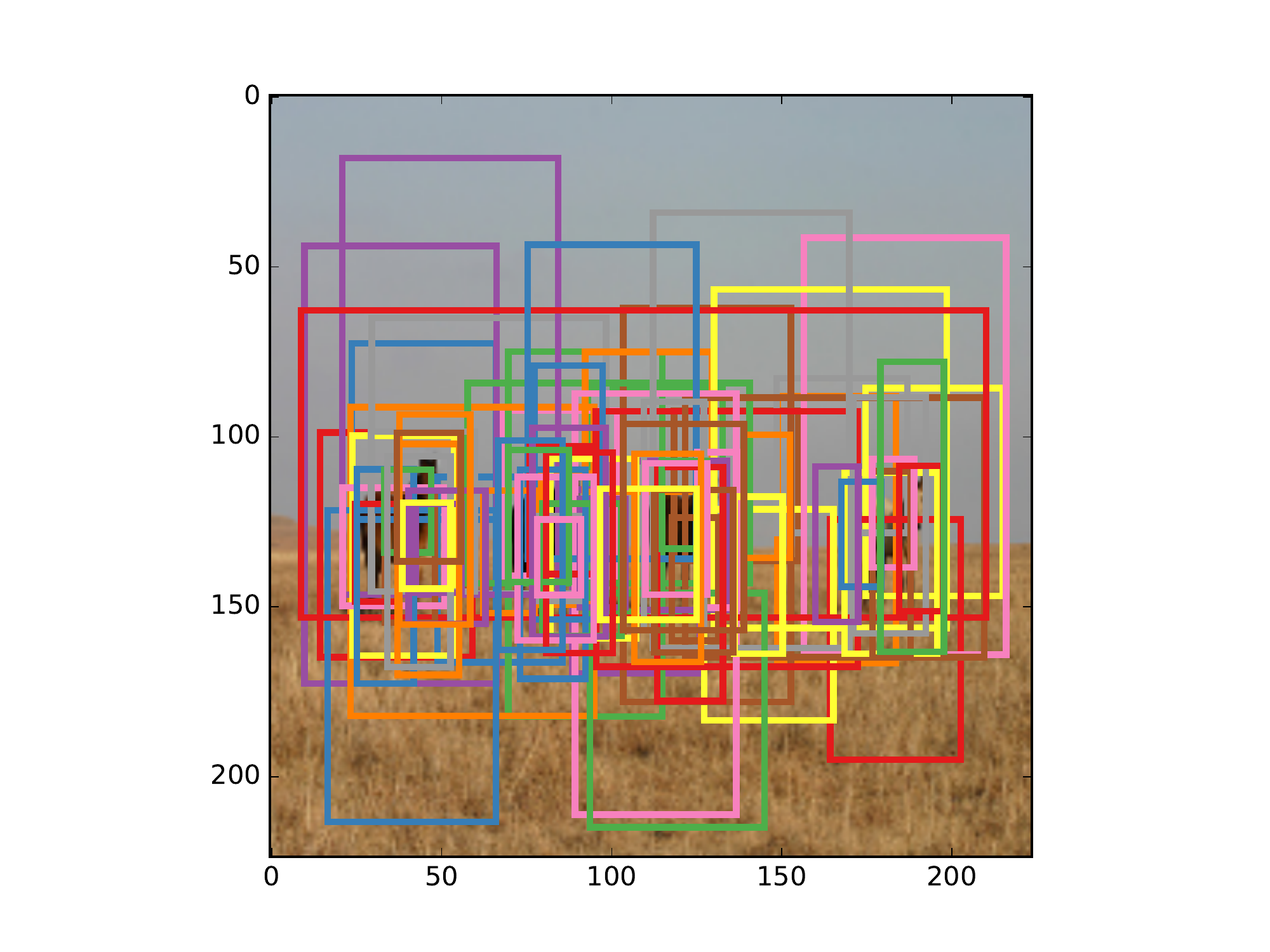}
\\
{\scriptsize (b)} & {\scriptsize (c)}
\\
\includegraphics[clip,trim=5cm 3cm 5cm 2cm,width=.45\linewidth]{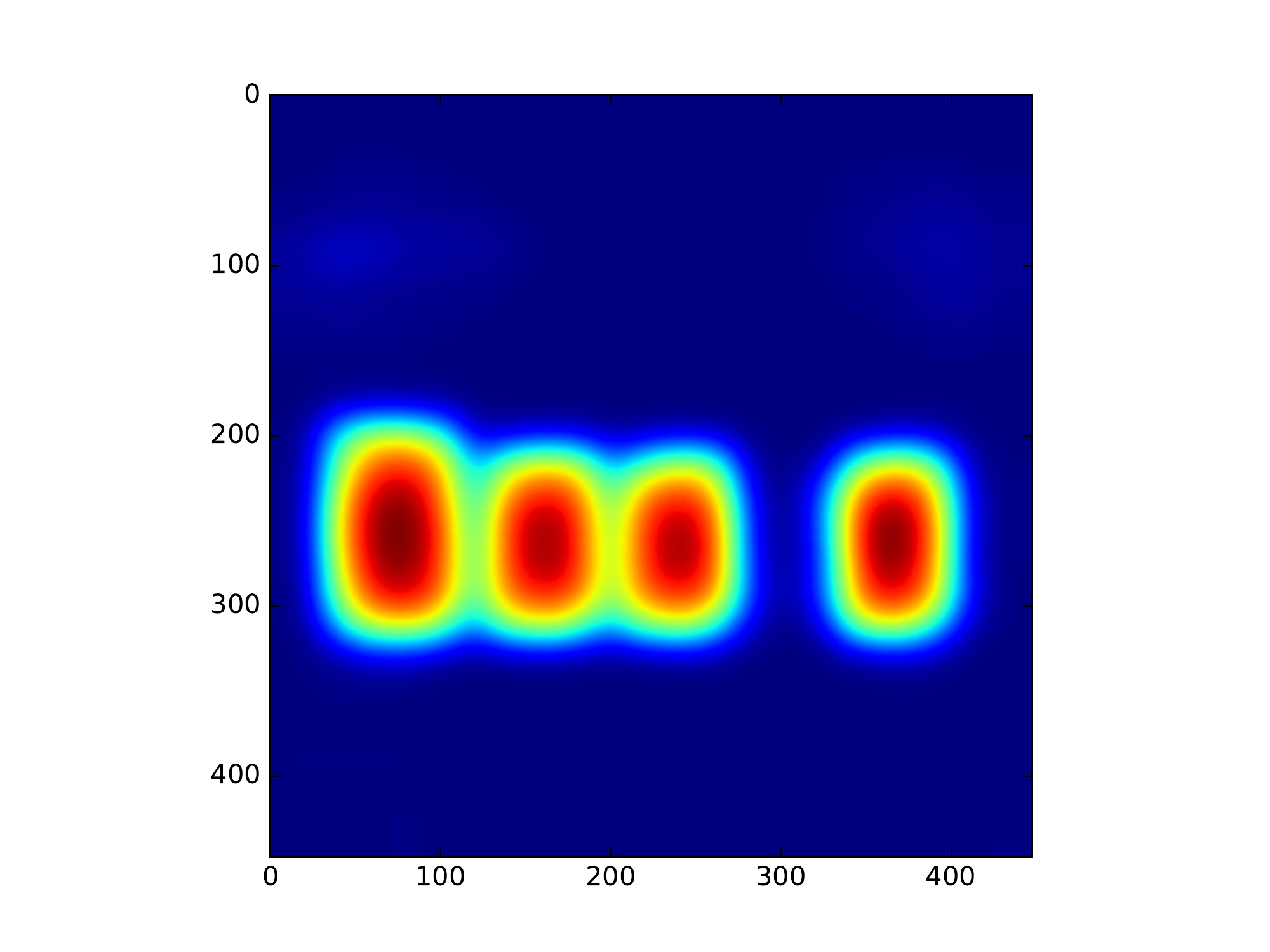}
\ &
\includegraphics[clip,trim=5cm 3cm 5cm 2cm,width=.45\linewidth]{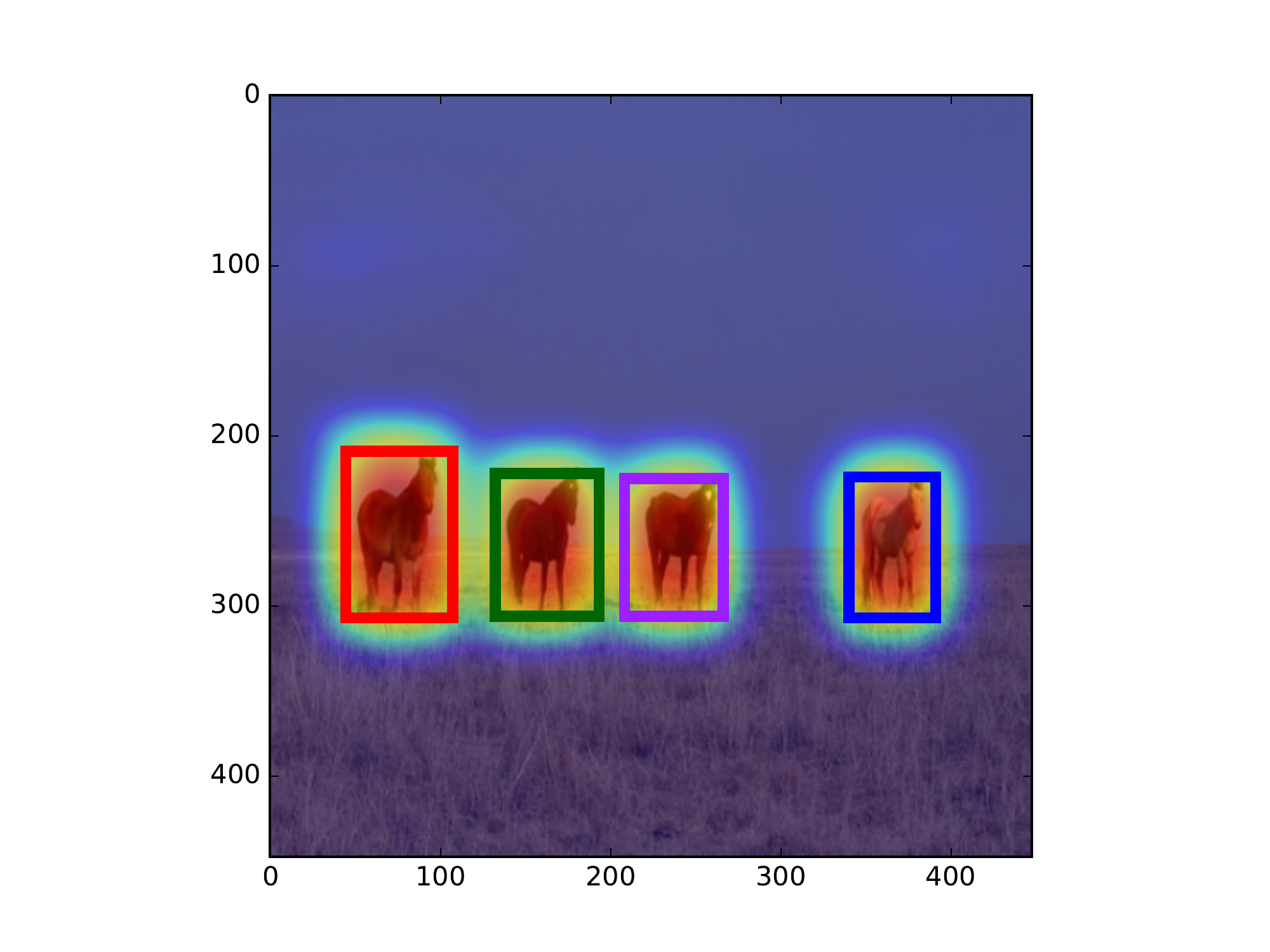}
\\
{\scriptsize (d)} & {\scriptsize (e)}
\end{tabular}
\end{minipage}
\caption{(a) Ground-truth image (b) Output of MultiBox~\cite{DBLP:conf/cvpr/ErhanSTA14} (c) Remaining MultiBox bounding boxes after pruning steps in~\cite{zhang2015SOD} (d) Saliency map predicted by our method without pixel-level labeling (e) Bounding boxes generated by our method from (d) without proposals.}
\label{fig:comparison}
\end{figure}
Saliency detection has been studied under three different scenarios. Early works attempt to predict human eye-fixation over an image~\cite{itti1998model}, while later works increasingly focus on salient foreground segmentation~\cite{lee2016deep,li2015visual,liu2016dhsnet,wang2015deep,zhao2015saliency}, \ie, predicting a dense, pixel-level binary map to differentiate salient objects from background. However, it cannot separate overlapping salient objects and requires pixel-level annotations that are expensive to acquire for large datasets. 
Different from salient foreground segmentation, salient object detection aims to locate and draw bounding boxes around salient objects. It only requires bounding box annotations, which significantly reduces the effort for human labeling, and can easily separate overlapping objects. {These advantages make the problem of salient object detection more valuable to investigate in terms of applicability to the real world.}

With the re-emergence of convolutional neural networks (CNN), computer vision community has witnessed numerous breakthroughs, including salient object detection, thanks to the extraordinary discriminative ability of CNNs~\cite{zhang2015SOD}. 
Prior to CNNs, Some works~\cite{luo2010saliency,suh2003automatic,valenti2009image,DBLP:conf/cvpr/WangWZFZL12} have proposed heuristics to detect single salient object in an image, while others~\cite{feng2011salient,siva2013looking} rank a fixed-sized list of bounding boxes which might contain salient objects without determining the exact detections. However, most of these methods do not solve the \textit{existence} problem, \ie, determining whether any salient objects exist in an image at all, and simply rely on external binary classifiers to address this problem.
Recently, saliency detection based on deep networks has achieved state-of-the-art performance. Zhang \etal~\cite{zhang2015SOD} propose to use the MultiBox proposal network~\cite{DBLP:conf/cvpr/ErhanSTA14} to generate hundreds of candidate bounding boxes that are further ranked to output a compact set of salient objects. A probabilistic approach is proposed to filter and re-rank candidate boxes as a substitution for non-maxima suppression (NMS). 
%
%
To accurately localize salient objects, \cite{zhang2015SOD} requires a large number of class-agnostic proposals covering the whole image (see Figure~\ref{fig:comparison}). However, its precision and recall significantly drop if one only uses tens of boxes. The reason is that generic object proposals have very low success rate of locating an object, \ie, only few of them tightly enclose the ground-truth objects, while most are redundant. Even though additional refinement steps are applied~\cite{zhang2015SOD}, there are still a lot of false positives (see Figure~\ref{fig:comparison}). The additional steps add more overhead and make this framework infeasible for real-time applications.

In this paper, we address this problem by moving towards the success rate of one, \ie, generating the exact number of boxes for salient objects without object proposals. We present an end-to-end deep network for real-time salient object detection, dubbed as RSD. Rather than generating lots of candidate boxes and filtering them, our network directly predicts a saliency map with Gaussian distributions centered at salient objects, and infers bounding boxes from these distributions. Our network consists of two branches trained with multi-task loss to perform saliency map prediction, salient object detection and subitizing simultaneously, all in a single pass within a unified framework. Notably, our RSD with VGG16 achieves more than $100$ fps on a single GPU during inference, significantly faster than existing CNN-based approaches. {To the best of our knowledge, this is the first work on real-time non-redundant bounding box prediction for simultaneous salient object detection, saliency map estimation and subitizing, without object proposals. We also show the possibility of generating accurate saliency maps without pixel-level annotations, formulating it as a weakly-supervised approach that is more practical than fully-supervised approaches.}

Our contributions are summarized as follows. First, we present a unified deep network performing salient object detection, saliency map prediction and subitizing simultaneously in a single pass.
Second, our network is trained with Gaussian distributions centered at ground-truth salient objects that are considered to be more informative and discriminative than bounding boxes to distinguish multiple salient objects. 
Third, our approach outperforms state-of-the-art methods using object proposals by a large margin, and also produces comparable results on salient foreground segmentation datasets, even though we do not use any pixel-level annotations.
Finally, our network achieves $100+$ fps during inference and is applicable to real-time systems. 

\section{Related Works}
{\textbf{Salient object detection} aims to mark important regions by rectangles in an image. Early works assume that there is only one dominant object in an image and utilize various hand-crafted features to detect salient objects~\cite{liu2011learning,DBLP:conf/cvpr/WangWZFZL12}. Salient objects are segmented out by a CRF model~\cite{liu2011learning} or bounding box statistics learned from a large image database~\cite{DBLP:conf/cvpr/WangWZFZL12}. Some works~\cite{feng2011salient,siva2013looking} demonstrate the ability of generating multiple overlapping bounding boxes in a single scene by combining multiple image features. Recently, Zhang \etal~\cite{zhang2015SOD} apply deep networks with object proposals to achieve state-of-the-art results. However, these methods are not scalable for real-time applications due to the use of sliding windows, complex optimization or expensive box sampling process.}

\textbf{Object proposal} have been used widely in object detection, which are either generated from grouping superpixels~\cite{DBLP:conf/cvpr/ArbelaezPBMM14,DBLP:journals/pami/CarreiraS12,DBLP:journals/ijcv/UijlingsSGS13} or sliding windows~\cite{DBLP:journals/pami/AlexeDF12,DBLP:conf/eccv/ZitnickD14}. However, it is a bottleneck to generate a large number of proposals for real-time detection~\cite{DBLP:conf/iccv/Girshick15,DBLP:conf/cvpr/GirshickDDM14}. Recently, deep networks are trained to generate proposals in an end-to-end manner to improve efficiency~\cite{DBLP:conf/cvpr/ErhanSTA14,DBLP:conf/nips/RenHGS15}.
While both SSD~\cite{DBLP:conf/eccv/LiuAESRFB16} and YOLO~\cite{DBLP:journals/corr/RedmonDGF15} instead adopt grid structure to generate candidate boxes, they still rely on a smaller set of proposals. Different from previous methods, our approach does not use any proposals.

\textbf{Object subitizing} addresses the object \emph{existence} problem by learning an external binary classifier~\cite{DBLP:conf/crv/ScharfenbergerWZC13,DBLP:conf/cvpr/WangWZFZL12}. Zhang \etal~\cite{DBLP:conf/cvpr/ZhangMSSBLSPM15} present a salient object subitizing model to remove detected boxes in images with no salient object. While the method in~\cite{zhang2015SOD} addresses \emph{existence} and \emph{localization} problems at the same time, it still requires generating proposals recursively, which is inefficient.

{\textbf{Saliency map prediction} produces a binary mask to segment salient objects from background. While both bottom-up methods using low-level image features~\cite{perazzi2012saliency,yang2013saliency,yang2013saliency,rc,dsr} and top-down methods~\cite{liu2011learning,DBLP:conf/cvpr/WangWZFZL12} have been proposed for decades, many recent works utilize deep neural networks for this task~\cite{zhao2015saliency,li2015visual,wang2015deep,liu2016dhsnet,DBLP:conf/eccv/WangWLZR16,DBLP:journals/tip/LiZWYWZLW16}. Li \etal~\cite{li2015visual} propose a model for visual saliency using multi-scale deep features computed by CNNs. Wang \etal~\cite{wang2015deep} develop two deep neural networks to learn local features and global contrast with geometric features to predict saliency score of each region. In ~\cite{zhao2015saliency}, both global and local context are combined into a single deep network, while a fully convolutional network is applied in ~\cite{DBLP:conf/eccv/WangWLZR16}. 
Note that existing methods heavily rely on pixel-level annotations~\cite{zhao2015saliency,liu2016dhsnet,DBLP:conf/eccv/WangWLZR16} or external semantic information, \ie, superpixels~\cite{DBLP:journals/tip/LiZWYWZLW16}, which is not feasible for large-scale problems, where human labeling is extremely sparse.
In contrast, our approach, as a weakly-supervised approach, only requires bounding box annotations and produces promising results as a free by-product, along with salient object detection and subitizing.}

\section{Proposed Approach}
Existing detection methods based on CNNs and object proposals~\cite{DBLP:conf/cvpr/ErhanSTA14,DBLP:conf/iccv/Girshick15,DBLP:conf/cvpr/GirshickDDM14,DBLP:conf/nips/RenHGS15,zhang2015SOD} convert the problem of selecting candidate locations in an image in the spatial domain to a parameter estimation problem, \eg, finding independent numbers as the coordinates of the bounding boxes. They use as many as billions of parameters in fully connected (\emph{fc}) layers~\cite{DBLP:conf/iccv/Girshick15,DBLP:conf/cvpr/GirshickDDM14}, which is computationally expensive and increases the possibility of overfitting on small datasets.
In contrast, our RSD approach discards proposals and directly solves the problem in the spatial domain. It reduces the number of parameters from billions to millions and achieves real-time speed. We predict a rough saliency map, from which we infer the exact number of boxes as the ground-truth objects based on the guidance of the subitizing output of our network. This unified framework addresses three closely related problems, saliency map prediction, subitizing and salient object detection, without allocating separate resources for each.


\subsection{Network architecture}

\begin{figure}[t]
\centering
\includegraphics[width=0.45\textwidth]{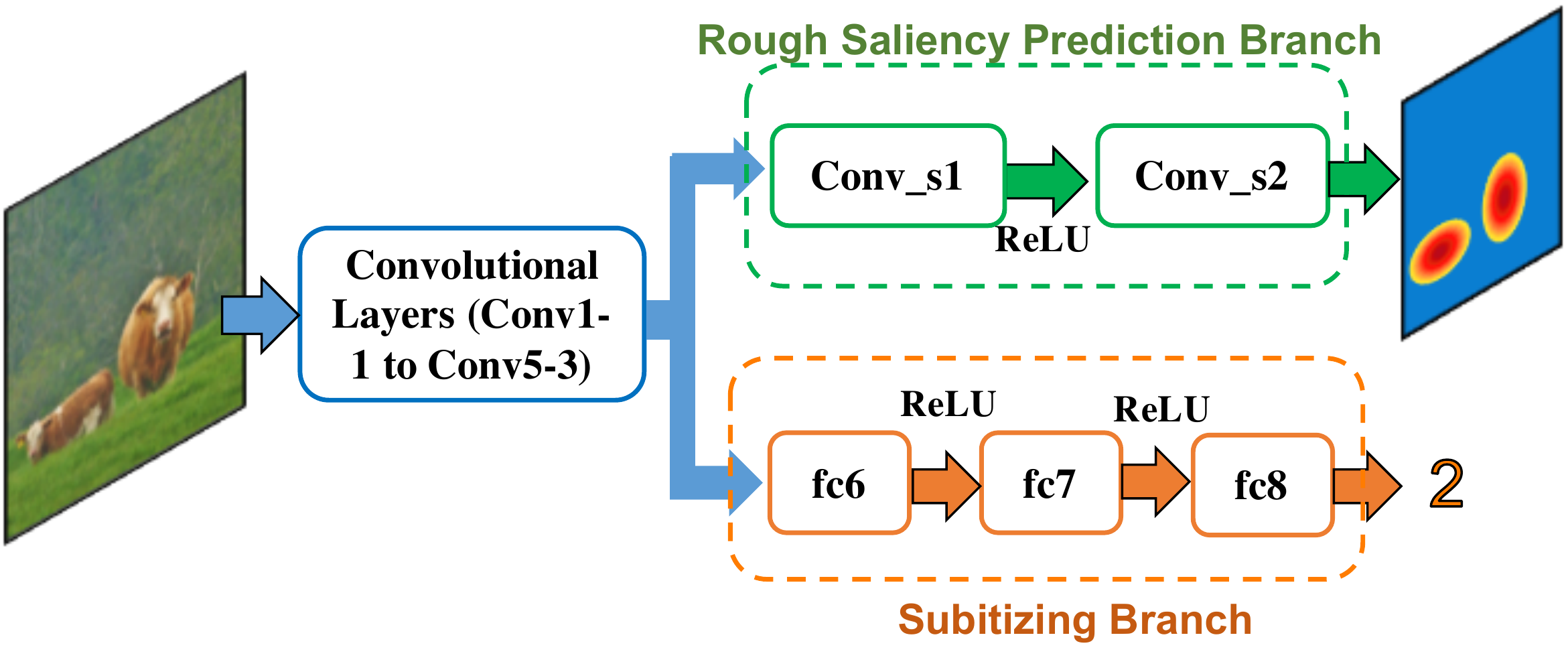}
\caption{Our RSD network based on VGG16.}
\label{fig:network}
\end{figure}

Our network is composed of the following components (see Figure~\ref{fig:network}). Images first go through a series of convolutional layers that can be any widely used models, such as VGG16 and ResNet-50. Specifically, we use the convolutional layers \emph{conv1\_1} through \emph{conv5\_3} from VGG16~\cite{DBLP:journals/corr/SimonyanZ14a}, and \emph{conv1} through \emph{res4f} from ResNet-50~\cite{DBLP:journals/corr/HeZRS15}. 
These layers capture low-level cues and high-level visual semantics.
Two branches are connected to the feature maps from the last convolutional layer: saliency map prediction branch and subitizing branch. 
The saliency map prediction branch consists of two convolutional layers, \emph{conv\_s1} and \emph{conv\_s2}, to continue processing the image in the spatial domain and produce a rough saliency map. 
The layer \emph{conv\_s1} has 80 $3\times3$ filters to produce intermediate saliency maps conditioned on different latent distributions of the objects (\eg, latent object categories). For instance, each of the 80 filters can be seen as a way to generate a rough saliency map for a specific type of category.
The layer \emph{conv\_s2} summarizes these conditional maps into a single saliency map by a $1\times1$ filter followed by a sigmoid function.  
The subitizing branch predicts the number of salient objects that can be 0, 1, 2, or $3+$. It contains the final \emph{fc} layers for VGG16, and all the remaining convolutional layers followed by a global average pooling layer and a single \emph{fc} layer for ResNet-50.

\subsection{Ground-truth preparation}
The ground-truth for salient object detection only contains a set of numbers defining coordinates of bounding boxes tightly enclosing the objects. Although we can generate a binary mask based on these coordinates, \ie, 1 inside the bounding boxes and 0 elsewhere, it cannot separate overlapping objects or encode non-rigid boundaries well.

To address this problem, we propose to generate Gaussian distributions to represent salient objects, and use images with Gaussian distributions as ground-truth saliency maps. Given a ground-truth bounding box $B_i$ for an image with width $W$ and height $H$, let $(cx_i,cy_i,w_i,h_i)$ represent the coordinates of its center, width, and height.
If the network has the stride of $s$ at the beginning of the saliency map prediction branch (\eg, 16 for VGG16), the ground-truth saliency map $\mathcal{M}_g$ is an image of size $\lfloor W/s \rfloor \times \lfloor H/s\rfloor$, where $\lfloor.\rfloor$ is the floor function. Its $(x,y)$-th element is then defined as
\begin{equation}
\mathcal{M}_g (x,y) = \sum_{i=1}^N e^{-\frac{1}{2}(\mathbf{v}_{xy}-\mathbf{\mu}_i)^T\mathbf{\Sigma_i} (\mathbf{v}_{xy}-\mathbf{\mu}_i)} 
\mathbbm{1}_{\mathbf{v}_{xy} \in \mathcal{R}_{B_i}},
\label{eq:gt}
\end{equation}
where $\mathbf{v}_{xy} = [x,y]^T$ is the location vector, and $\mathbf{\mu}_i = [\lfloor cx_i/s \rfloor,\lfloor cy_i/s \rfloor]^T$ is the mean value.
$N$ is the number of ground-truth bounding boxes in the image.
$\mathcal{R}_{B_i}$ represents the ROI inside bounding box $B_i$.
$\mathbbm{1}$ is an indicator function.
The covariance matrix $\mathbf{\Sigma}_n$ can be represented as 
\begin{equation}
\mathbf{\Sigma}_i = \begin{bmatrix}
\lfloor \frac{w_i}{s}\rfloor ^2/4,0\\
0,\lfloor \frac{h_i}{s}\rfloor ^2/4
\end{bmatrix}.
\end{equation}

\begin{figure}[t]
\begin{center}
\begin{tabular}{ccc}
\frame{\includegraphics[clip,trim=4.5cm 1.5cm 4cm 1.5cm, width=0.25\linewidth]{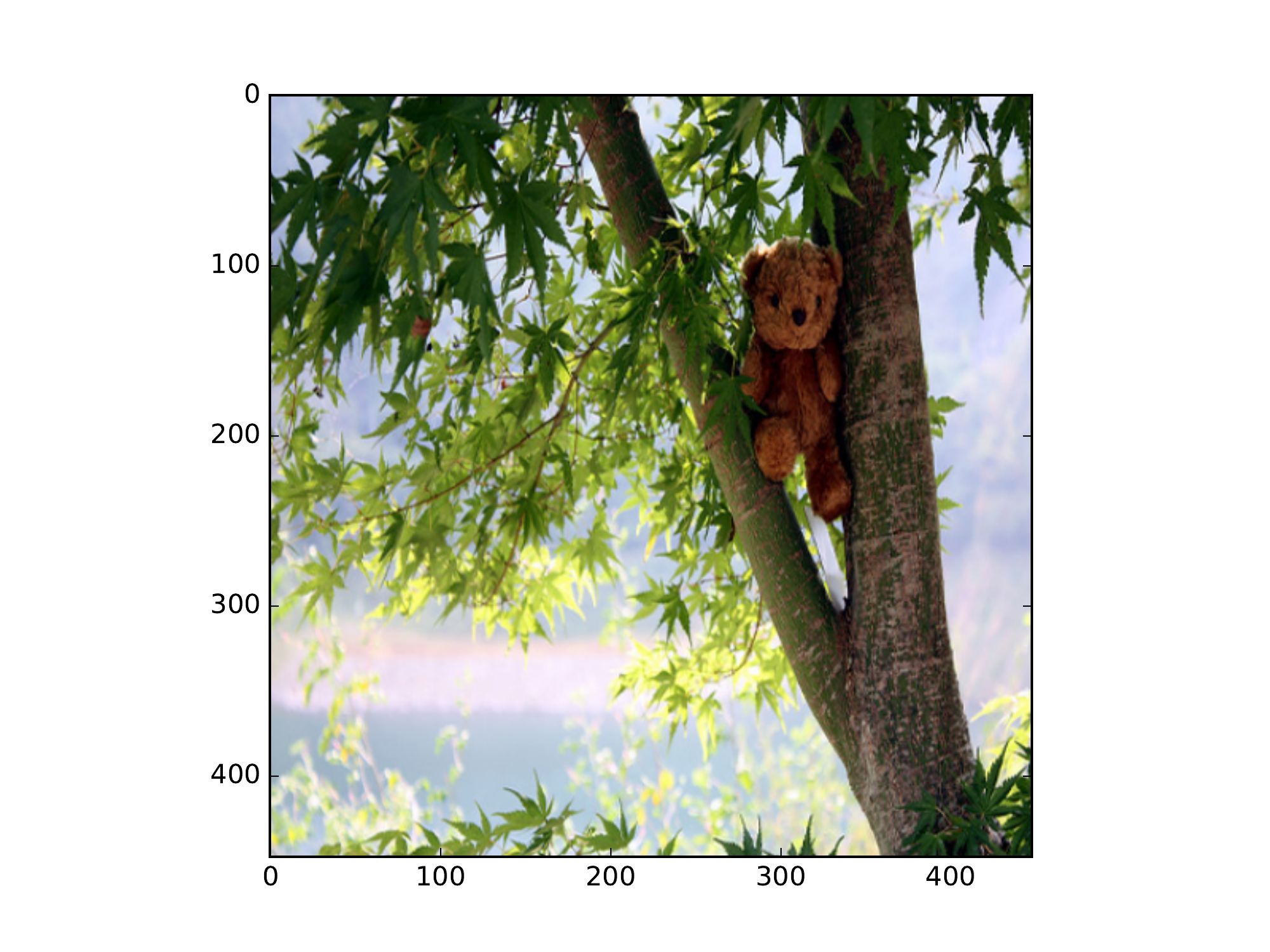}}
&
\frame{\includegraphics[clip,trim=4.5cm 1.5cm 4cm 1.5cm, width=0.25\linewidth]{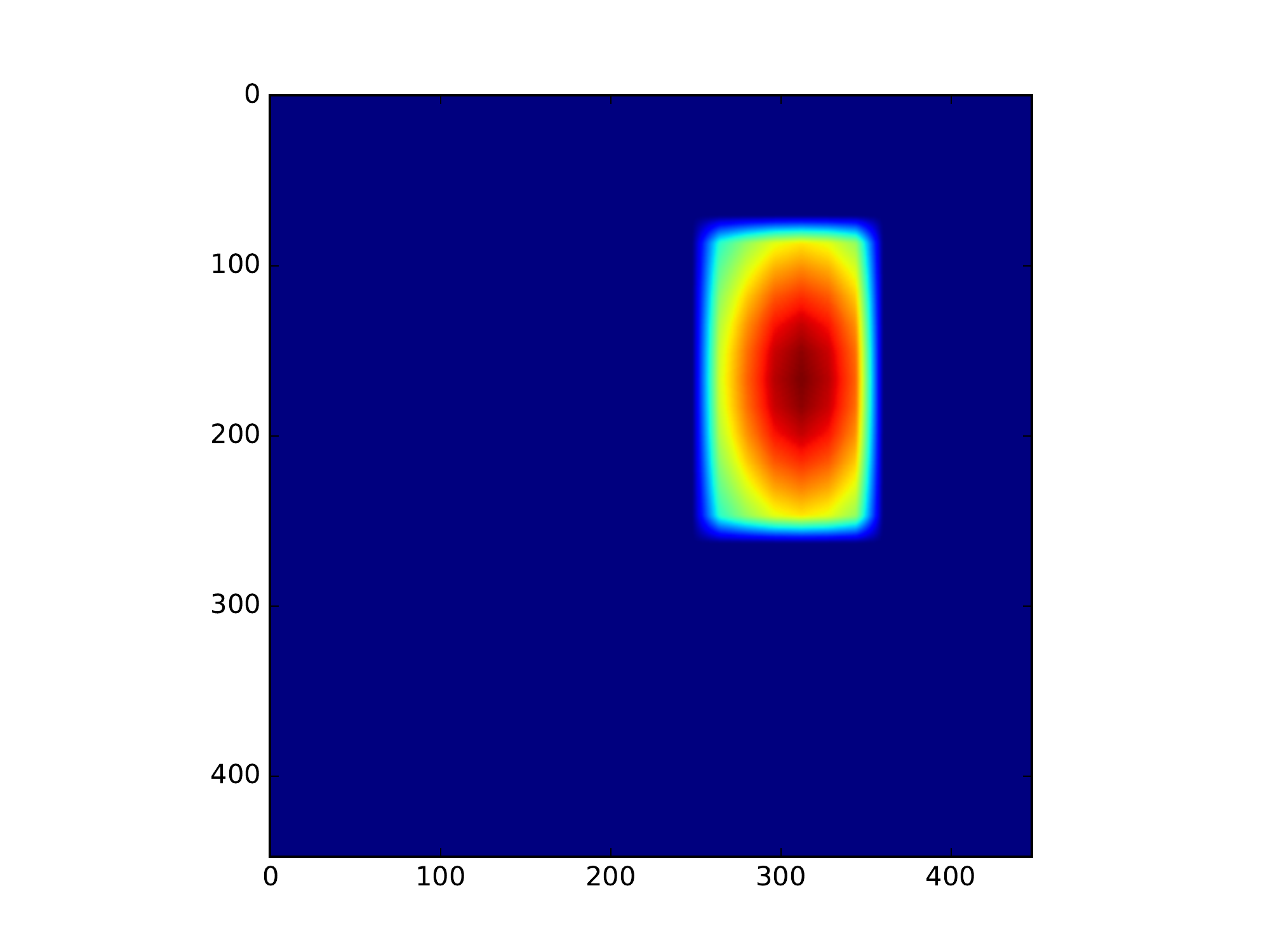}}
&
\frame{\includegraphics[clip,trim=4.5cm 1.5cm 4cm 1.5cm, width=0.25\linewidth]{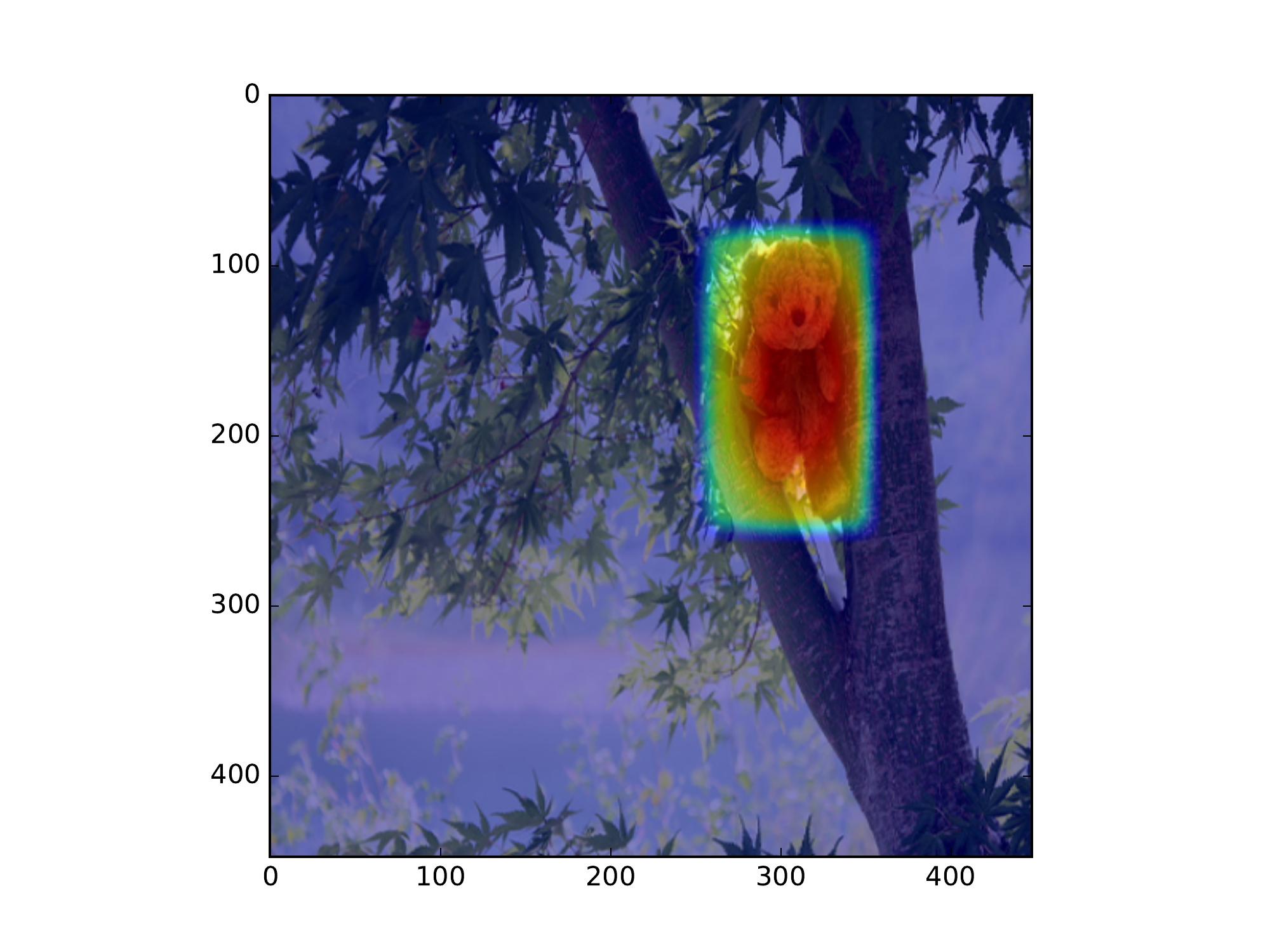}}\\
\frame{\includegraphics[clip,trim=4.5cm 1.5cm 4cm 1.5cm, width=0.25\linewidth]{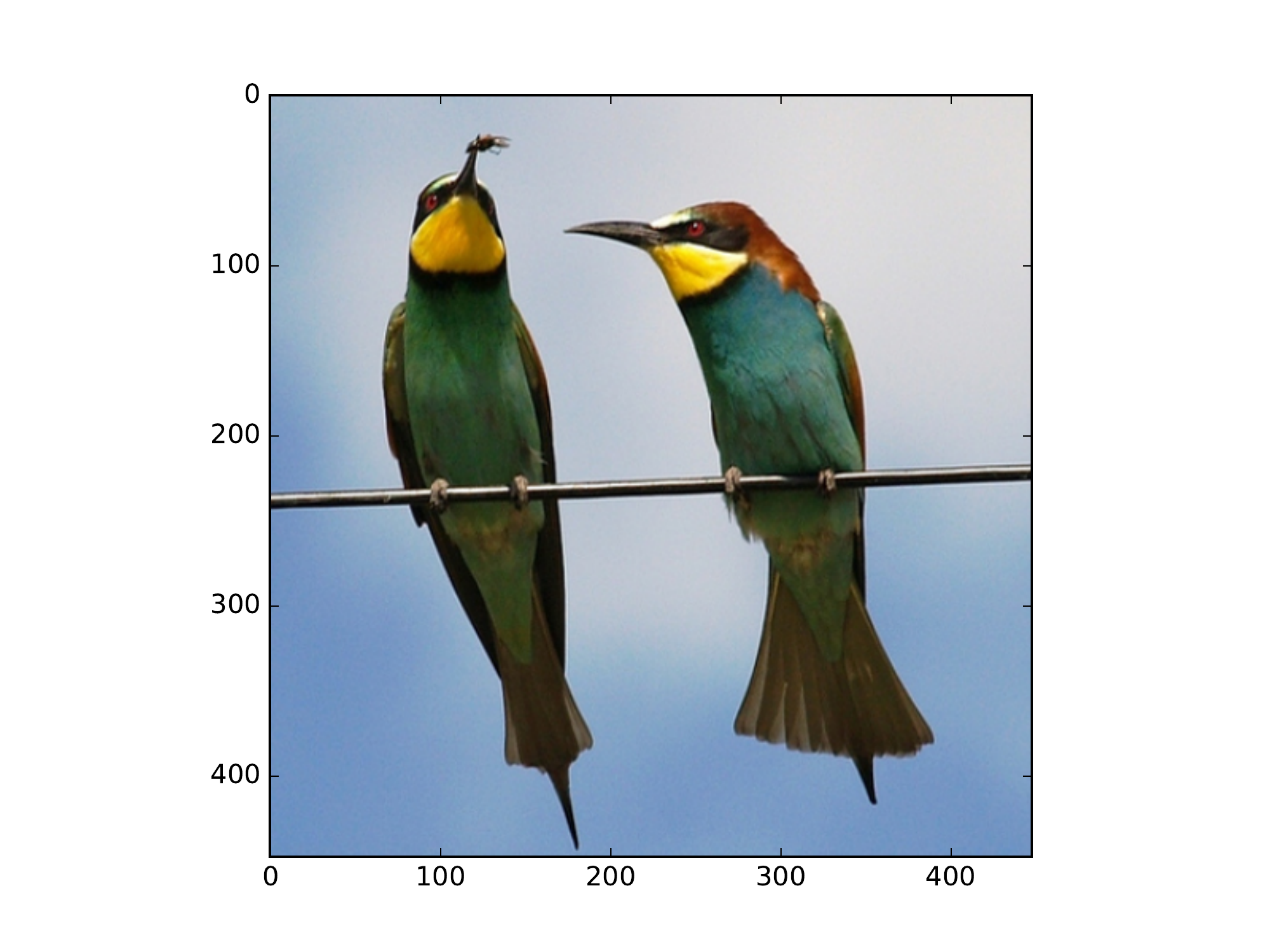}}
&
\frame{\includegraphics[clip,trim=4.5cm 1.5cm 4cm 1.5cm, width=0.25\linewidth]{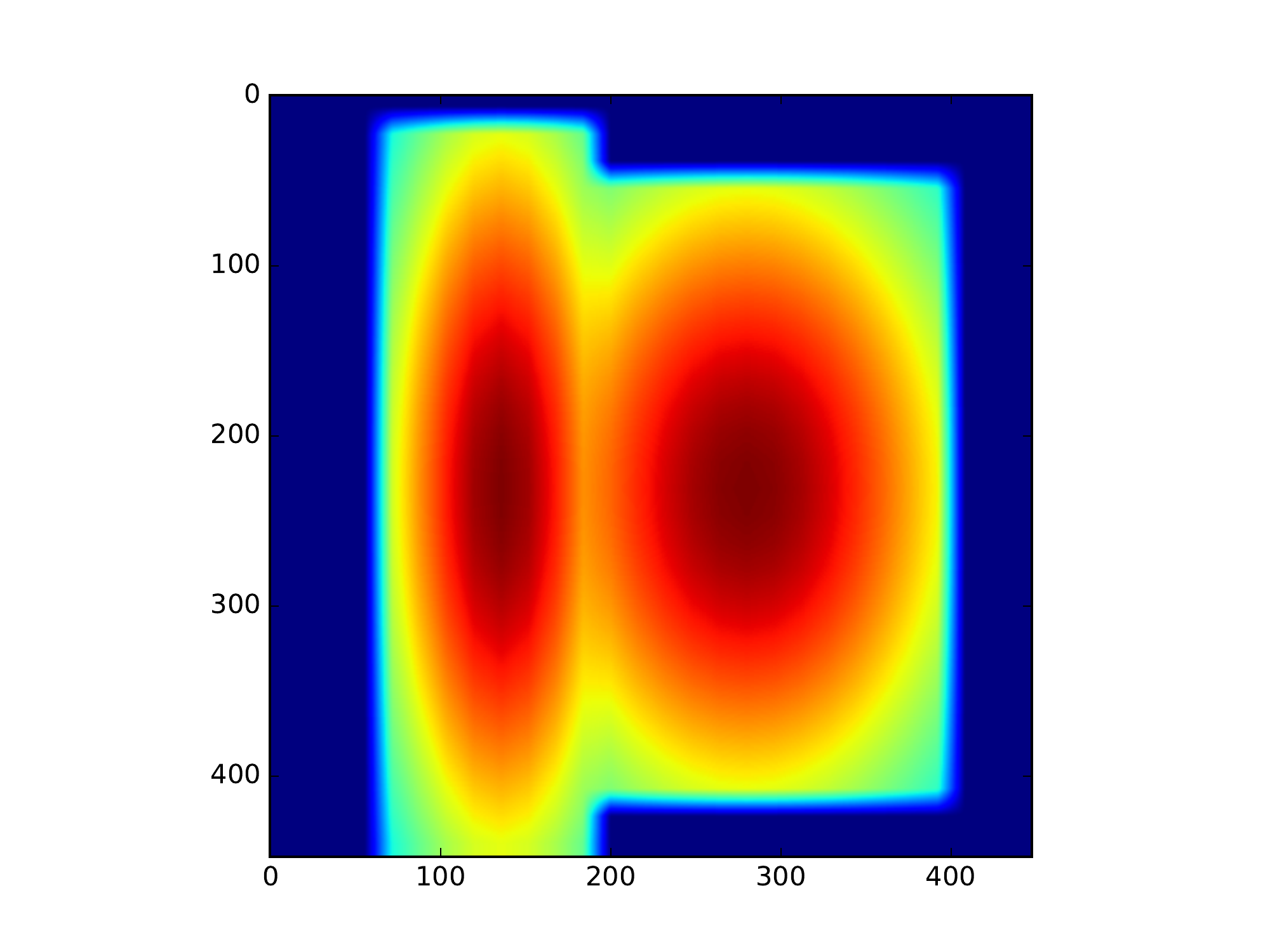}}
&
\frame{\includegraphics[clip,trim=4.5cm 1.5cm 4cm 1.5cm, width=0.25\linewidth]{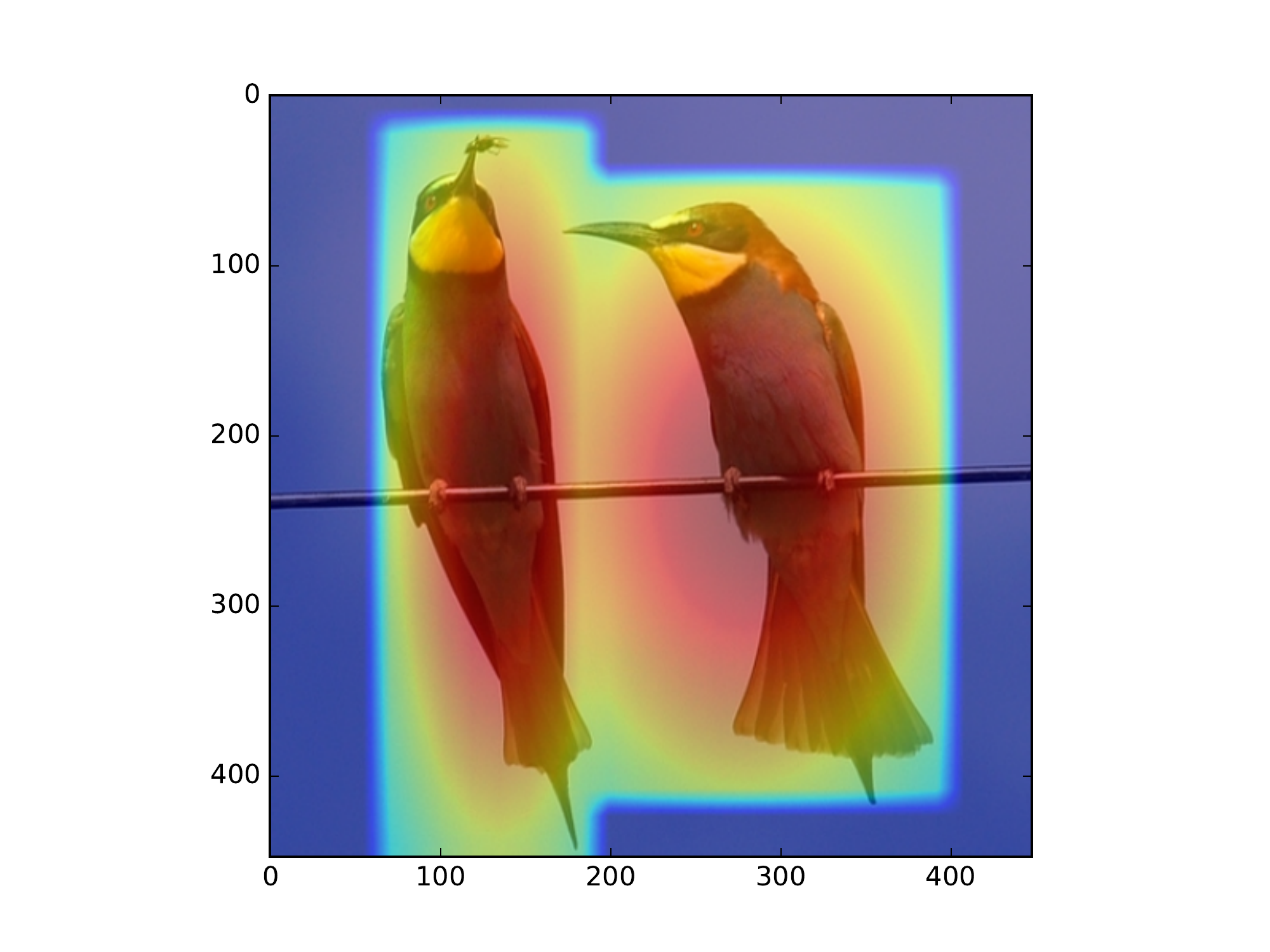}}
\end{tabular}
\end{center}
\caption{Samples of generated ground-truth saliency maps.}
\label{fig:generated_gt}
\end{figure}

By \eqref{eq:gt}, we represent each bounding box as a normalized 2D Gaussian distribution, located at the center of the bounding box, with the co-variance determined by the bounding box's height and width and truncated at the box boundary. 
As shown in Figure~\ref{fig:generated_gt}, the Gaussian shape ground-truth provides better separation for multiple objects compared to rectangular bounding boxes. It also naturally acts as spatial weighting to the ground-truth, so that the network learns to focus more on the center of objects instead of being distracted by background.


\subsection{Multi-task loss}
Our network predicts a saliency map from an image and performs subitizing as well. During training, the network tries to minimize the difference between the ground-truth map and the predicted saliency map. 
Although Euclidean loss is widely used to measure the pixel-wise distance, it pushes gradients towards 0 if the values of most pixels are 0, which is the case in our application when there are images with no salient object. Therefore, we use a weighted Euclidean loss to better handle this scenario, defined as
\begin{equation}
\ell_{sal} (\mathbf{x},\mathbf{g})= \frac{1}{2d} \sum_{i=1}^{d} \alpha^{\mathbbm{1}_{g_i>0.5}} (x_i-g_i)^2 ,
\label{eq:weighted_euc_loss}
\end{equation}
where $\mathbf{x} \in \mathbb{R}^d$ and $\mathbf{g} \in \mathbb{R}^d$ are the vectorized predicted and ground-truth saliency maps, respectively. $x_i$ and $g_i$ represent the corresponding $i$-th element.
$\alpha$ is a constant weight set to 5 in all our experiments.
Essentially, the loss $\ell_{sal}$ assigns more weight to the pixels with the ground-truth value higher than 0.5, compared to those with the value close to 0. 
In this way, the problem of gradient vanishing is alleviated since the loss focuses more on pixels belonging to the real salient objects and is not dominated by background pixels. As a classifier, the subitizing branch minimizes the multinomial logistic loss $\ell_{sub}(y,n)$ between the ground-truth number of objects $n$, and the predicted number of objects $y$.
The two losses are combined as our final multi-task loss
\begin{equation}
\ell(\mathbf{x},\mathbf{g},y,n) = \ell_{sal}(\mathbf{x},\mathbf{g}) + \lambda \ell_{sub}(y,n),
\label{eq:full_loss}
\end{equation}
where $\lambda$ is a weighting factor to balance the importance of the two losses. We set $\lambda=0.25$ to make the magnitude of the loss values comparable. 

\subsection{Training}
\label{sec:training}
The loss in \eqref{eq:full_loss} defines a multi-task learning problem previously studied by other vision applications~\cite{DBLP:conf/iccv/Girshick15,DBLP:conf/nips/RenHGS15}. It reduces required resources by sharing  weights between different tasks, and acts as a regularization to avoid over-fitting. We use standard SGD with momentum and weight decay for learning the parameters of the network. 

To ensure fair comparison, we adopt the same two-stage training scheme suggested by~\cite{zhang2015SOD}. In the first stage, we initialize the network using the weights trained on ImageNet~\cite{DBLP:conf/cvpr/DengDSLL009} for classification and fine-tune it on the ILSVRC2014 detection dataset~\cite{ILSVRC15} by treating all objects as salient. 
In the second stage, we continue fine-tuning the network on the SOS dataset~\cite{DBLP:conf/cvpr/ZhangMSSBLSPM15} for salient object subitizing and detection. Although all the images in SOS are annotated for subitizing, some are not labled for detection. Therefore, we do not back-propagate gradients to our saliency map prediction branch for these images labeled as containing salient objects but without bounding box annotations. The loss function to fine-tune on the SOS dataset $\ell(\mathbf{x},\mathbf{g},y,n)$ is denoted as $\mathbbm{1}_{n_{box}\ge n} \ell_{sal}(\mathbf{x},\mathbf{g}) + \lambda \ell_{sub}(y,n)$,
where $n_{box}$ indicates the number of bounding box annotations in the image.

\subsection{Bounding box generation}
{Our method leverages the saliency prediction branch and subtizing branch to infer the correct number and location of bounding boxes. Given the output of the subitizing branch $n_{sub}$ and the rough saliency prediction map $\mathcal{M}$, the goal is to find $K$ Gaussians $\mathcal{N}(\mu_k, \sigma_k^2), k=1,\dots,K$ that align with the predicted saliency map and are supported by the subitizing output, which can be formulated as
}
\begin{align}
\argmin_{\{\mathbf{\mu}_k\},\{\mathbf{\sigma}_k\},K} &\ell_{s}(\sum_{k=1}^K \mathcal{N}(\mu_k, \sigma_k^2),\mathcal{M})\nonumber \\
& +\mathbbm{1}_{n_{sub}<n_{M}}
\mathbbm{1}_{\theta_{sub}>\theta_c}\ell_c(K,n_{sub}),
\label{eq:bbox}
\end{align}
{where $\ell_s$ captures the discrepancy between the predicted saliency map and the generated Gaussian map. $\ell_c$ measures the disagreement between the subitizing branch and the number of Gaussians, from which boxes' locations can be inferred. $n_M$ is the maximal possible output of the subitizing branch, \ie, maximal number of salient objects. $\theta_{sub}$ is the confidence score of the subitizing branch, and $\theta_c$ is a fixed confidence threshold that will be discussed later. In other words, if $n_{sub}= n_M$ or $\theta_{sub}$ is lower than the threshold, we rely only on the predicted saliency map to determine the number and locations of salient objects.
Since solving \eqref{eq:bbox} directly is intractable, we propose a feasible and efficient greedy algorithm to approximate it, which predicts the center and scale of boxes, while optimizing the objective function. 
\ignore{In section \ref{sec:single_detection}, we present the algorithm assuming a single object inside the image and generalize the algorithm for multiple objects in section \ref{sec:multi_detection}. Moreover to tackle the \emph{existence} sub-problem, we use the output of subitizing branch $n_{sub}$ and do not generate any bounding boxes if $n_{sub}=0$.}
If $n_{sub}=0$, our method does not generate any bounding boxes; otherwise it generates either single or multiple objects.
}

\subsubsection{Single salient object detection}
\label{sec:single_detection}

If $n_{sub}=1$ and the confidence of subitizing branch $c$ is larger than a pre-defined threshold $\theta_c$, we think there is only a single object.
We 
convert the saliency map $\mathcal{M}$ to a binary map $\mathcal{M}_b$ using $\theta_c$, and then perform contour detection using the fast Teh-Chin chain approximation~\cite{teh1989detection} on $\mathcal{M}_b$ to detect connected components $\mathcal{C}$ and infer bounding boxes $\mathcal{B}$.
We define the ROI of box $B_i$ on the original map as $\mathcal{M}_{\mathcal{R}_{B_i}}$, from which the maximal value is assigned as its score $S_{B_i}$.
The one with the highest score is selected as the salient object.
The entire process is summarized in Algorithm \ref{alg:single}.
\begin{algorithm}[t]
\caption{Single bounding box generation}
\label{alg:single}
\setstretch{0.9}
\begin{algorithmic}[1]
{\small
\Statex \textbf{Parameters:} $\theta_c \gets$ Fixed confidence threshold
\Procedure{SingleDetect}{$\mathcal{M}$}
\State $\mathcal{M}_b$ $\gets$ \texttt{threshold}$(\mathcal{M}, \theta_c)$
\State $\mathcal{C}$ $\gets$ \texttt{detectContours}$(\mathcal{M}_b)$
\State $\mathcal{B}$ $\gets$  \texttt{generateBoxes}$(\mathcal{C})$
\For {$B_i \in \mathcal{B}$}
	\State $\mathcal{R}_{B_i}$ $\gets$  ROI of box $B_i$
	\State $S_{B_i}$ $\gets$ $\max \mathcal{M}_{\mathcal{R}_{B_i}}$
\EndFor
\State \textbf{return} $B = \argmax\limits_{B_i} S_{B_i}$
\EndProcedure
}
\end{algorithmic}
\end{algorithm}

\subsubsection{Multiple salient object detection}
\label{sec:multi_detection}

If $n_{sub}>1$, there may be multiple salient objects.
When {the subitizing branch outputs $n_M$, or its confidence score $\theta_{sub}<\theta_c$, we rely on the predicted saliency map $\mathcal{M}$ to find as many reliable peaks as possible. Therefore, our method is able to detect arbitrary number of salient objects (see Table~\ref{tab:existance}).}  Otherwise, we try to find at least $n_{sub}$ {reliable} peaks.
A multi-level thresholding scheme is proposed for robust peak detection {and balancing the losses, $\ell_s$ and $\ell_c$, in \eqref{eq:bbox}}. Starting from a high threshold, a peak $P_i=[x_i,y_i]^T$ is discovered from $\mathcal{M}$ following similar steps in Algorithm~\ref{alg:single}. Peaks are continuously identified and added to the set of peaks $\mathcal{P}$ by reducing the threshold and repeating the process until the cardinality of $\mathcal{P}$ reaches or exceeds $n_{sub}$. {Note that the predicted number of boxes depends on both the subitizing and saliency map prediction branches, which could be less or more than $n_{sub}$, if no threshold can separate $n_{sub}$ reliable peaks or more peaks are detected in different thresholds.}

After the initial set of peaks are determined, peaks with low confidence are treated as noise and removed.
Then we try to find separating lines $\mathcal{L}$ to isolate remaining peaks into different non-overlapping regions.
Each line perpendicular to the line segment connecting a pair of peaks is associated with a score. The score is the maximal value of the pixels this line passes on $\mathcal{M}$. The one with the minimal score is selected as the separating line of the two peaks. In this way, we ensure that the separating line passes through the boundaries between objects rather than the objects themselves. 
These lines $\mathcal{L}$ divide $\mathcal{M}$ into different regions. 
Finally, for each peak $P_i \in \mathcal{P}$, we apply Algorithm~\ref{alg:single} to its corresponding region on the saliency map to obtain a bounding box.
Algorithm~\ref{alg:multiple} summarizes the process.
\begin{algorithm}[t]
\caption{Multiple bounding box generation}
\label{alg:multiple}
\setstretch{0.9}
\begin{algorithmic}[1]
{\small
\Statex \textbf{Parameters:}
\Statex $\mathbf{\Theta} \gets $ Fixed thresholds for peak detection
\Statex $\theta_c \gets$ Fixed confidence threshold
\Procedure{MultiDetect}{$\mathcal{M}$}
\State $\mathcal{P}$ $\gets$ $\emptyset$, $\mathcal{B}$ $\gets$ $\emptyset$
\While {$\left\vert\mathcal{P}\right\vert \leq n_{sub}$}
	\For {$\theta_i \in \mathbf{\Theta}$}
			\State $\mathcal{M}_b$ $\gets$ \texttt{threshold}$(\mathcal{M}, \theta_i)$
      		\State $\mathcal{C}$ $\gets$ \texttt{detectContours}$(\mathcal{M}_b)$ 
            \For {$C_j \in \mathcal{C}$}
      			\State $\mathcal{P} \gets \mathcal{P} \cup \argmax \mathcal{M}_{C_j}$
            \EndFor
  \EndFor
\EndWhile
\State $\mathcal{P}$ $\gets$ $\mathcal{P} \setminus P_i$ \textbf{where}  $\mathcal{M}(P_i) <\theta_c$, $\forall i$
\State $\mathcal{L}$ $\gets$ \texttt{findSeparatingLines} $(P_i,P_j)$, $\forall i\neq j$
\For {$P_i \in \mathcal{P}$}
\State $\mathcal{R}_{P_i}$ $\gets$  ROI formed by $\mathcal{L}$ containing $P_i$
\State $\mathcal{B}$ $\gets$ $\mathcal{B}$ $\cup$ \Call{SingleDetect}{$\mathcal{M}_{\mathcal{R}_{P_i}}$}
\EndFor
\State \textbf{return} $\mathcal{B}$
\EndProcedure
}
\end{algorithmic}
\label{alg:train}
\end{algorithm}

\vspace{-2mm}
\section{Experiments}
\subsection{Experimental setup}

{\flushleft \textbf{Datasets.}} We evaluate our salient object detection method on {four} datasets, MSO\cite{DBLP:conf/cvpr/ZhangMSSBLSPM15}, {PASCAL-S\cite{li2014secrets}}, MSRA\cite{liu2011learning} and DUT-O\cite{yang2013saliency}. The MSO dataset is the test set of the SOS dataset annotated for salient object detection. It contains images of multiple salient objects and many background images with no salient object. {PASCAL-S is a subset of PASCAL VOC 2010 validation set~\cite{Everingham15} annotated for saliency segmentation problem. It contains images with multiple salient objects and 8 subjects decided on the saliency of each object segment. As suggested by~\cite{li2014secrets}, we define salient objects as those having a saliency score of at least 0.5, \ie, half of the subjects believe that the object is salient, and consider the surrounding rectangles as the ground-truth.}
We also compare the performance of our method and existing methods for subitizing on this dataset. The MSRA and DUT-O datasets only contain images of single salient object. For every image in the MSRA and DUT-O datasets, five raw bounding box annotations are provided, which are later converted to one ground-truth following the same protocol in~\cite{zhang2015SOD}. We use only the SOS dataset for training and others for evaluation. To verify that our RSD can also generate accurate pixel-wise saliency map, we additionally compare our method with existing methods on ESSD~\cite{hs} and PASCAL-S~\cite{li2014secrets} datasets. 

{\flushleft \textbf{Parameters and settings.}} In Algorithm \ref{alg:single} and \ref{alg:multiple}, we set $\theta_c = 0.7$ as our strong evidence threshold, $\mathbf{\Theta} = [0.95,0.9,0.8,0.6]$ as our peak detection thresholds, and use vertical and horizontal lines as our separating lines. 
In our real-time network based on VGG16, we use an image size of $224 \times 224$ and for our network based on ResNet-50 we use $448 \times 448$ instead. We smooth predicted saliency maps by a Gaussian filter before converting them to binary maps. We use a Gaussian kernel with standard deviation of $10$ for $448 \times 448$ input and $2$ for $224 \times 224$ input. 
In the first training step, we use Xavier initialization for \emph{conv\_s1} and \emph{conv\_s2} and Gaussian initializer for the final \emph{fc} layer in the subitizing branch. For fine-tuning on SOS, we use a momentum of $0.9$, weight decay of $5e^{-4}$, and learning rates of $1e^{-4}$ and $1e^{-5}$ for our VGG16 and ResNet-50 based methods, respectively. All timings are measured on an NVIDIA Titan X Pascal GPU, and a system with 128GB RAM and Intel Core i7 6850K CPU.

\subsection{Results}

\begin{figure*}[t]
    \centering
        \includegraphics[height=.17\textwidth]{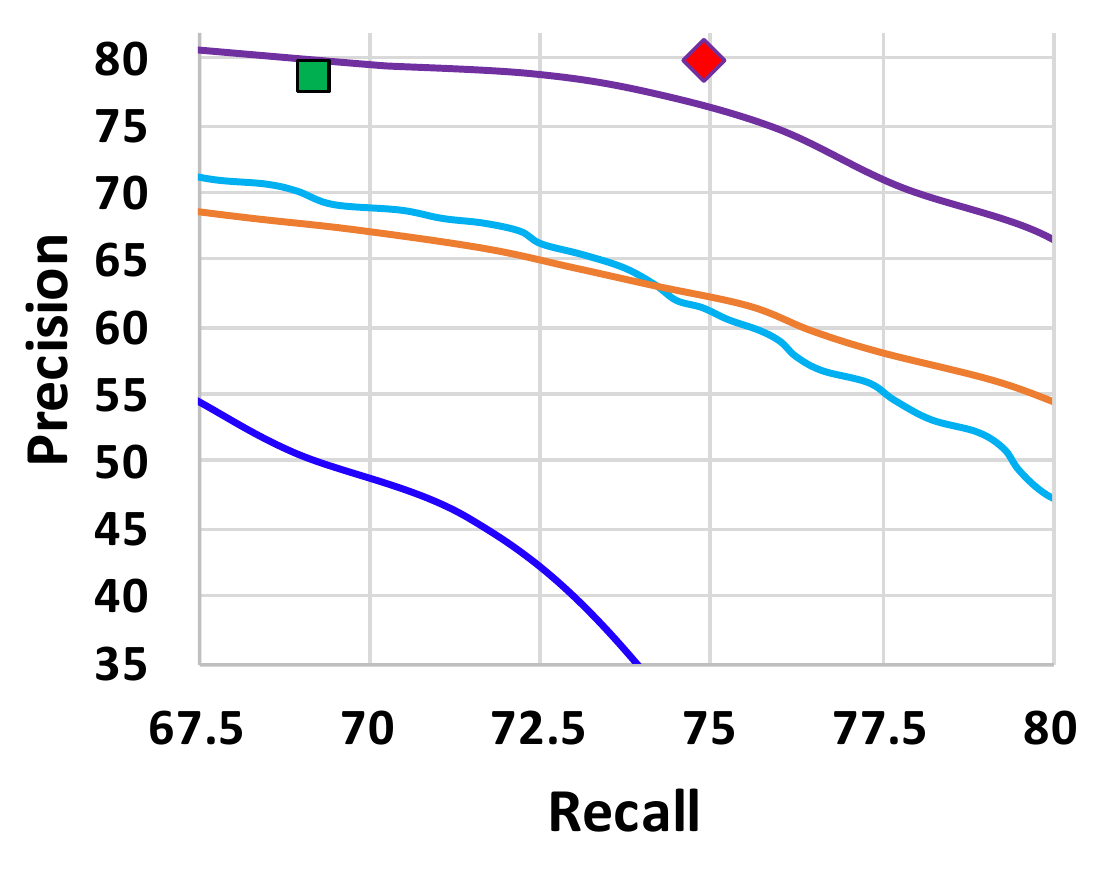}
        \includegraphics[height=.17\textwidth]{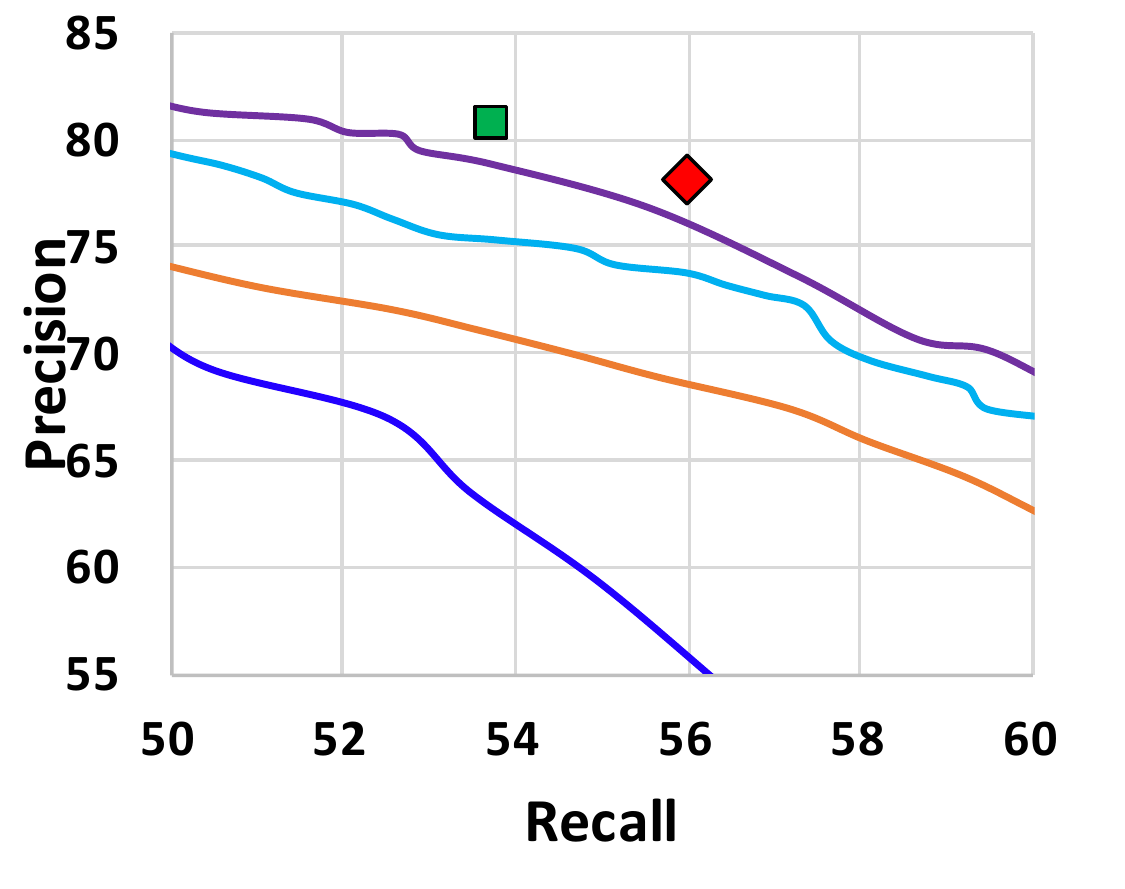}
        \includegraphics[height=.17\textwidth]{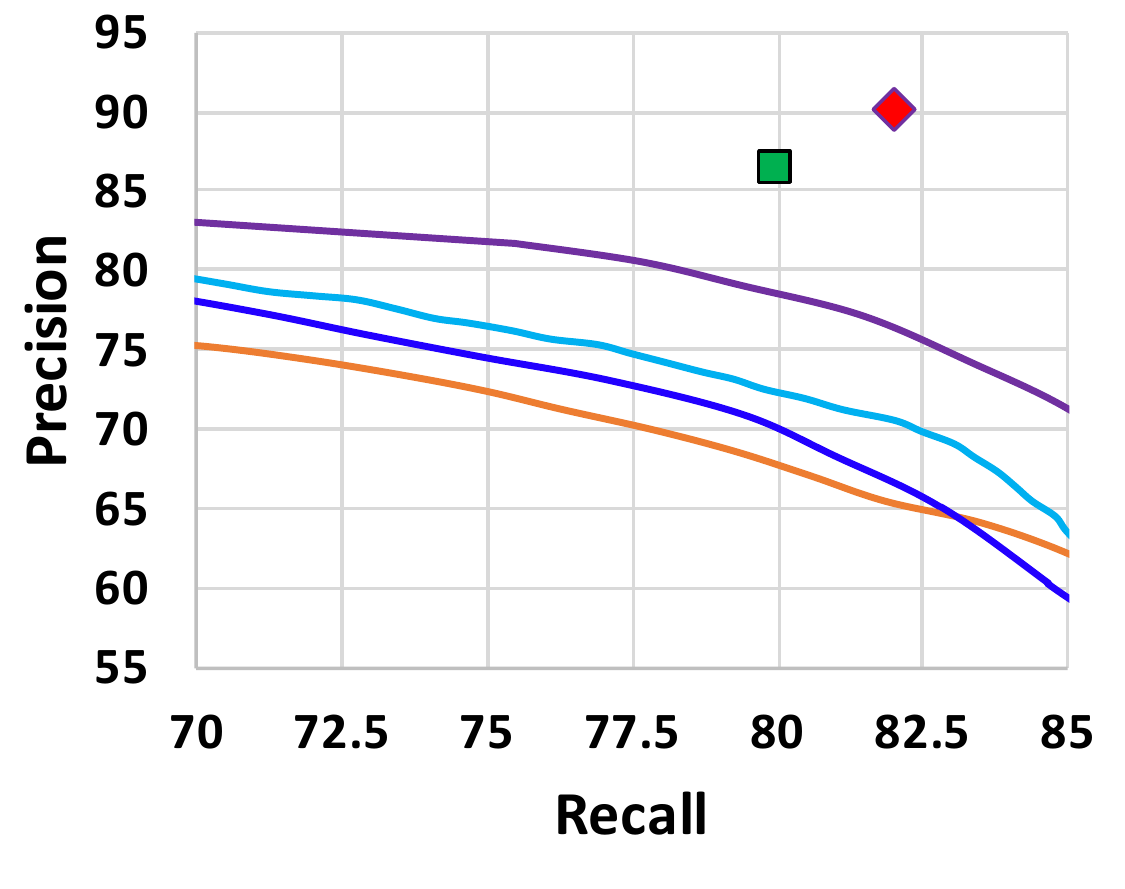}
        \includegraphics[height=.17\textwidth]{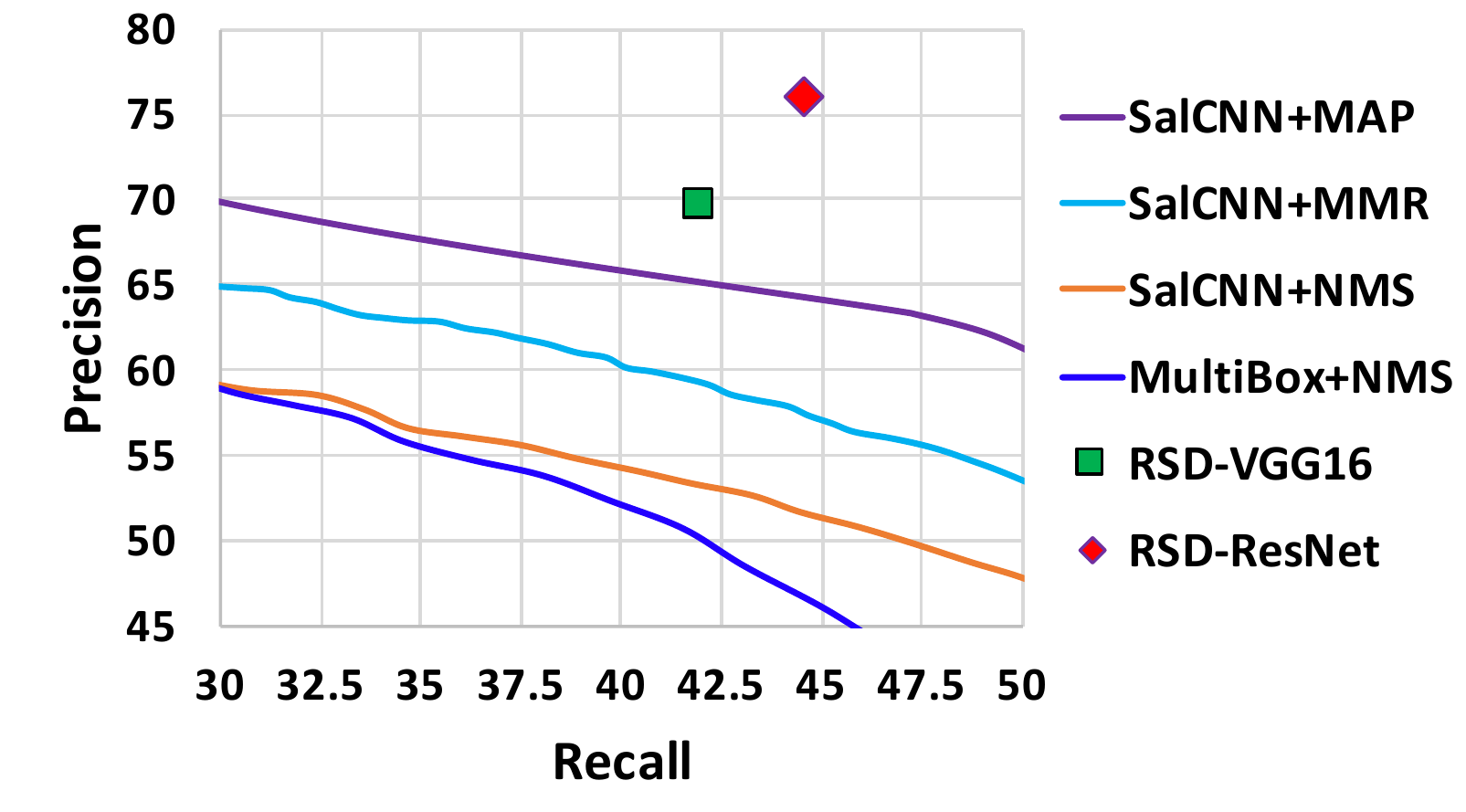}
        \vspace{-2mm}
\caption{{Comparisons of precision/recall by different methods on MSO, PASCAL-S, MSRA, and DUT-O (from left to right) datasets.} 
We let others methods to generate different number of boxes by varying the threshold for confidence scores of boxes and present the performance change as precision-recall curves. The IoU threshold for evaluation is set to $0.5$.}
\label{fig:detection_results}
\end{figure*}

{\flushleft \textbf{Salient Object detection.}} We compare RSD with several existing methods including the state-of-the-art approach in~\cite{zhang2015SOD}, which are SalCNN+MAP, SalCNN+NMS, SalCNN with Maximum Marginal Relevance (SalCNN+MMR), and MultiBox~\cite{DBLP:conf/cvpr/ErhanSTA14} with NMS. 
Unlike our RSD that generates the exact number of bounding boxes as salient objects, other methods have free parameters to determine the number of selected bounding boxes from hundreds of proposals, which greatly changes their performance. 
For fair comparison, we change these free parameters and show their best results with our performance point in Figure~\ref{fig:detection_results}. 
It should be noted that we use the same set of parameters on all datasets, while for other methods different parameters lead to their best performance on different datasets. 

{On the MSO and PASCAL-S datasets that contain multiple salient objects, our RSD-ResNet produces the best results at the same precision or recall rate. RSD-VGG achieves comparable precision/recall as the state-of-the-art methods while being nearly $12\times$ faster. Although our subitizing branch has a range of three, Table~\ref{tab:existance} shows that our RSD-ResNet also achieves the best results on images with $4+$ objects based on the predicted saliency map.} 
On the MSRA and DUT-O datasets that contain single salient object in an image, both of our RSD-VGG and RSD-ResNet outperform the state-of-the-arts by a large margin. Notably, our RSD-ResNet achieves nearly 15\% and 10\% absolute improvement in precision at the same recall rate on the MSRA and DUT-O datasets, respectively, which clearly indicates that our method, without any object proposals, is more powerful and robust even when it is allowed to generate only a single bounding box. 
%
%
\begin{table*}[t]
\renewcommand{\arraystretch}{0.95}
\centering
\footnotesize
\caption{Number of false positives in images containing no salient objects and F1 score for different number of ground-truth objects in the MSO dataset. Results of other methods are obtained at the same recall rate of RSD-ResNet and RSD-VGG, respectively, for fair comparison.}
%
%
\begin{tabular}{ccccc}
\hline
Method                 & RSD-ResNet/RSD-VGG & SalCNN\cite{zhang2015SOD}+MAP & SalCNN+MMR & SalCNN+NMS \\
\hline
\multicolumn{1}{c}{False Positives} & \textbf{36}/\textbf{30}              & 54/40      & 95/50      & 53/34    \\
\multicolumn{1}{c}{F1 Score ($1 \sim 3$ objects)} & \textbf{79.2}/\textbf{77.4}          & 78.9/77.0      & 71.6/72.6      &72.5/70.7 \\
\multicolumn{1}{c}{F1 Score ($4+$ objects)} & \textbf{57.5}/26.8              & 55.2/\textbf{50.9}      & 46.1/47.7      & 47.7/48.5 \\
\hline
\end{tabular}
\label{tab:existance}
\end{table*}
\begin{table}[t]
  \centering
  \footnotesize
  \caption{The accuracy of the counting branch and comparison with the baselines.}
    \begin{tabular}{c|cccc}
    \hline
    {Method} & ResNet~\cite{DBLP:journals/corr/HeZRS15} & RSD-ResNet & VGG~\cite{DBLP:journals/corr/SimonyanZ14a} & RSD-VGG \\
    \hline
    {Accuracy} & 83.33 & \textbf{86.19} & 83.25   & 83.97 \\
    \hline
    \end{tabular}
  \label{tab:counting}
\end{table}

{\flushleft \textbf{Object subitizing.}} We evaluate the subitizing performance of our RSD on the MSO dataset. First, we compare our RSD with state-of-the-art methods in terms of solving the \emph{existence} problem in Table~\ref{tab:existance}.
While our parameters are fixed, we vary the parameters of other methods on different datasets to match their performance. For example, we tune the parameters of other methods when comparing with our RSD-ResNet, so that they can achieve the same recall as ours. Then we compare the number of false positives in the background images. We do the same thing for the comparison with our RSD-VGG as well. 

For predicting \emph{existence}, both our RSD-ResNet and RSD-VGG produce fewer false positives when there is no salient object. 
Additionally, we compare the counting performance of RSD with two baselines using vanilla ResNet-50 and VGG16 in Table~\ref{tab:counting}. For fair comparison, we use exactly the same training scheme and initialization for all networks. 
Our RSD method successfully produces better accuracy compared with vanilla ResNet-50 and VGG16, verifying that the multi-task training facilitates the subitizing branch to learn a better classifier by utilizing the information from saliency map prediction. 
%
\ignore{
\begin{figure}[t]
    \centering
        \includegraphics[width=.42\linewidth]{Figs/confusion_ours_resnet.pdf}
        \includegraphics[width=.42\linewidth]{Figs/confusion_ours_vgg.pdf}
 	\caption{The confusion matrix of the results by our subitizing branch of RSD-ResNet (left) and RSD-VGG (right).}
    \label{fig:confusion}
\end{figure}
}

{\flushleft \textbf{Saliency map prediction.}} {In real world scenarios, pixel-level annotations are difficult to collect. It is challenging to generate precise saliency maps without such detailed labeling.} As a weakly-supervised approach only using bounding boxes for salient foreground segmentation, we will show that our RSD still generates accurate saliency map that aligns well with multiple salient objects in the scene. We compare our RSD against five powerful unsupervised salient object
segmentation algorithms, RC~\cite{rc}, SF~\cite{perazzi2012saliency}, GMR~\cite{yang2013saliency}, 
PCAS~\cite{zhang2008sun},
GBVS~\cite{harel2006graph} and three state-of-the-art supervised methods,
HDCT~\cite{kim2014salient},
DRFI~\cite{jiang2013salient},
GBVS+PatchCut~\cite{yang2015patchcut}.We also evaluate the performance using precision-recall curves. Specifically, the precision and recall are computed by binarizing the grayscale saliency map using varying thresholds~\cite{achanta2009frequency,perazzi2012saliency,yang2013saliency,hs} and comparing the binary mask against the ground-truth.
Our RSD approach is surprisingly good considering that it only uses rough Gaussian maps as ground-truth. In particular, the RSD-ResNet approach produces comparable results with the fully-supervised methods in terms of precision/recall, making it readily applicable for salient foreground segmentation without any pixel-level annotations.
\begin{figure}
    \centering
    \includegraphics[height=2.8cm,width=.49\linewidth]{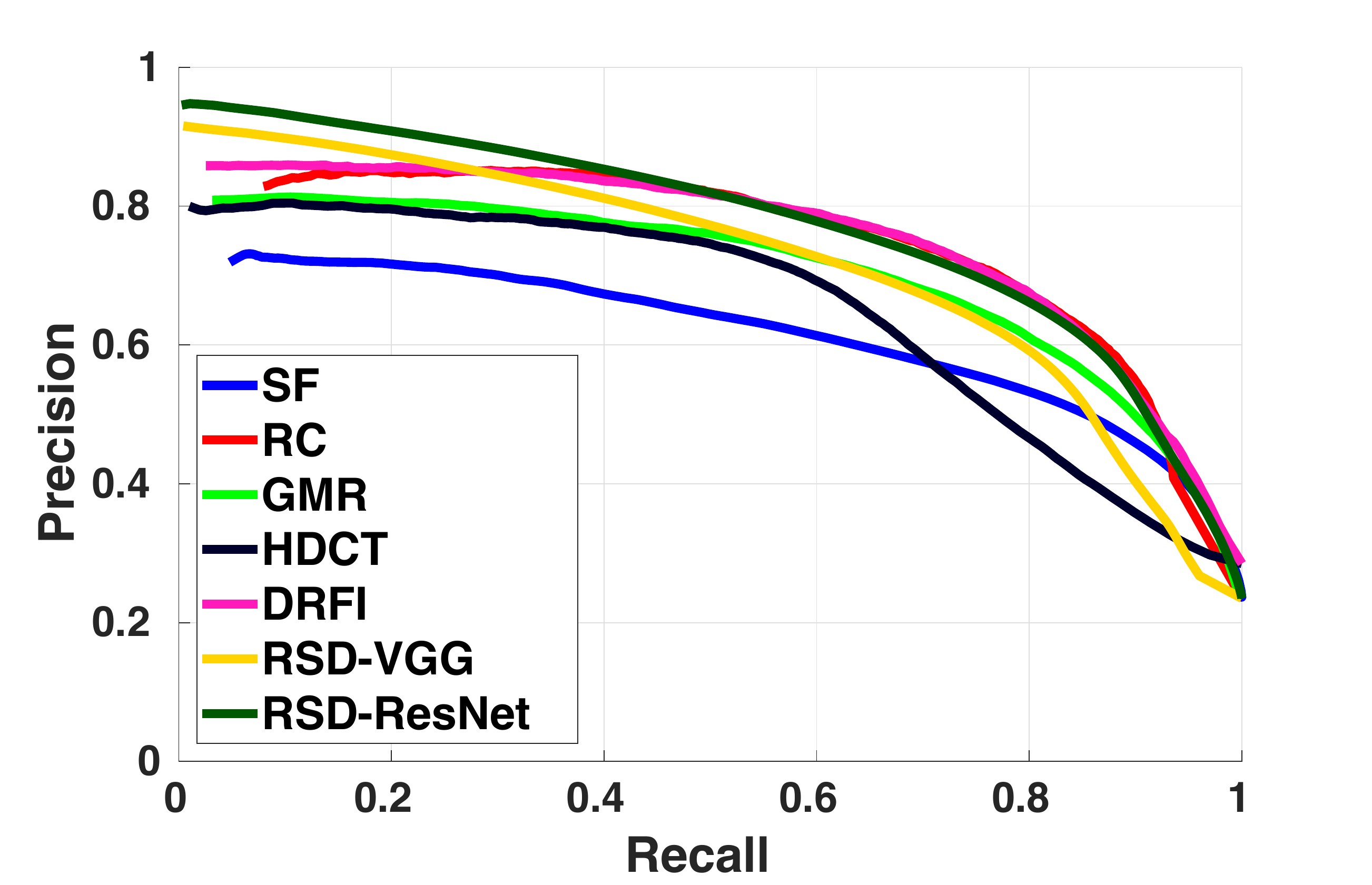}
    \includegraphics[height=2.8cm,width=.49\linewidth]{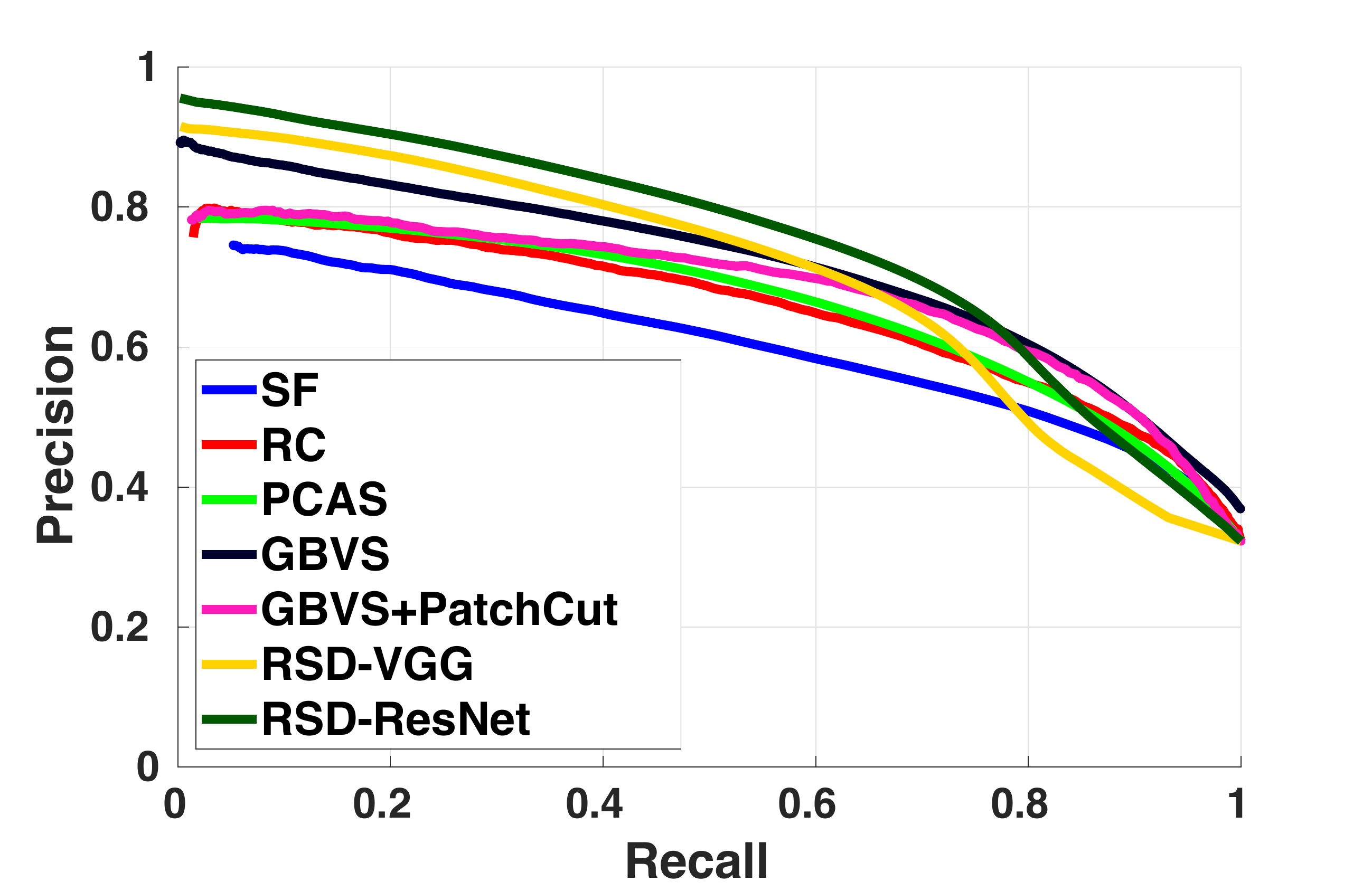}
 	\caption{The pixel-wise saliency map prediction performance on the ESSD~\cite{hs} (left) and PASCAL-S~\cite{li2014secrets} (right) datasets.}
    \label{fig:saliency_map}
\end{figure}

\subsection{Ablation study}
\begin{figure*}[t]
    \centering
    \begin{subfigure}[b]{0.12\textwidth}
        \includegraphics[width=\textwidth]{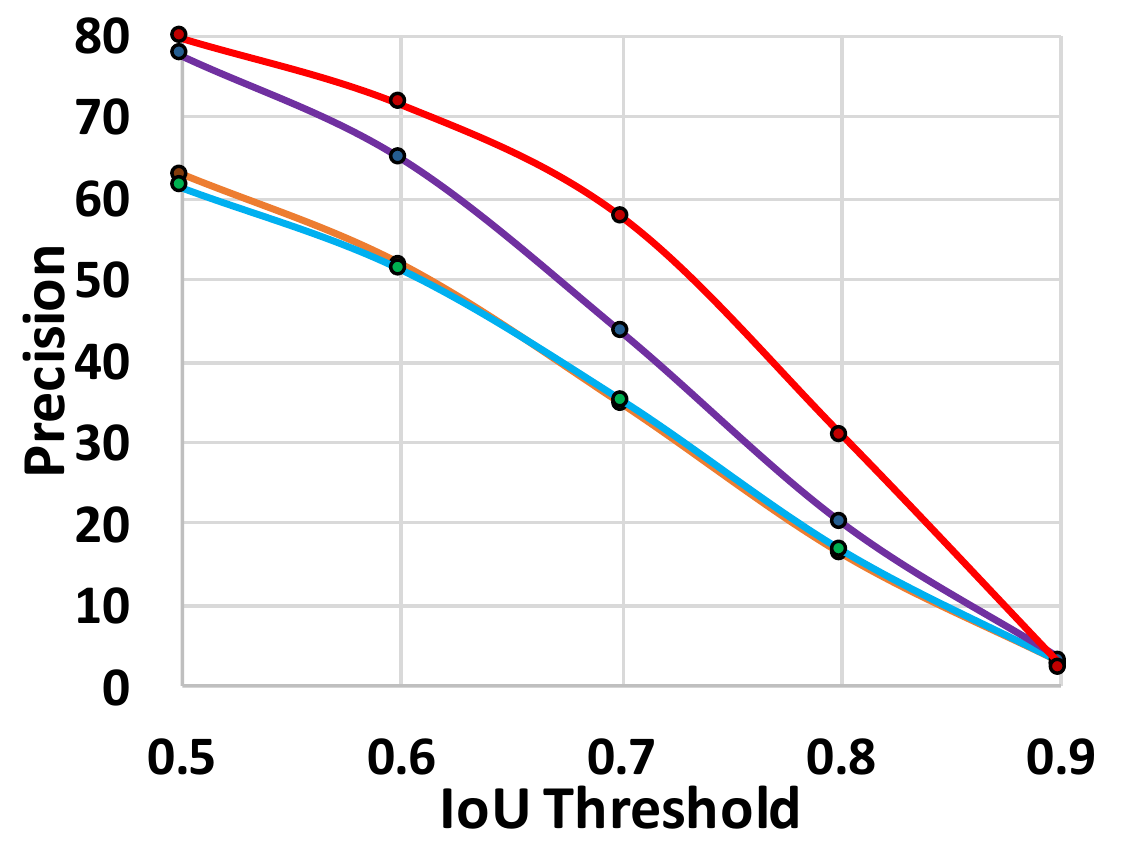}
        \caption{{\scriptsize MSO-P}}
    \end{subfigure}
    \hspace{-0.2cm}
    \begin{subfigure}[b]{0.12\textwidth}
        \includegraphics[width=\textwidth]{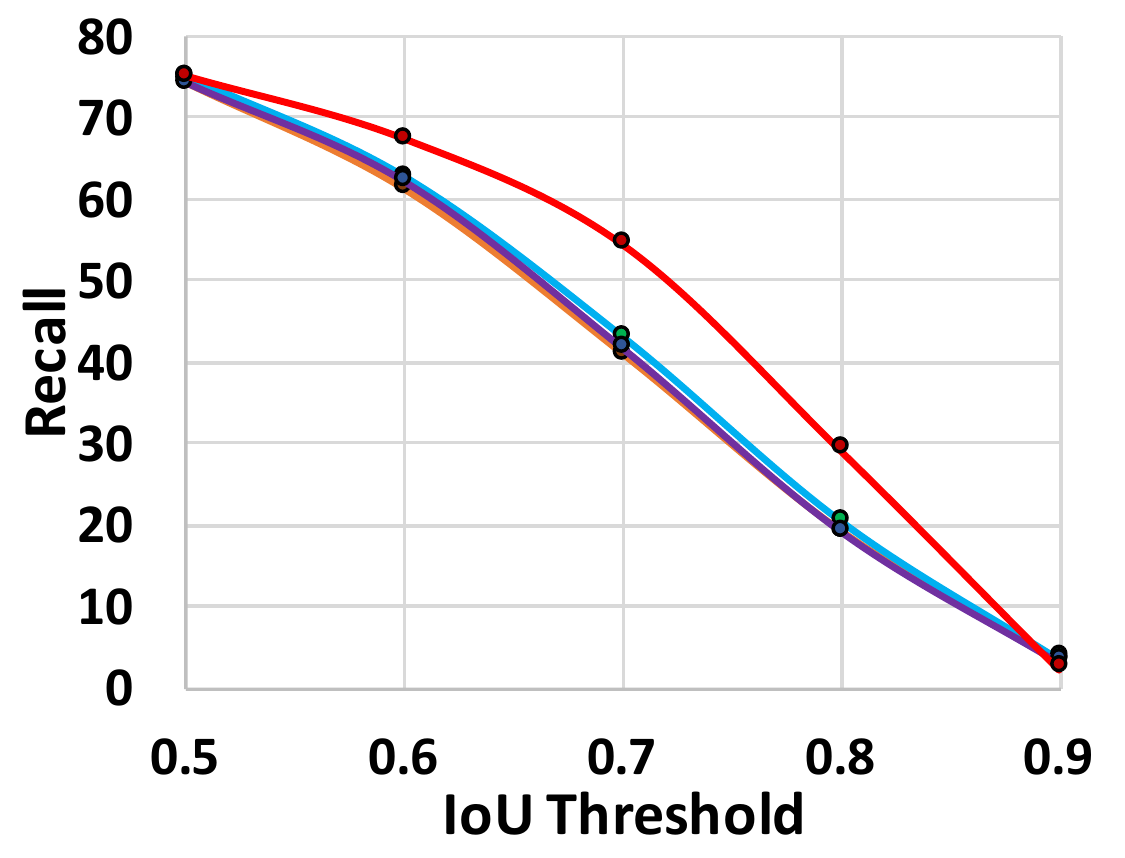}
        \caption{{\scriptsize MSO-R}}
    \end{subfigure}
    \hspace{-0.2cm}
    \begin{subfigure}[b]{0.12\textwidth}
        \includegraphics[width=\textwidth]{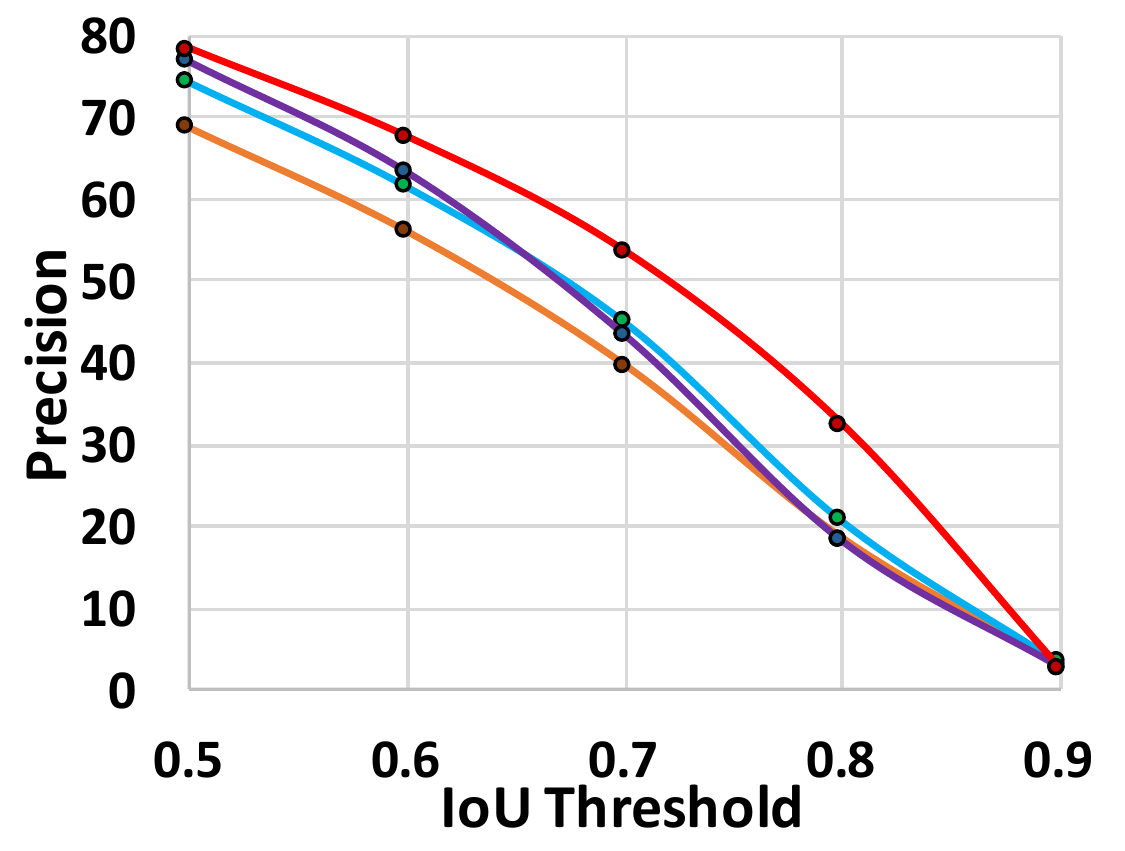}
        \caption{{\scriptsize PASCAL-S-P}}
    \end{subfigure}
        \hspace{-0.2cm}
    \begin{subfigure}[b]{0.12\textwidth}
        \includegraphics[width=\textwidth,trim={0 0 4.5cm 0},clip]{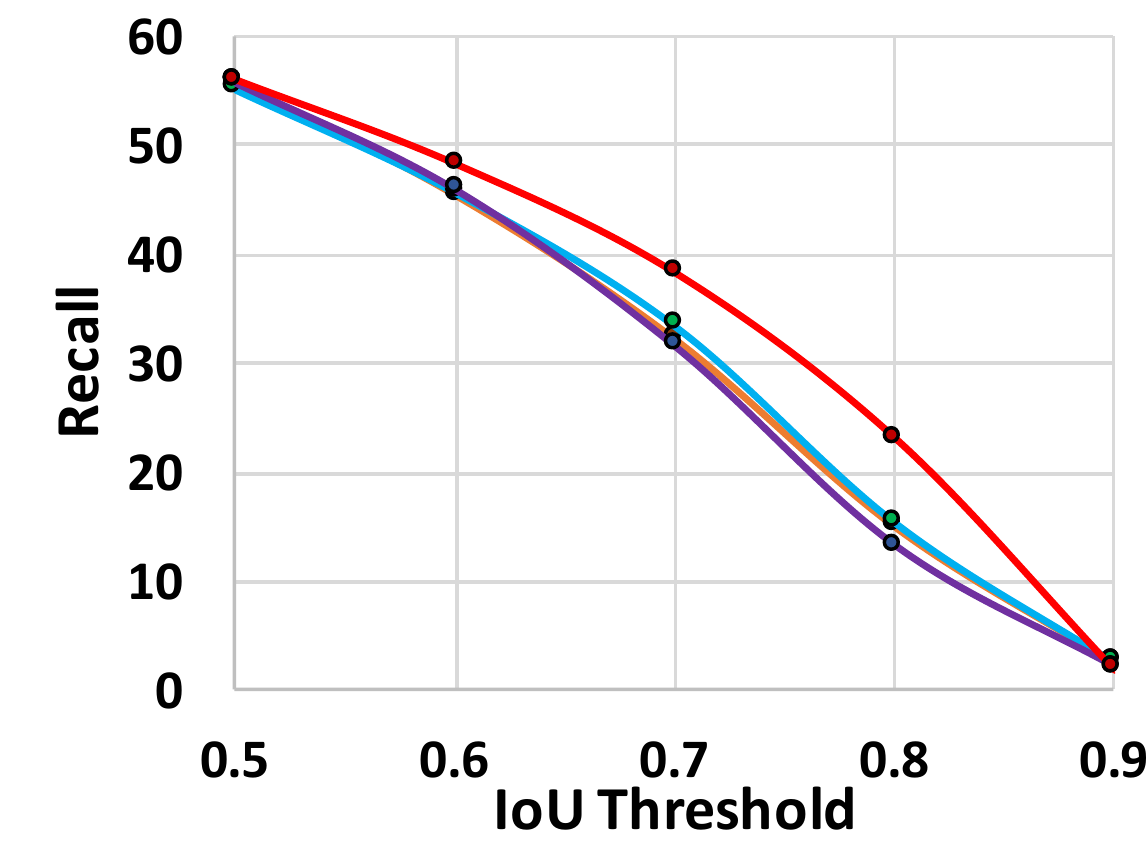}
        \caption{{\scriptsize PASCAL-S-R}}
    \end{subfigure}    
    \hspace{-0.2cm}
    \begin{subfigure}[b]{0.12\textwidth}
        \includegraphics[width=\textwidth]{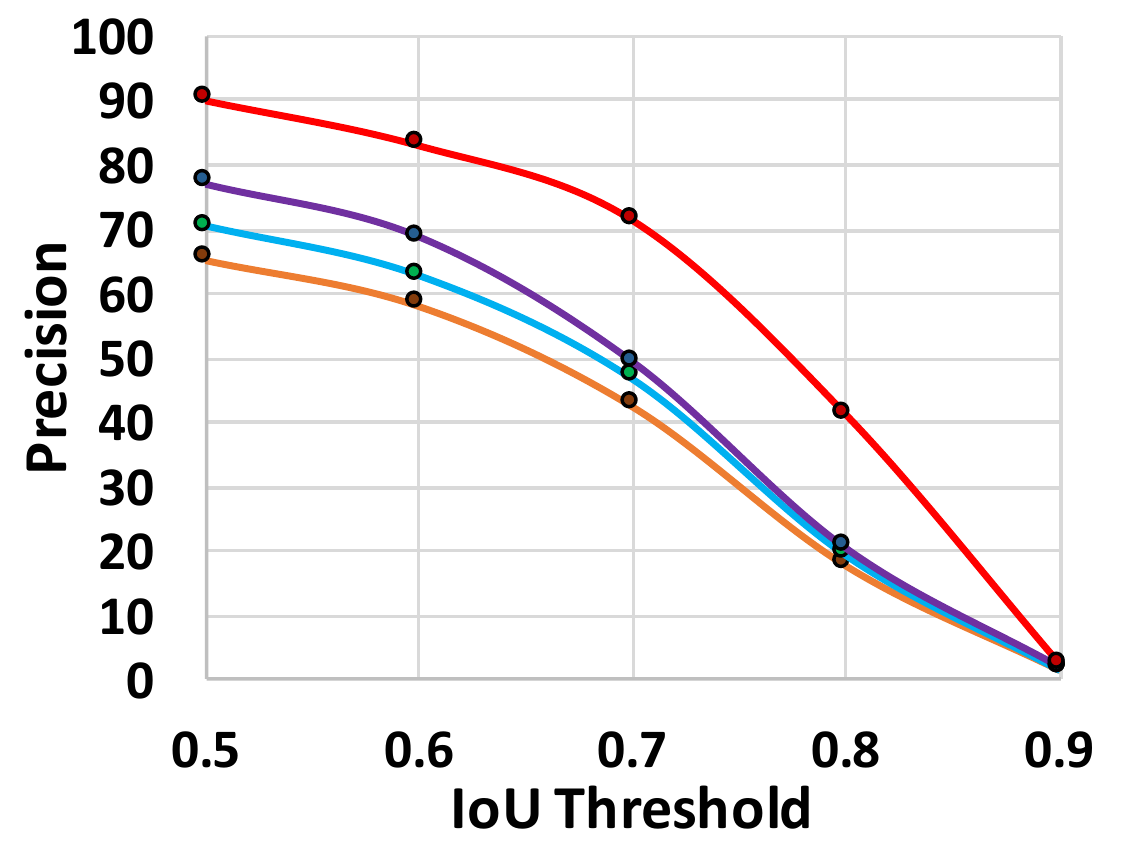}
        \caption{{\scriptsize MSRA-P}}
    \end{subfigure}
        \hspace{-0.2cm}
        \begin{subfigure}[b]{0.12\textwidth}
        \includegraphics[width=\textwidth]{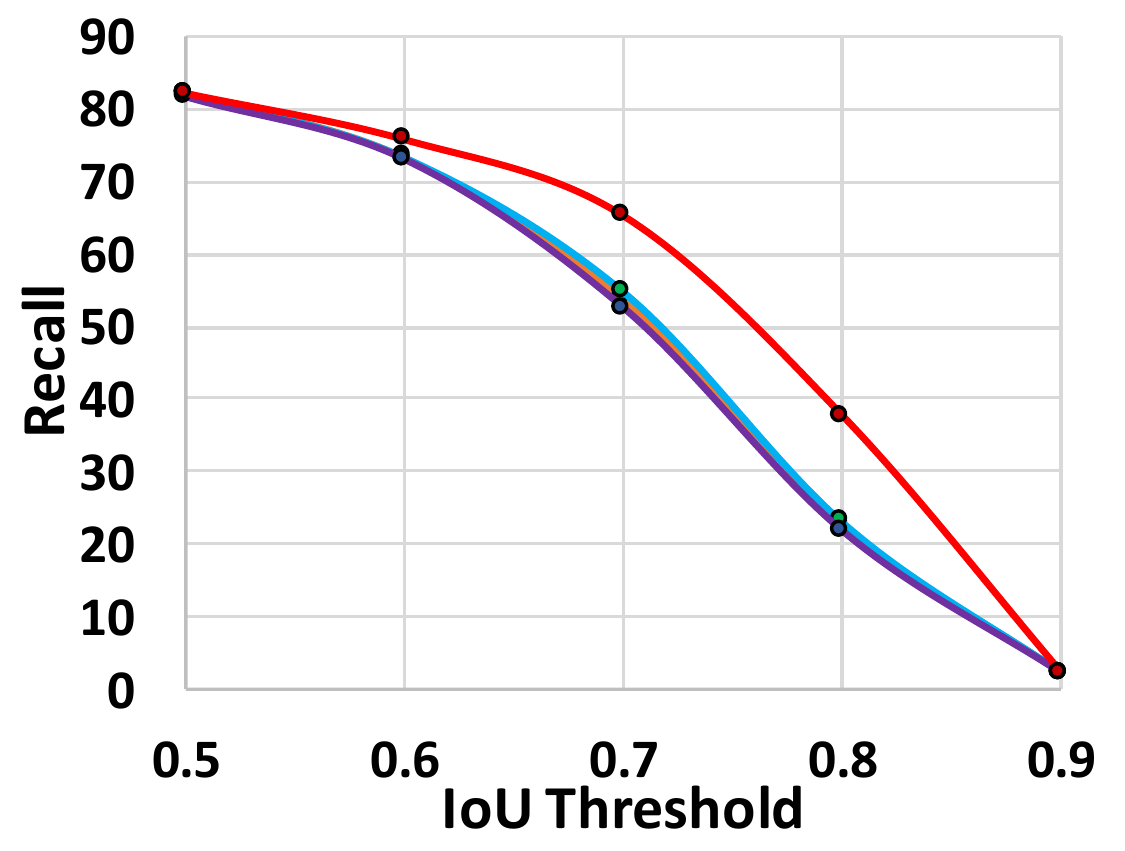}
        \caption{{\scriptsize MSRA-R}}
    \end{subfigure}
        \hspace{-0.2cm}
        \begin{subfigure}[b]{0.12\textwidth}
        \includegraphics[width=\textwidth]{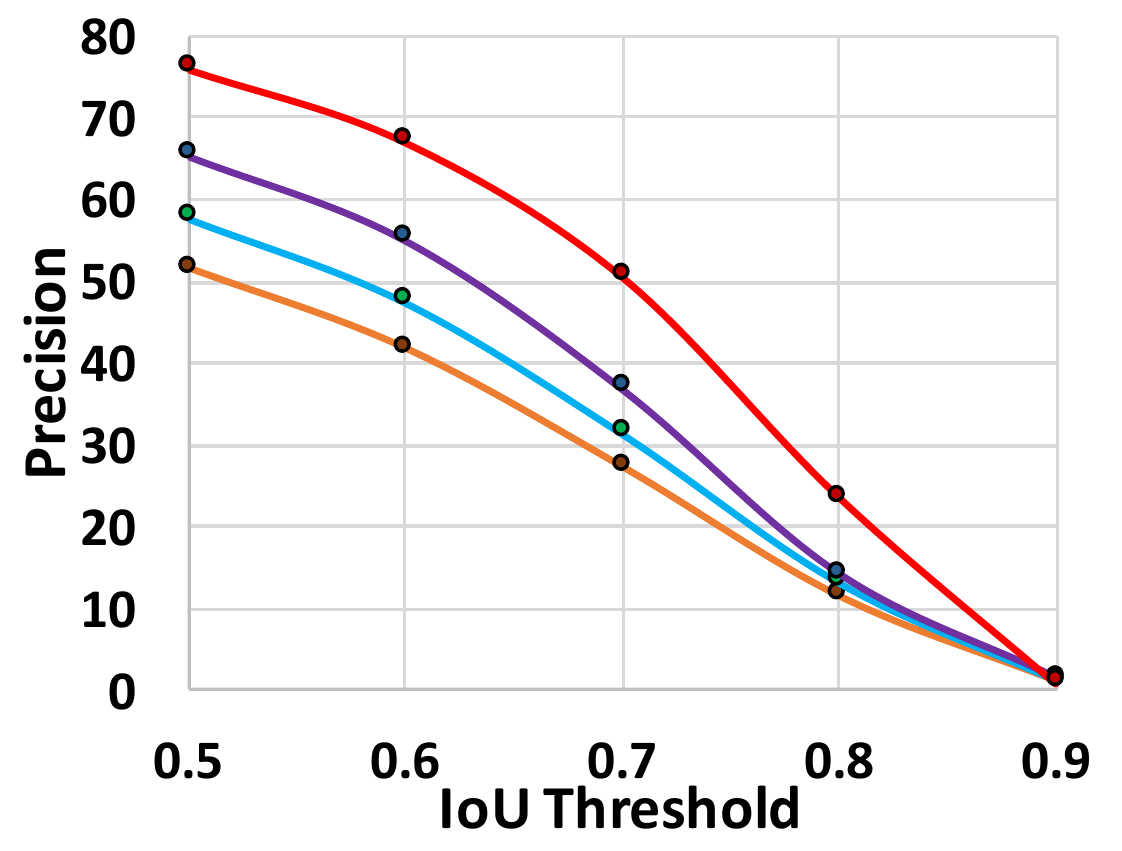}
        \caption{{\scriptsize DUT-O-P}}
    \end{subfigure}
        \hspace{-0.25cm}
        \begin{subfigure}[b]{0.17\textwidth}
        \includegraphics[width=\textwidth]{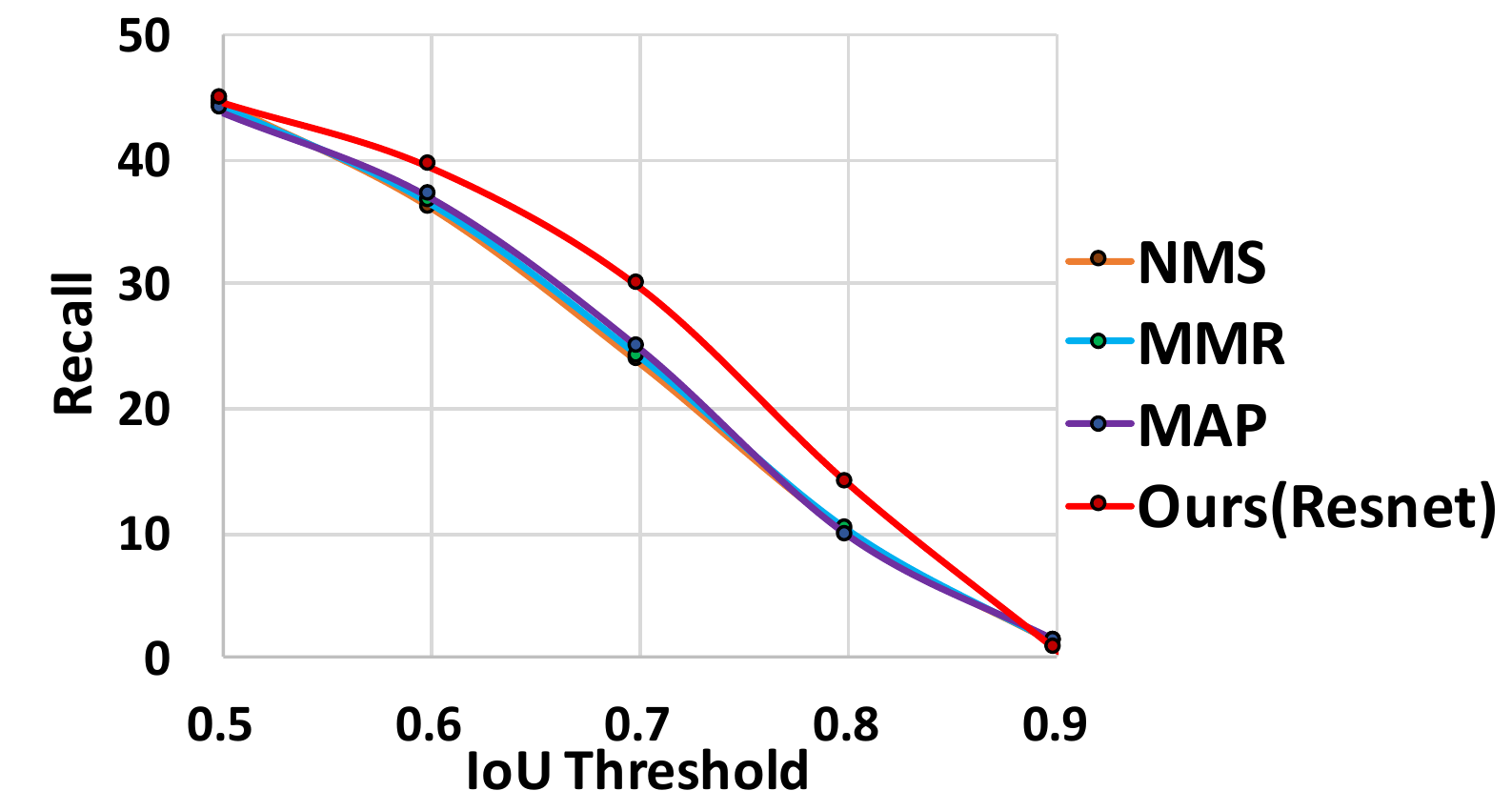}
        \caption{{\scriptsize DUT-O-R}}
    \end{subfigure}
    \hspace{-0.2cm}
    \begin{subfigure}[b]{0.12\textwidth}
        \includegraphics[width=\textwidth]{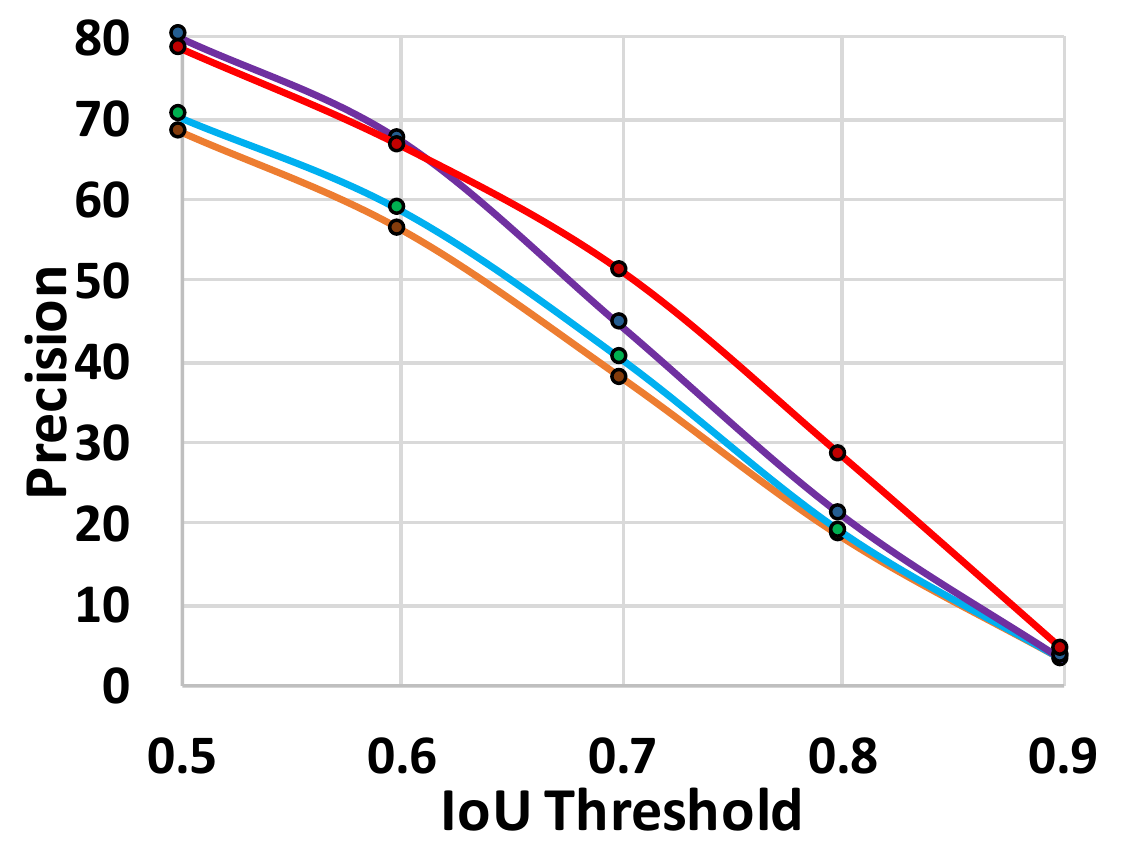}
        \caption{{\scriptsize MSO-P}}
        \label{fig:mso_res}
    \end{subfigure}
            \hspace{-0.2cm}
    \begin{subfigure}[b]{0.12\textwidth}
        \includegraphics[width=\textwidth]{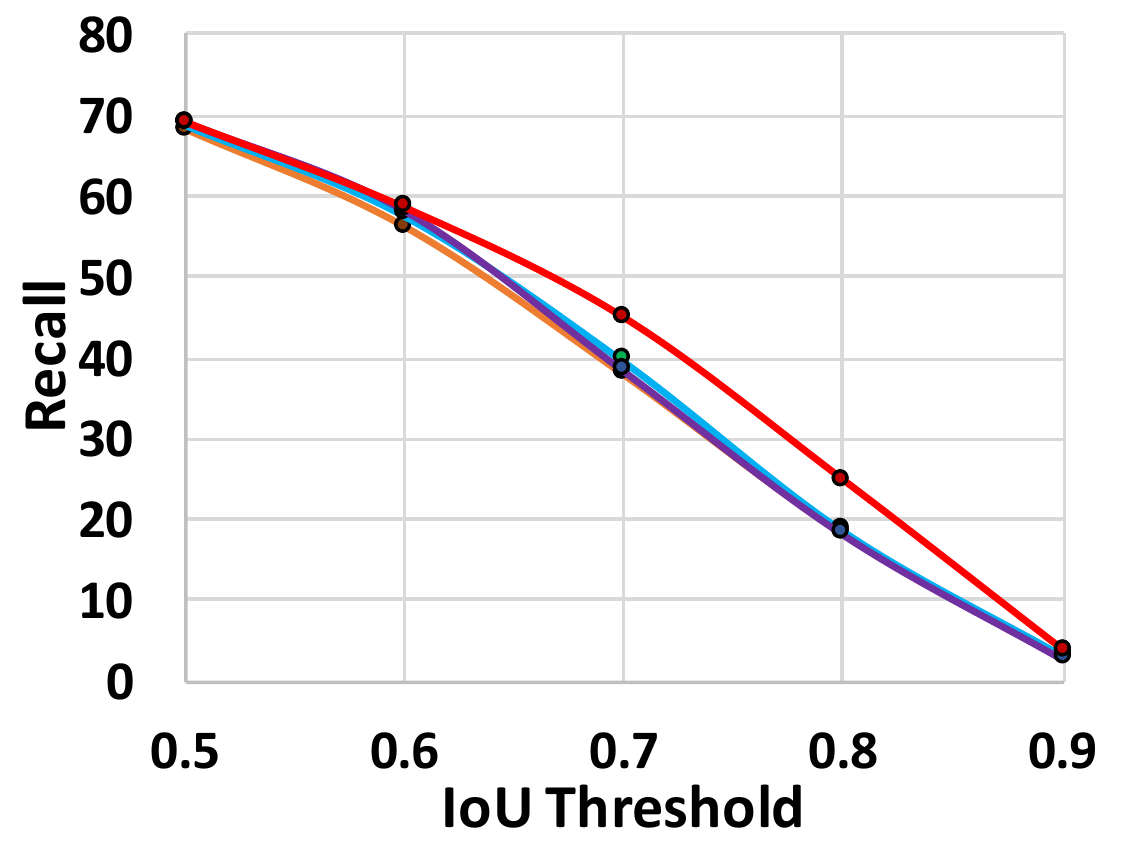}
        \caption{{\scriptsize MSO-R}}
        \label{fig:MSRA_res}
    \end{subfigure}
            \hspace{-0.2cm}
     \begin{subfigure}[b]{0.12\textwidth}
        \includegraphics[width=\textwidth]{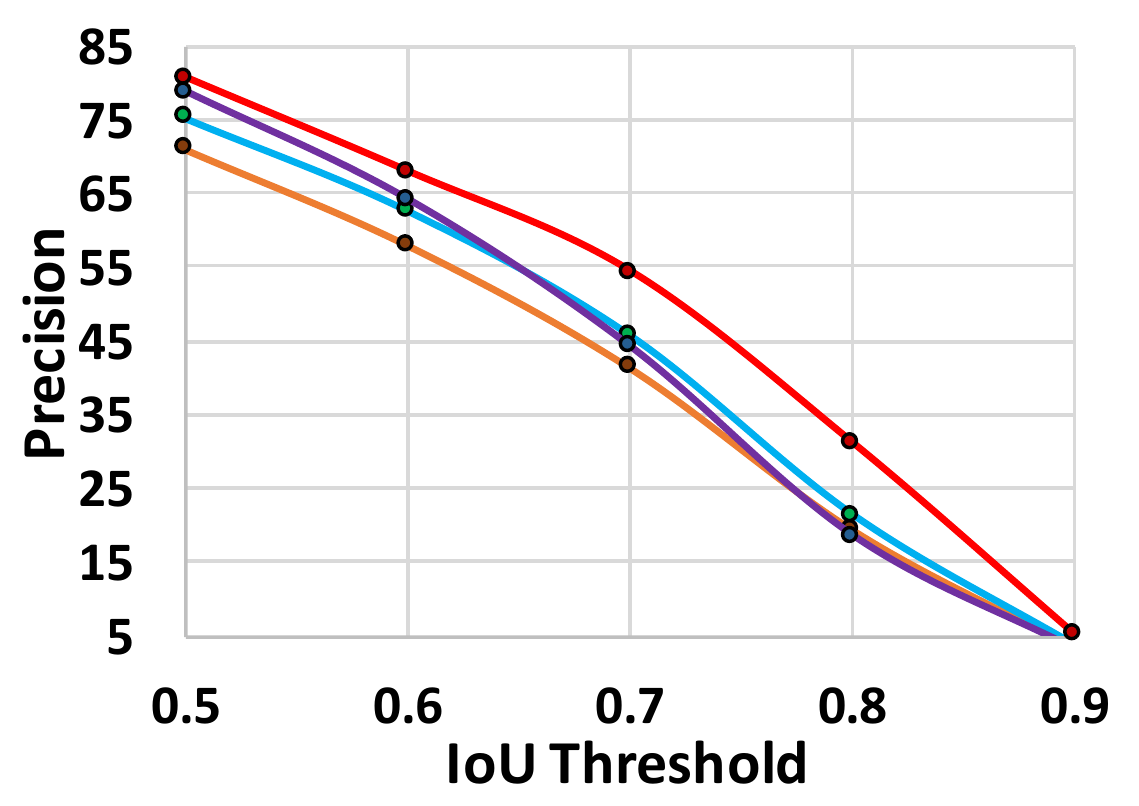}
        \caption{{\scriptsize PASCAL-S-P}}
    \end{subfigure}
        \hspace{-0.2cm}
    \begin{subfigure}[b]{0.12\textwidth}
        \includegraphics[width=\textwidth,trim={0 0 4.5cm 0},clip]{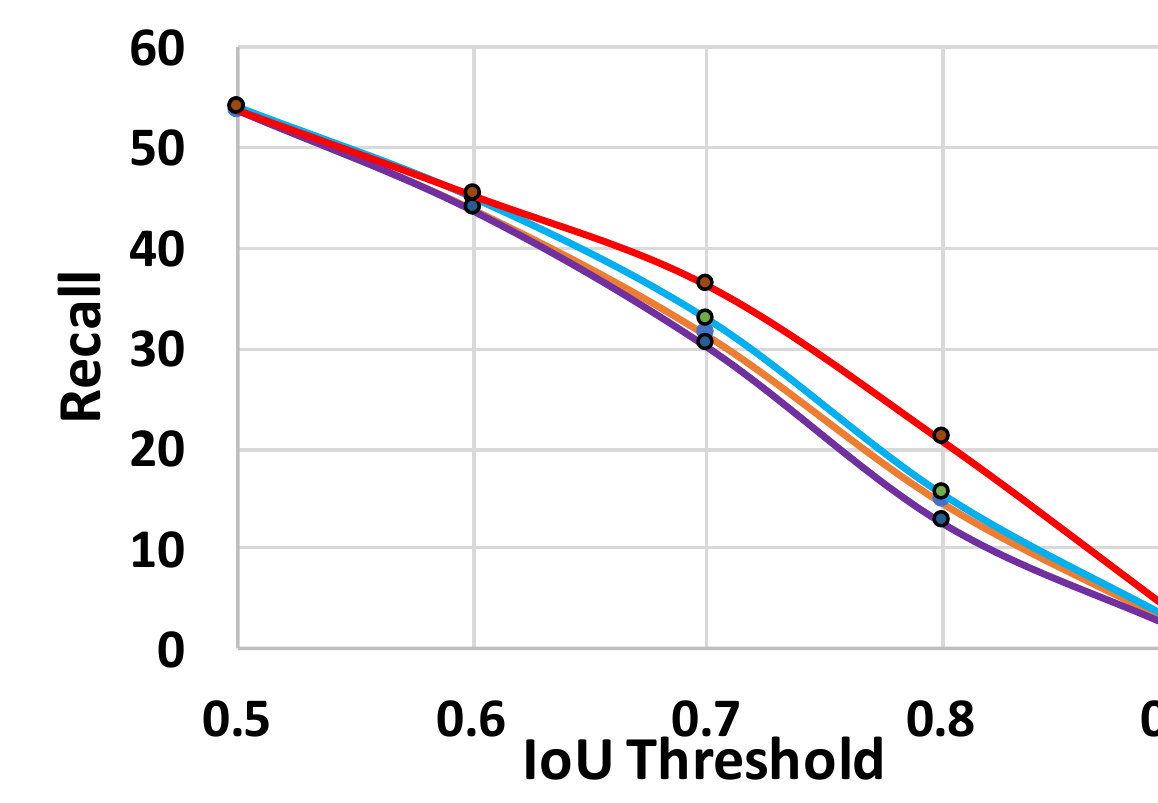}
        \caption{{\scriptsize PASCAL-S-R}}
    \end{subfigure}
     \hspace{-0.2cm}
    \begin{subfigure}[b]{0.12\textwidth}
        \includegraphics[width=\textwidth]{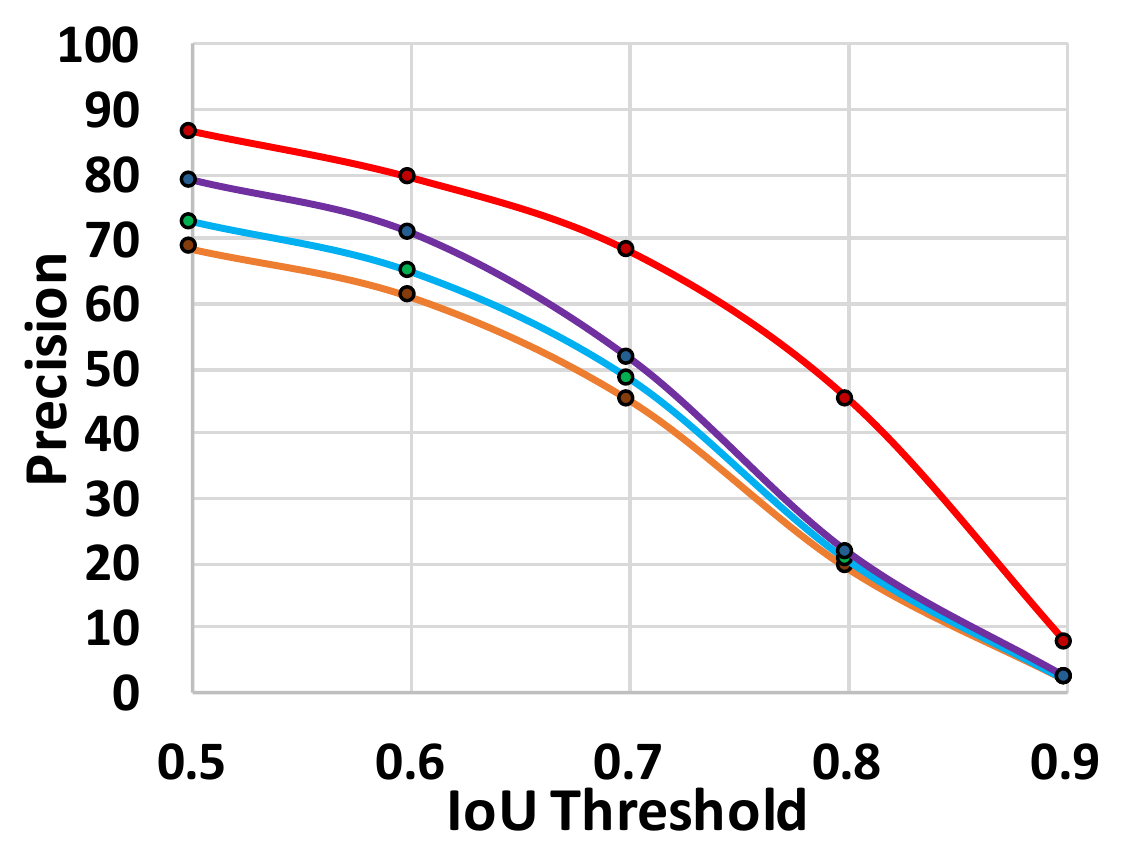}
        \caption{{\scriptsize MSRA-P}}
    \end{subfigure}
            \hspace{-0.2cm}
        \begin{subfigure}[b]{0.12\textwidth}
        \includegraphics[width=\textwidth]{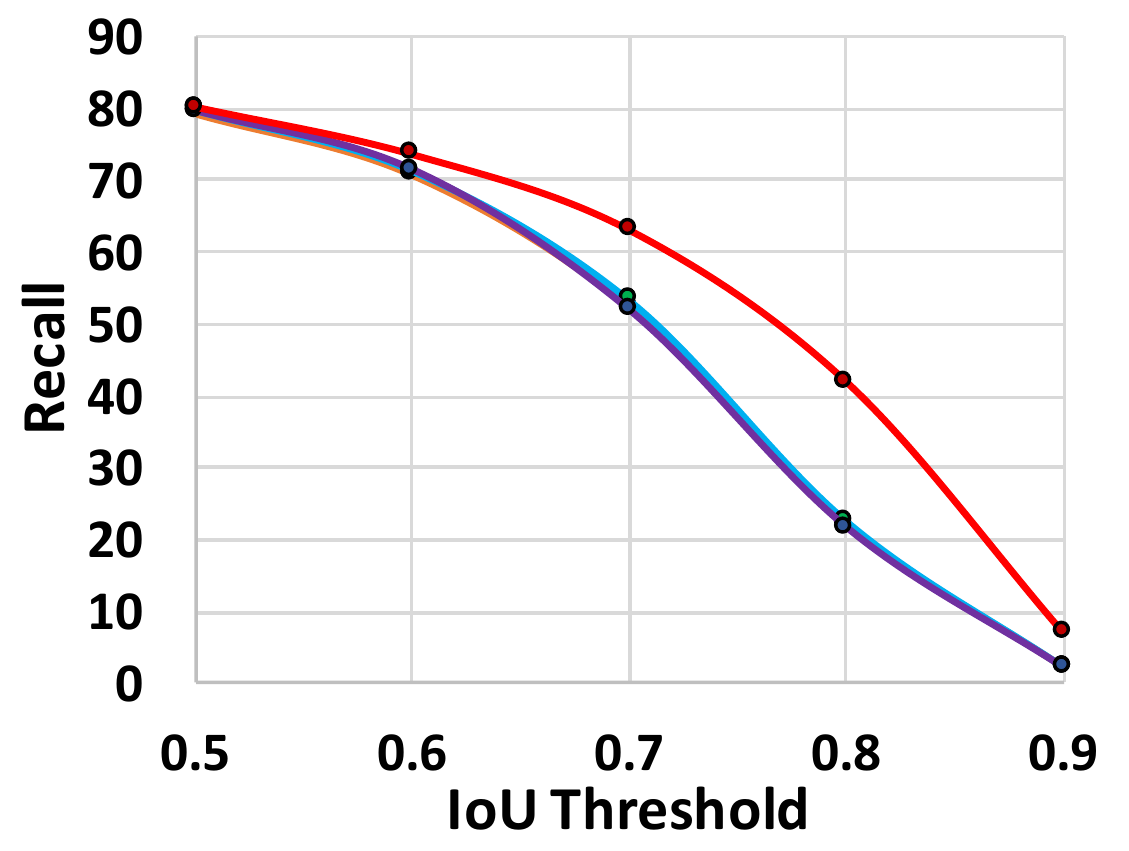}
        \caption{{\scriptsize MSRA-R}}
    \end{subfigure}
            \hspace{-0.2cm}
        \begin{subfigure}[b]{0.12\textwidth}
        \includegraphics[width=\textwidth]{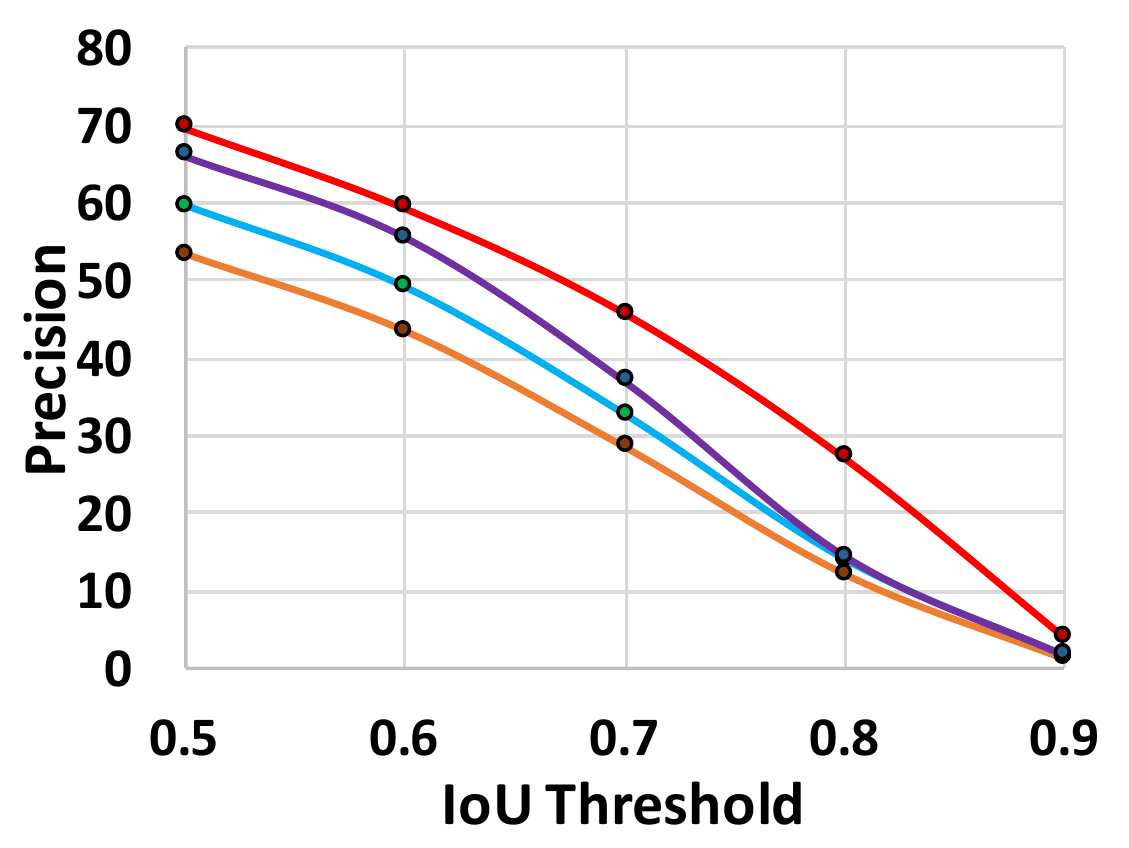}
        \caption{{\scriptsize DUT-O-P}}

    \end{subfigure}
            \hspace{-0.2cm}
        \begin{subfigure}[b]{0.17\textwidth}
        \includegraphics[width=\textwidth]{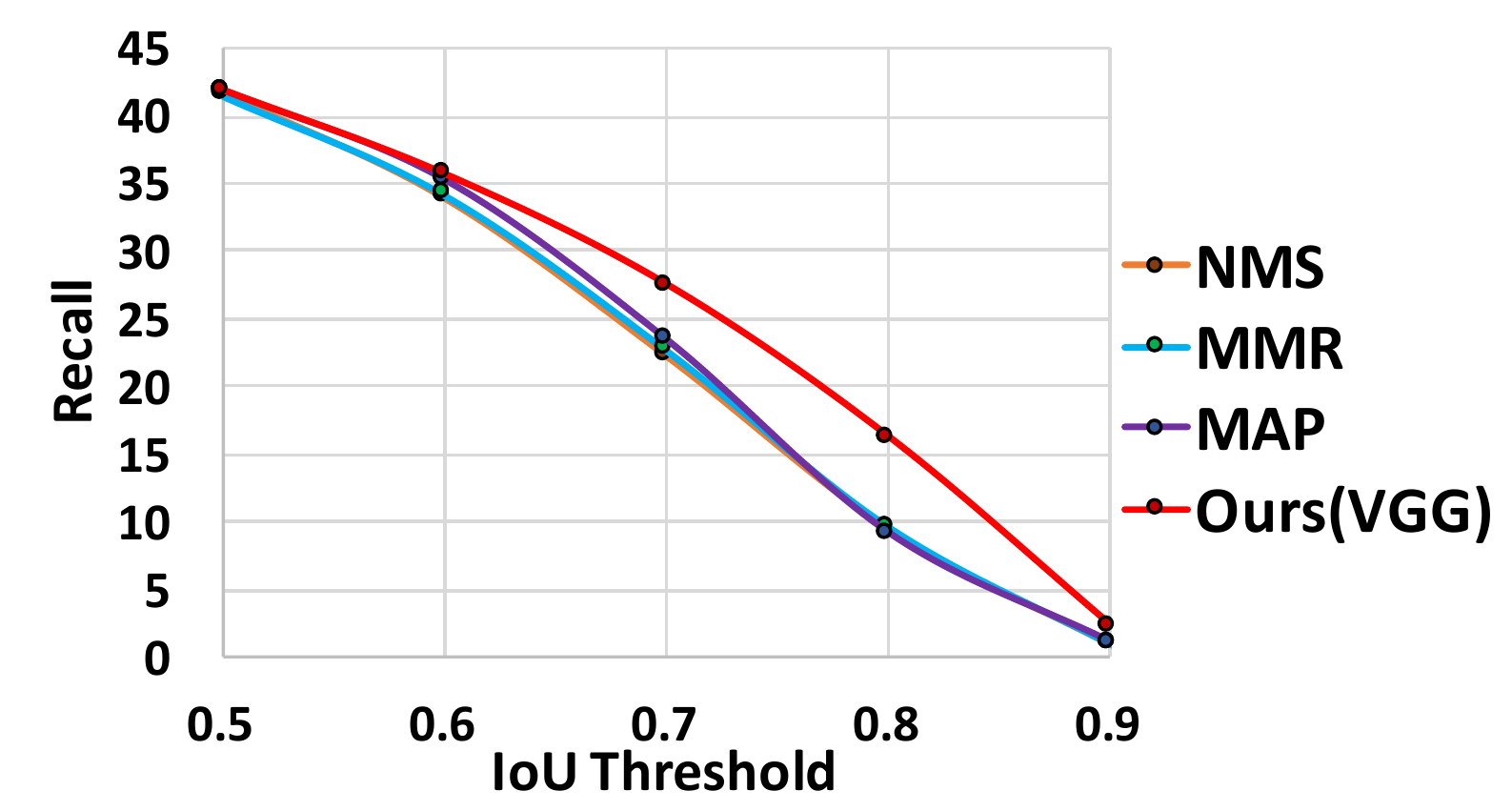}
        \caption{{\scriptsize DUT-O-R}}
    \end{subfigure}
    \vspace{-2mm}
    \caption{Comparison of RSD-ResNet (top), RSD-VGG (bottom) with other methods in terms of precision at the same recall and recall at the same precision under different localization thresholds. {``P'' stands for precision and ``R'' stands for recall.}}
    \label{fig:iou}
\end{figure*}

\begin{figure*}
\begin{tabular}{c>{\columncolor{green!15}}c>{\columncolor{green!15}}c>{\columncolor{green!15}}c>{\columncolor{green!15}}c>{\columncolor{green!15}}c>{\columncolor{green!15}}c>{\columncolor{red!15}}c>{\columncolor{red!15}}c}
\rotatebox[origin=c]{90}{{\footnotesize \textbf{GT}}}& \hspace{-5mm}
\begin{minipage}{0.1\linewidth}\includegraphics[trim={4.5cm 2cm 4cm 2cm},clip,width=\linewidth]{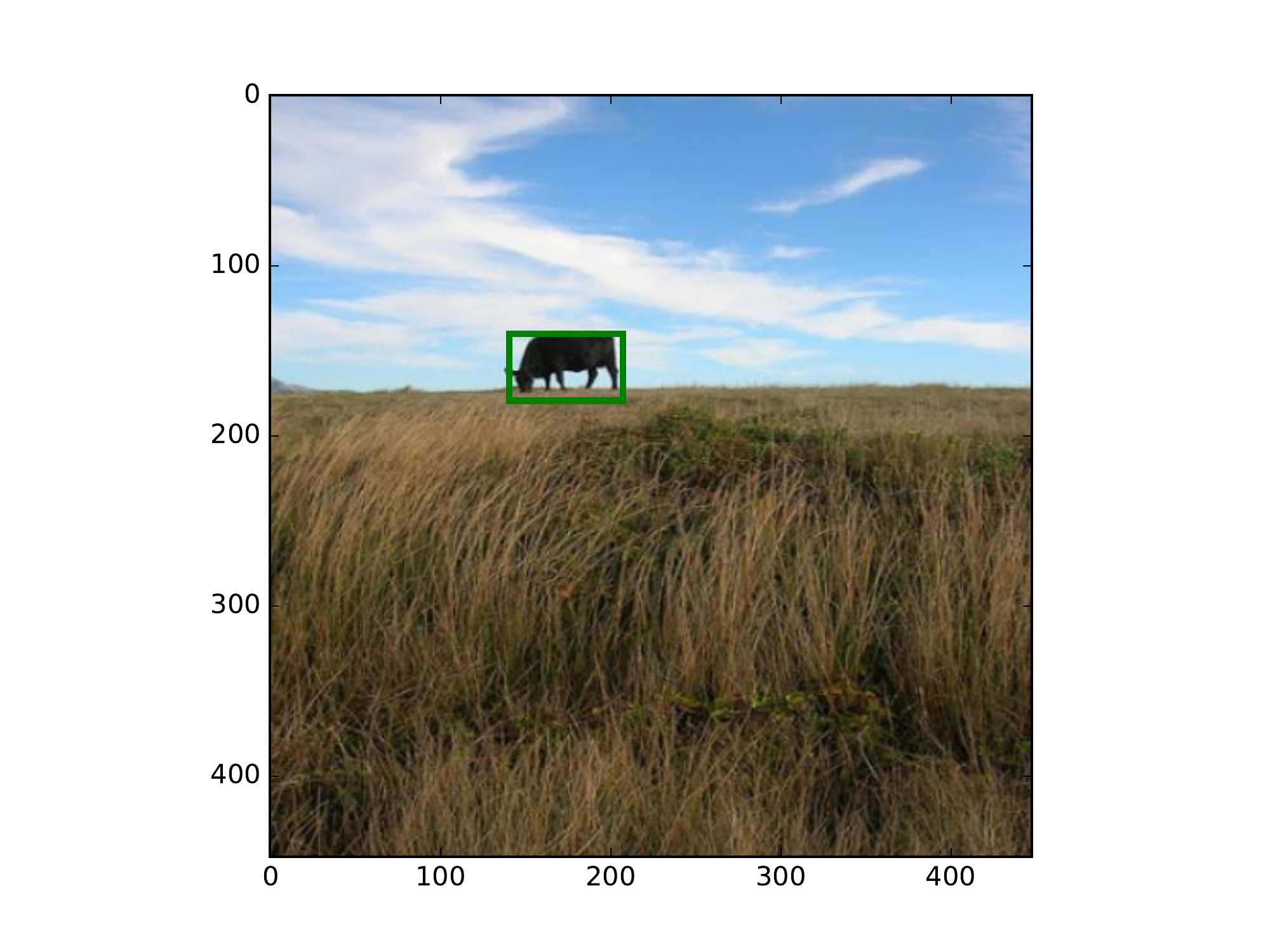}\end{minipage}&
\begin{minipage}{0.1\linewidth}\includegraphics[trim={4.5cm 2cm 4cm 2cm},clip,width=\linewidth]{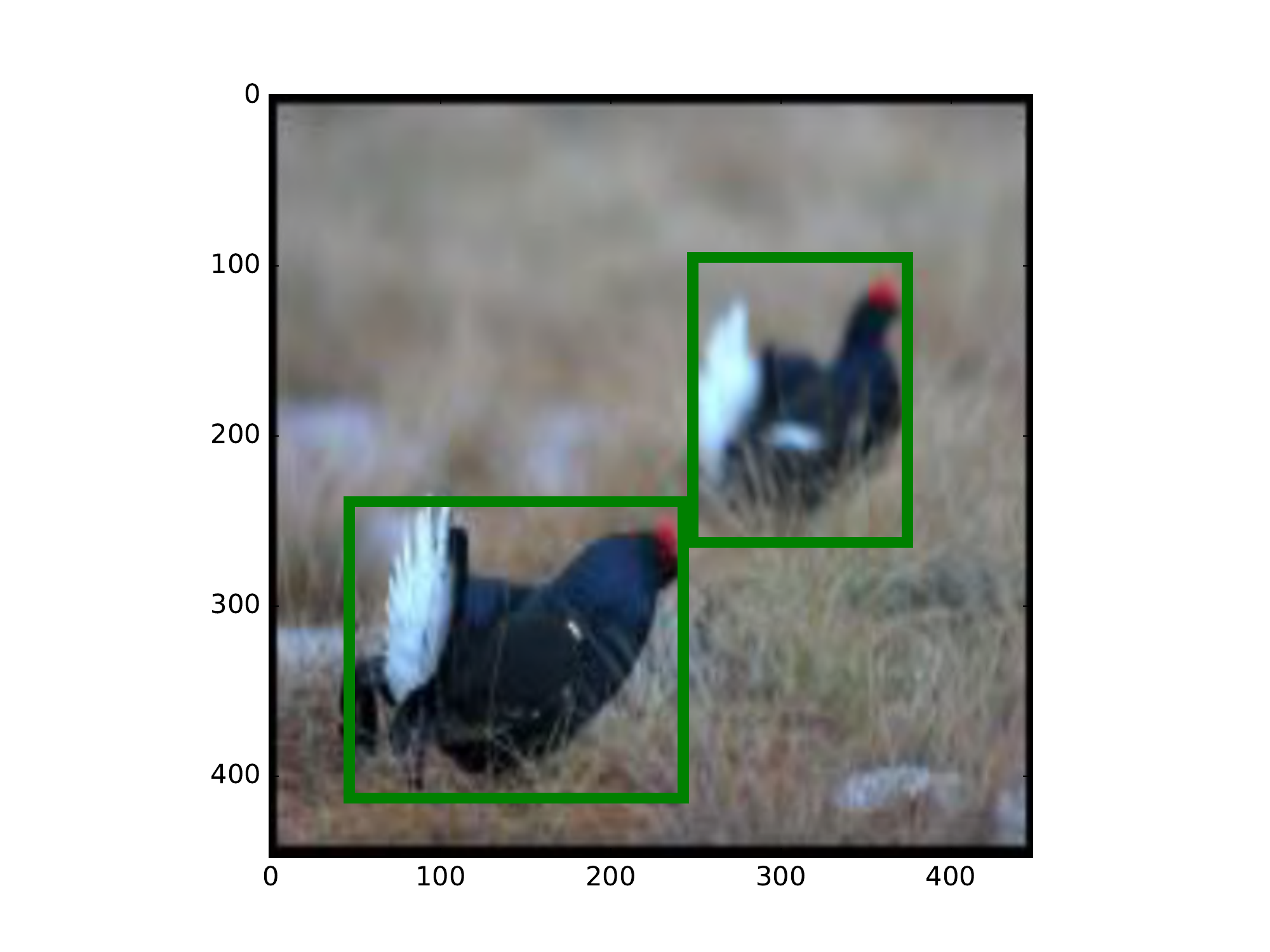}\end{minipage}&
\begin{minipage}{0.1\linewidth}\includegraphics[trim={4.5cm 2cm 4cm 2cm},clip,width=\linewidth]{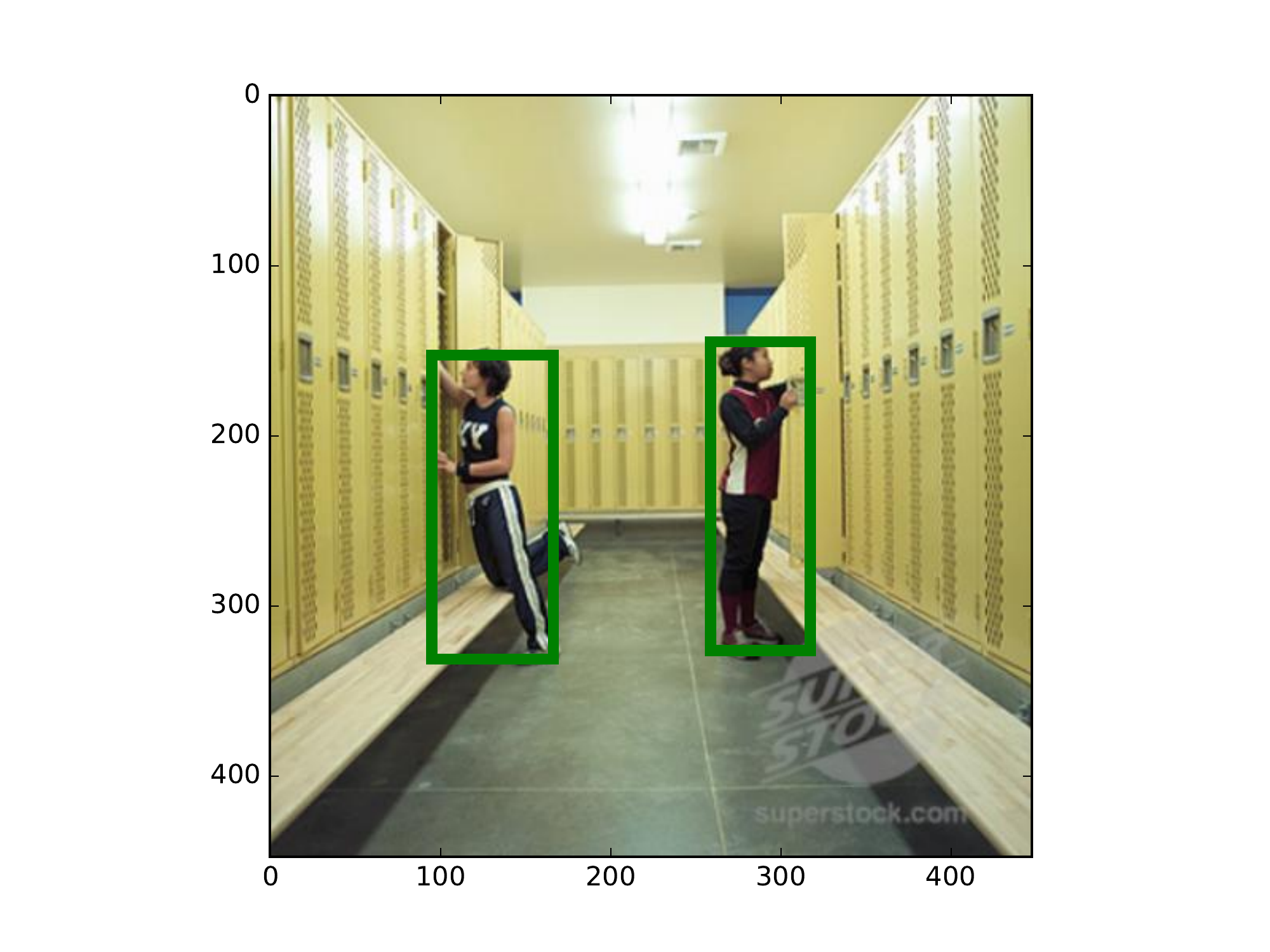}\end{minipage}&
\begin{minipage}{0.1\linewidth}\includegraphics[trim={4.5cm 2cm 4cm 2cm},clip,width=\linewidth]{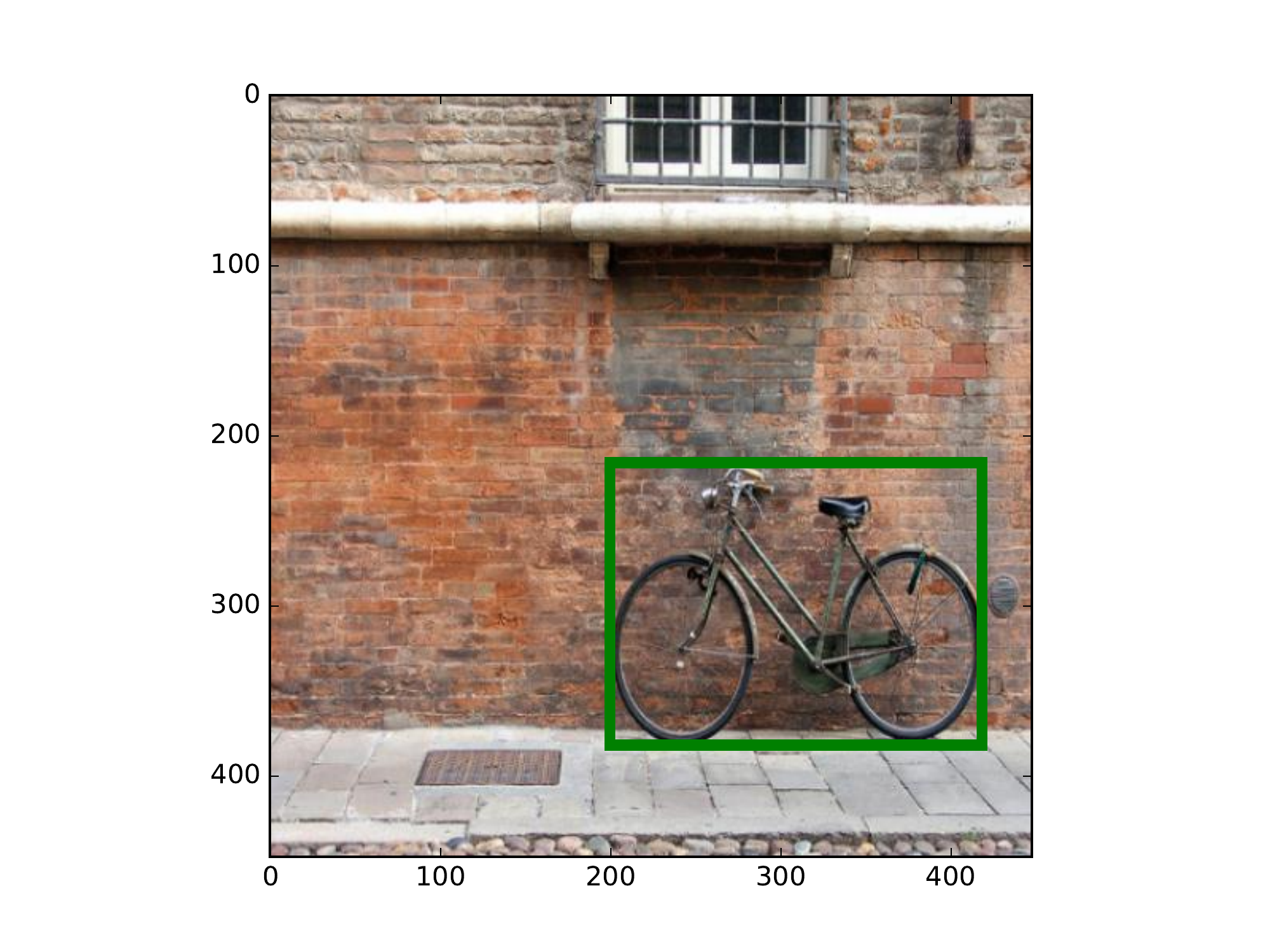}\end{minipage}&
\begin{minipage}{0.1\linewidth}\includegraphics[trim={4.5cm 2cm 4cm 2cm},clip,width=\linewidth]{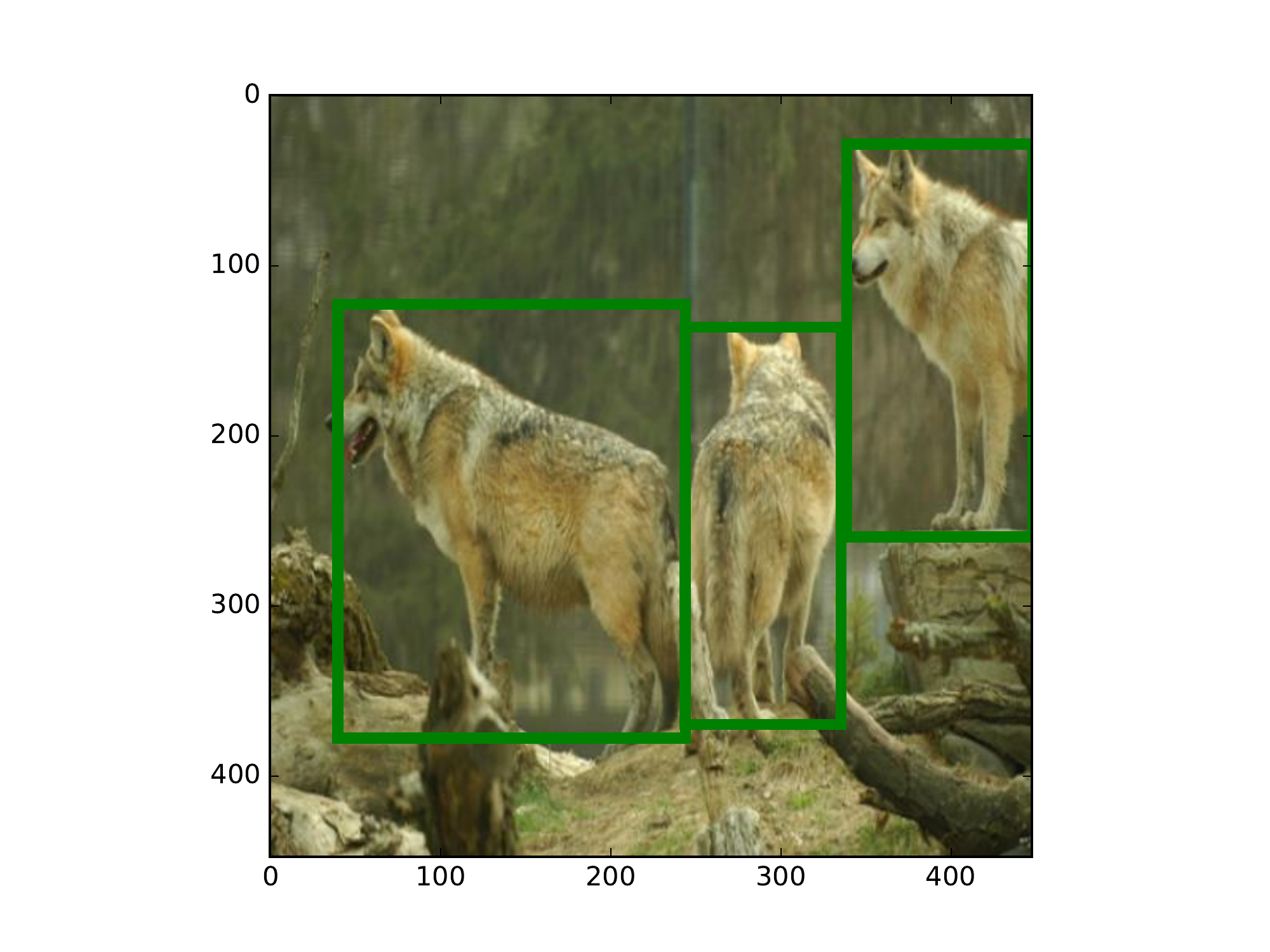}\end{minipage}&
\begin{minipage}{0.1\linewidth}\includegraphics[trim={4.5cm 2cm 4cm 2cm},clip,width=\linewidth]{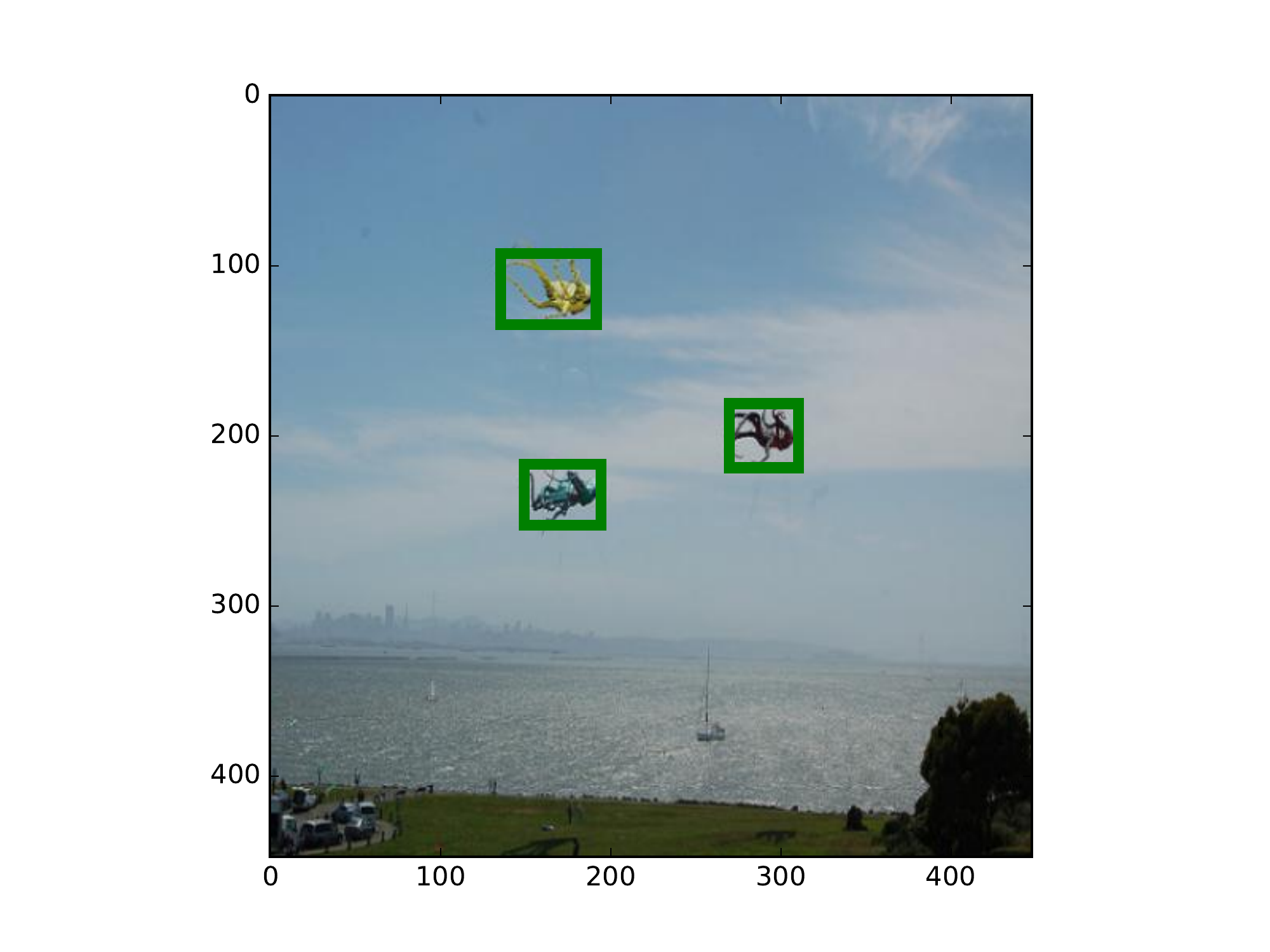}\end{minipage}&
\begin{minipage}{0.1\linewidth}\includegraphics[trim={4.5cm 2cm 4cm 2cm},clip,width=\linewidth]{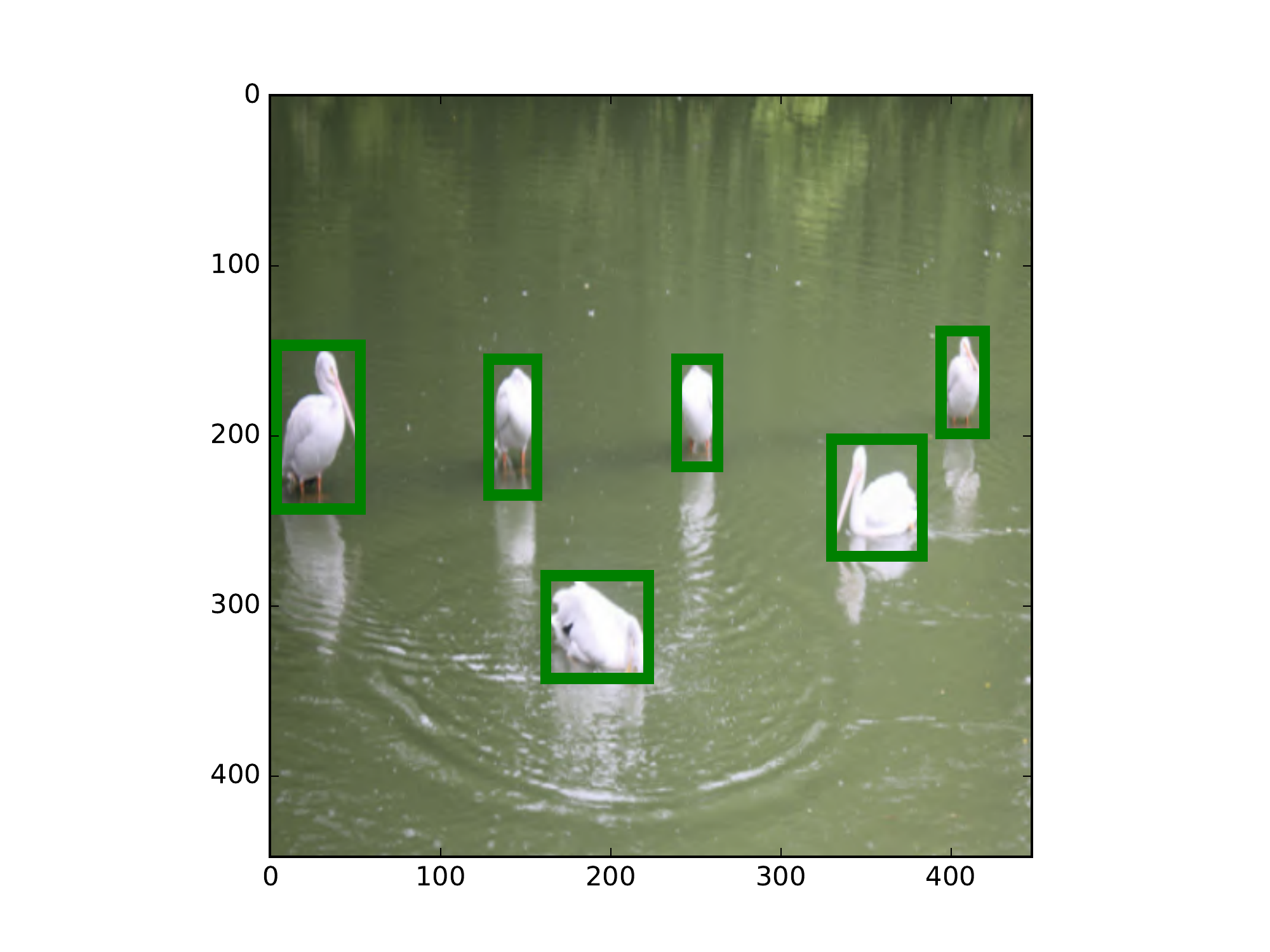}\end{minipage}&
\begin{minipage}{0.1\linewidth}\includegraphics[trim={4.5cm 2cm 4cm 2cm},clip,width=\linewidth]{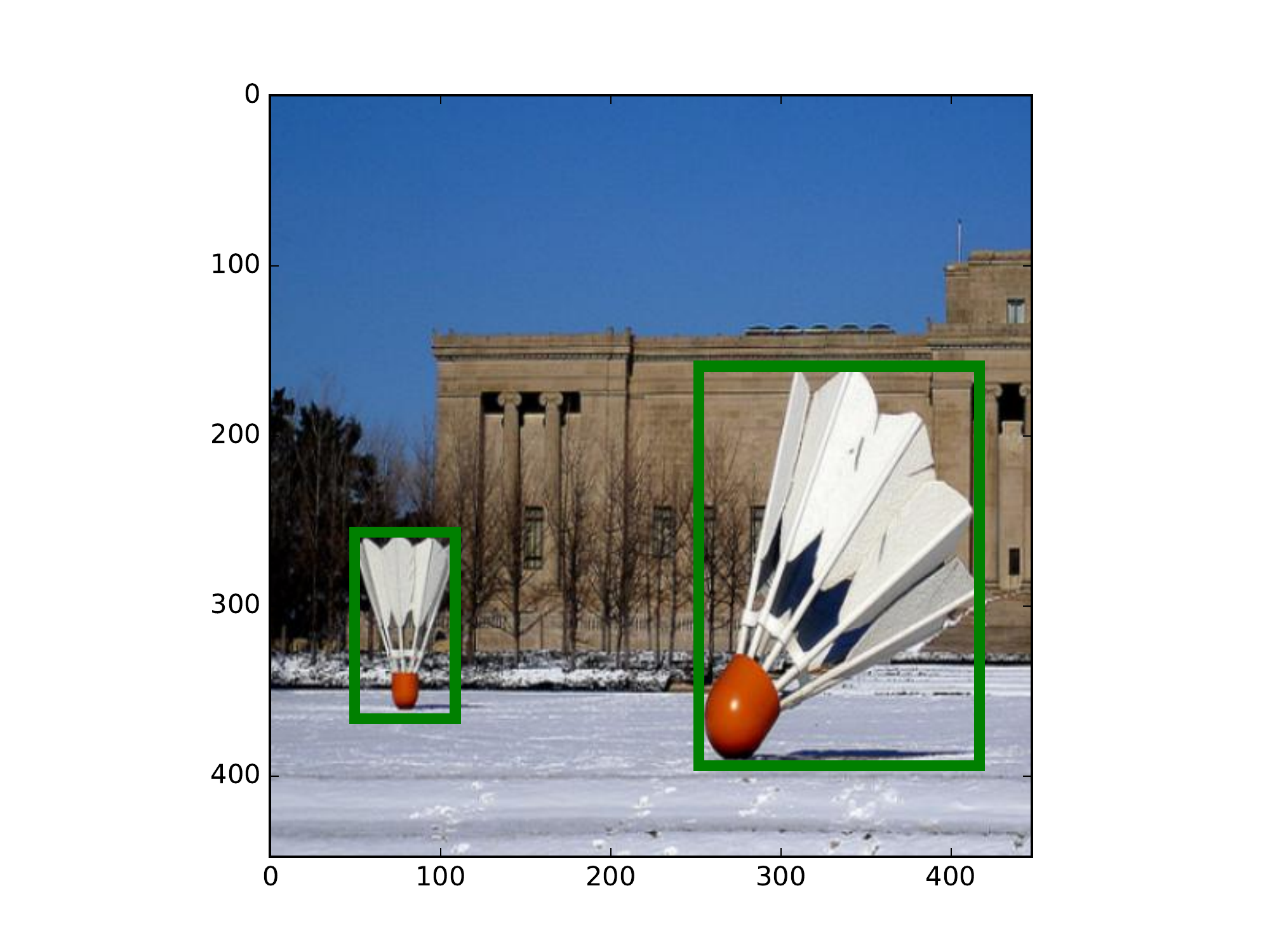}\end{minipage}
\\
\rotatebox[origin=c]{90}{{\footnotesize \textbf{RSD Map}}}& \hspace{-5mm}
\begin{minipage}{0.1\linewidth}\includegraphics[trim={4.5cm 2cm 4cm 2cm},clip,width=\linewidth]{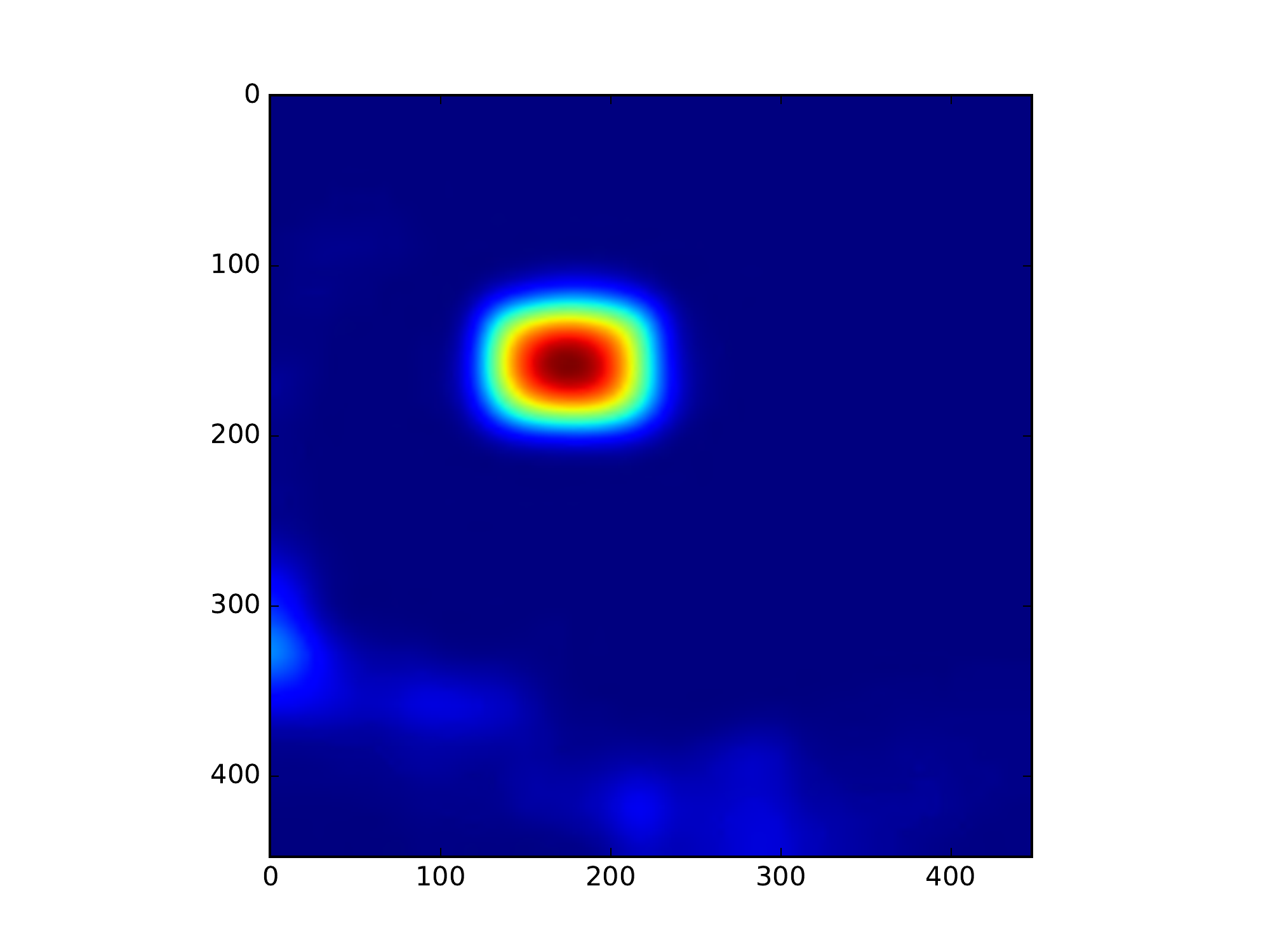}\end{minipage}&
\begin{minipage}{0.1\linewidth}\includegraphics[trim={4.5cm 2cm 4cm 2cm},clip,width=\linewidth]{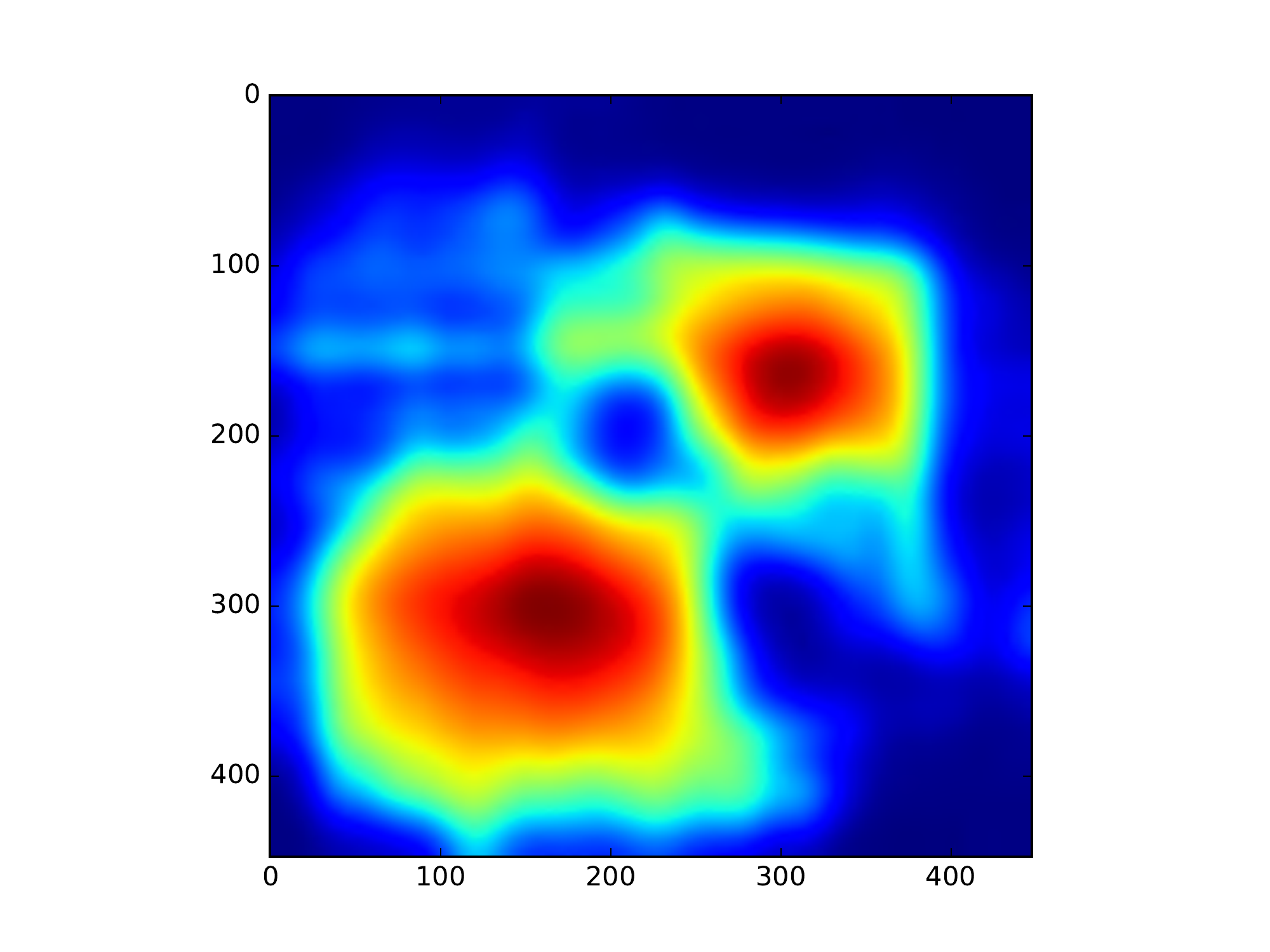}\end{minipage}&
\begin{minipage}{0.1\linewidth}\includegraphics[trim={4.5cm 2cm 4cm 2cm},clip,width=\linewidth]{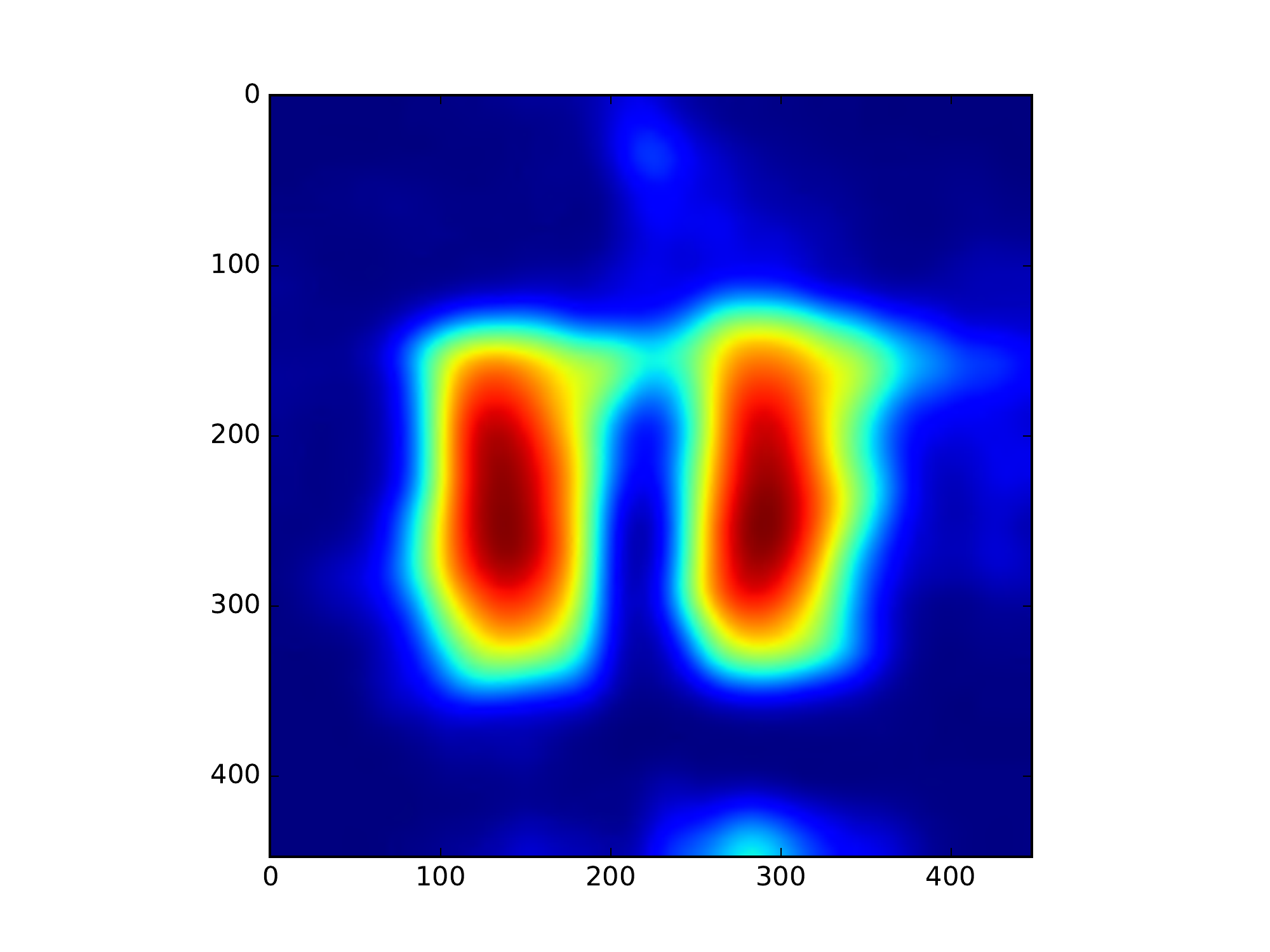}\end{minipage}&
\begin{minipage}{0.1\linewidth}\includegraphics[trim={4.5cm 2cm 4cm 2cm},clip,width=\linewidth]{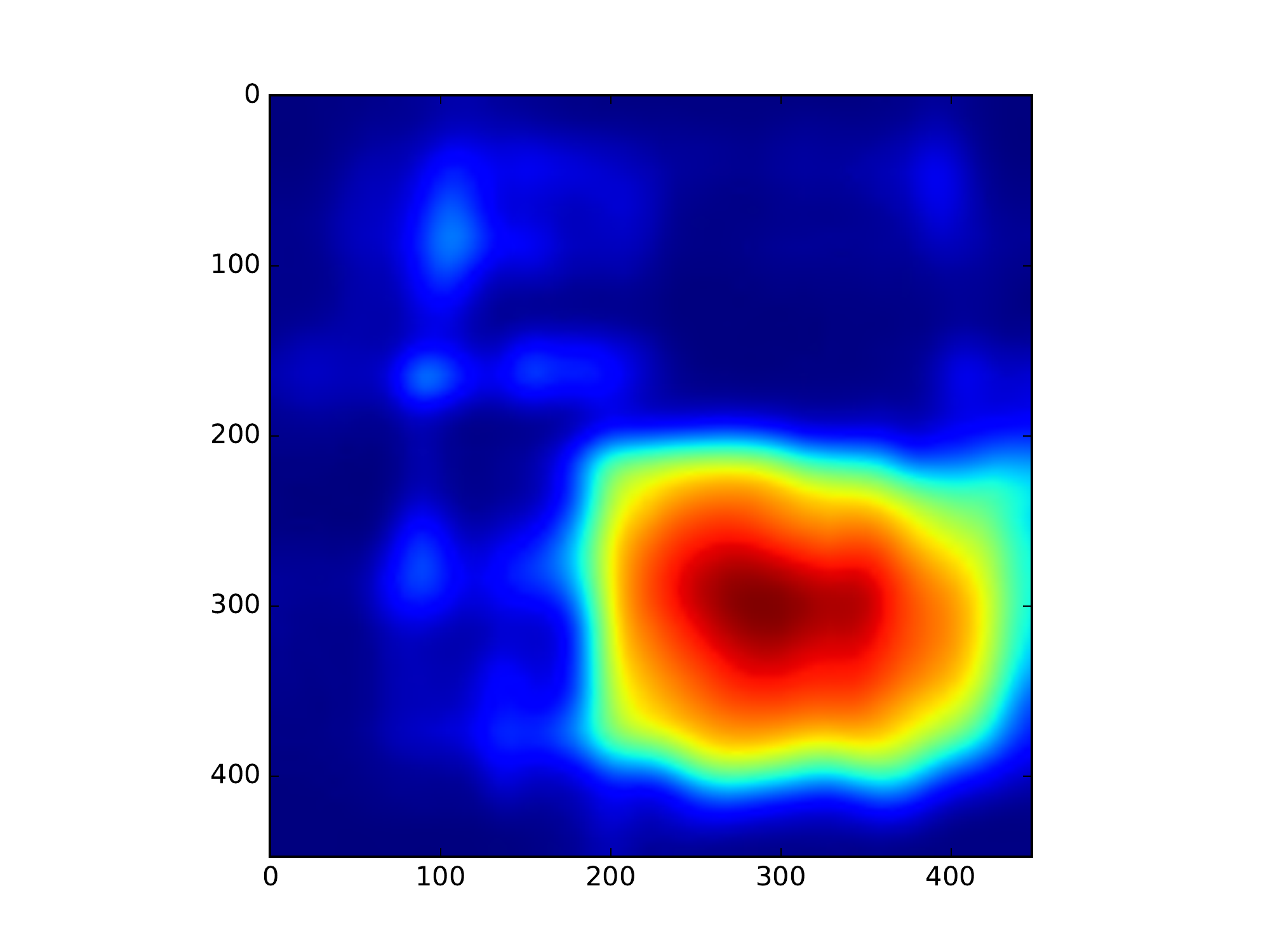}\end{minipage}&
\begin{minipage}{0.1\linewidth}\includegraphics[trim={4.5cm 2cm 4cm 2cm},clip,width=\linewidth]{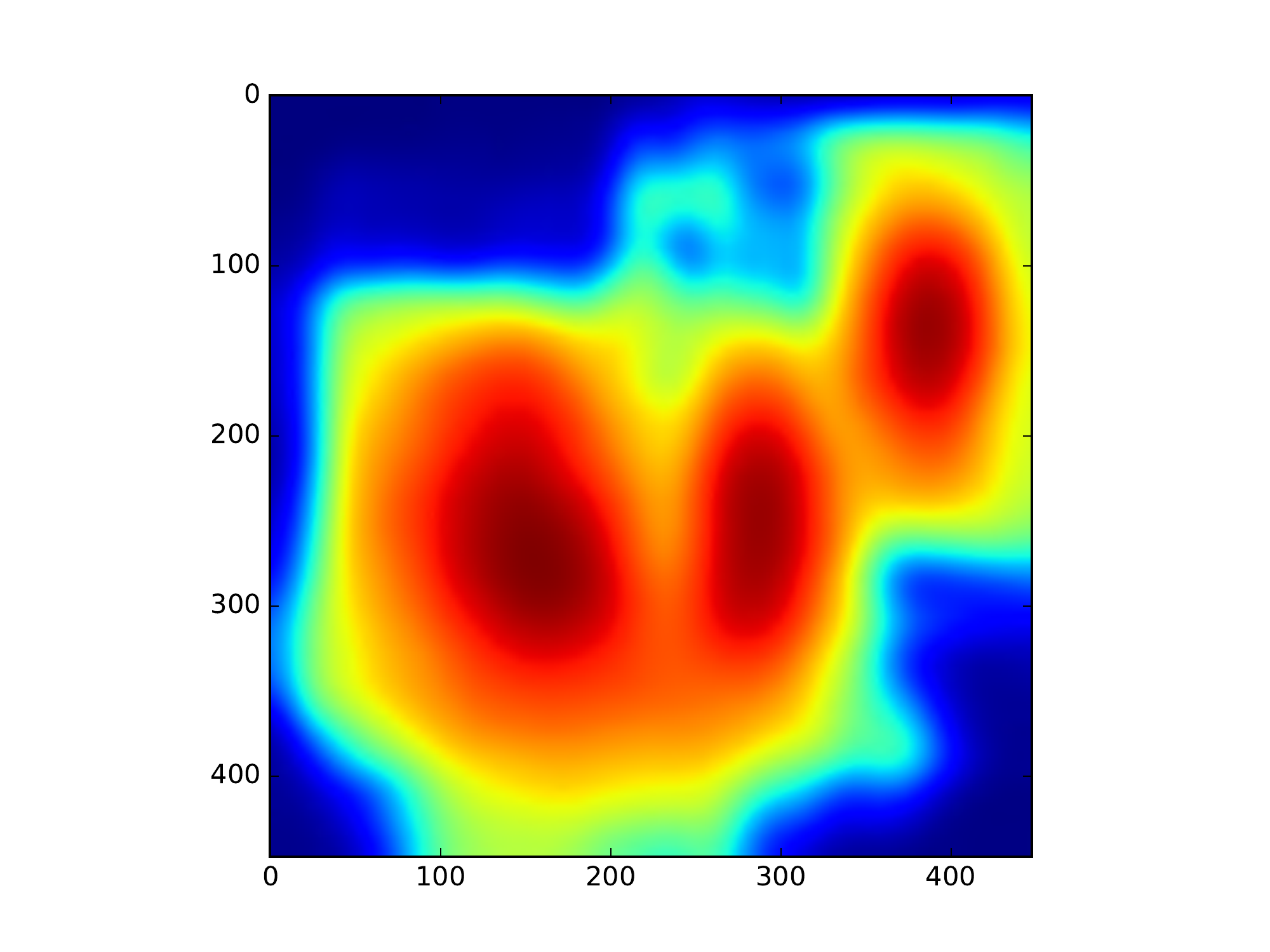}\end{minipage}&
\begin{minipage}{0.1\linewidth}\includegraphics[trim={4.5cm 2cm 4cm 2cm},clip,width=\linewidth]{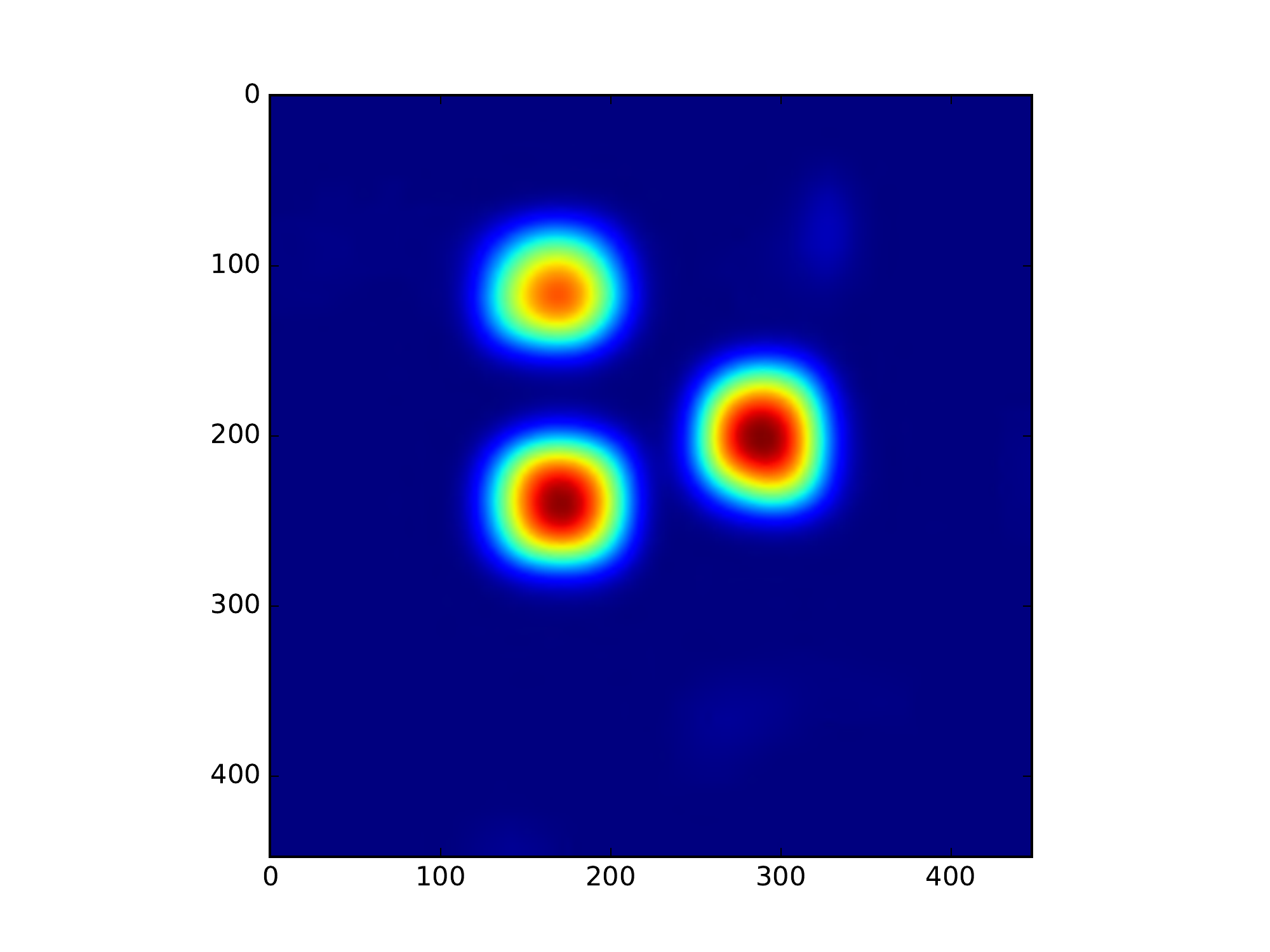}\end{minipage}&
\begin{minipage}{0.1\linewidth}\includegraphics[trim={4.5cm 2cm 4cm 2cm},clip,width=\linewidth]{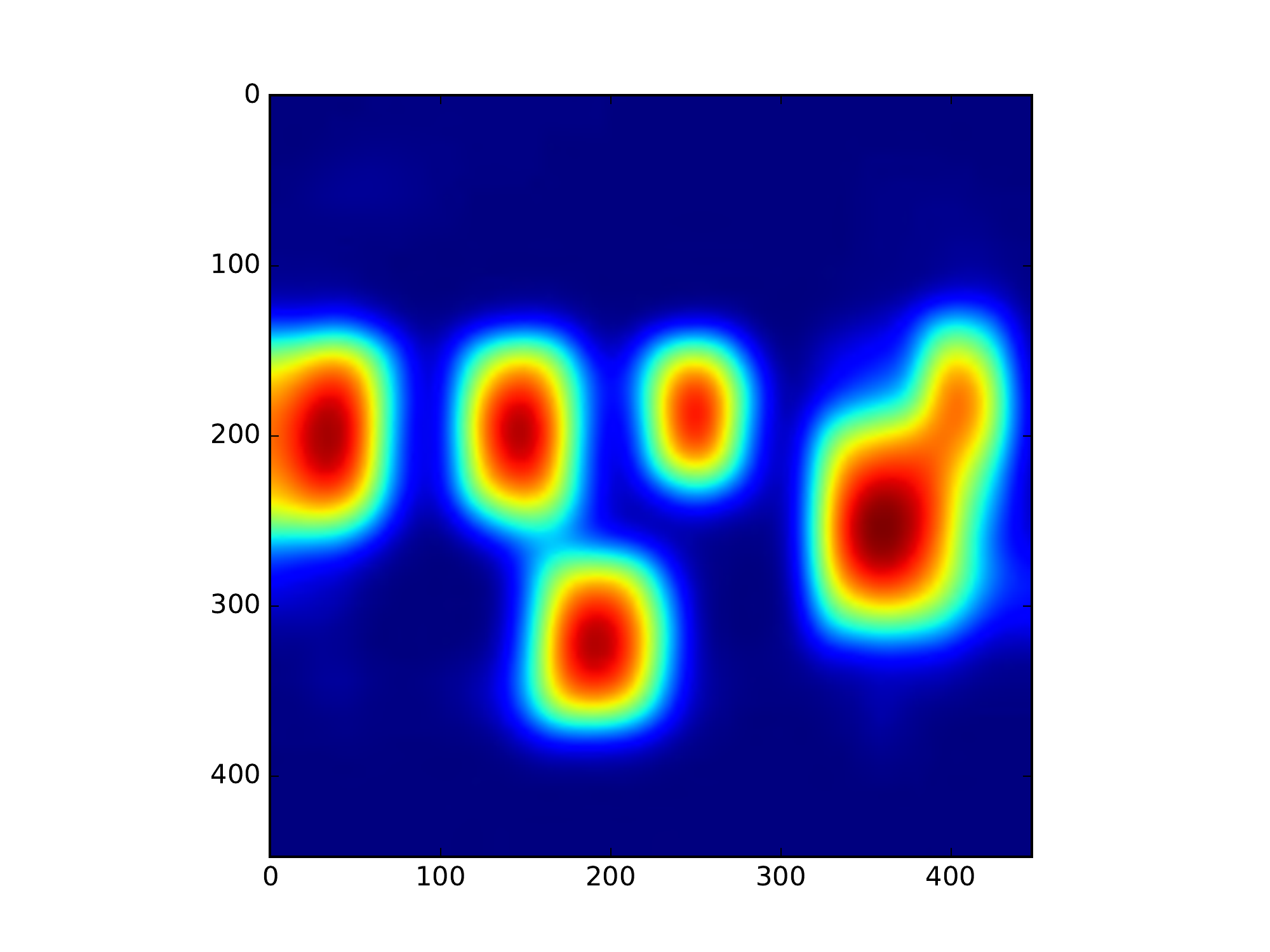}\end{minipage}&
\begin{minipage}{0.1\linewidth}\includegraphics[trim={4.5cm 2cm 4cm 2cm},clip,width=\linewidth]{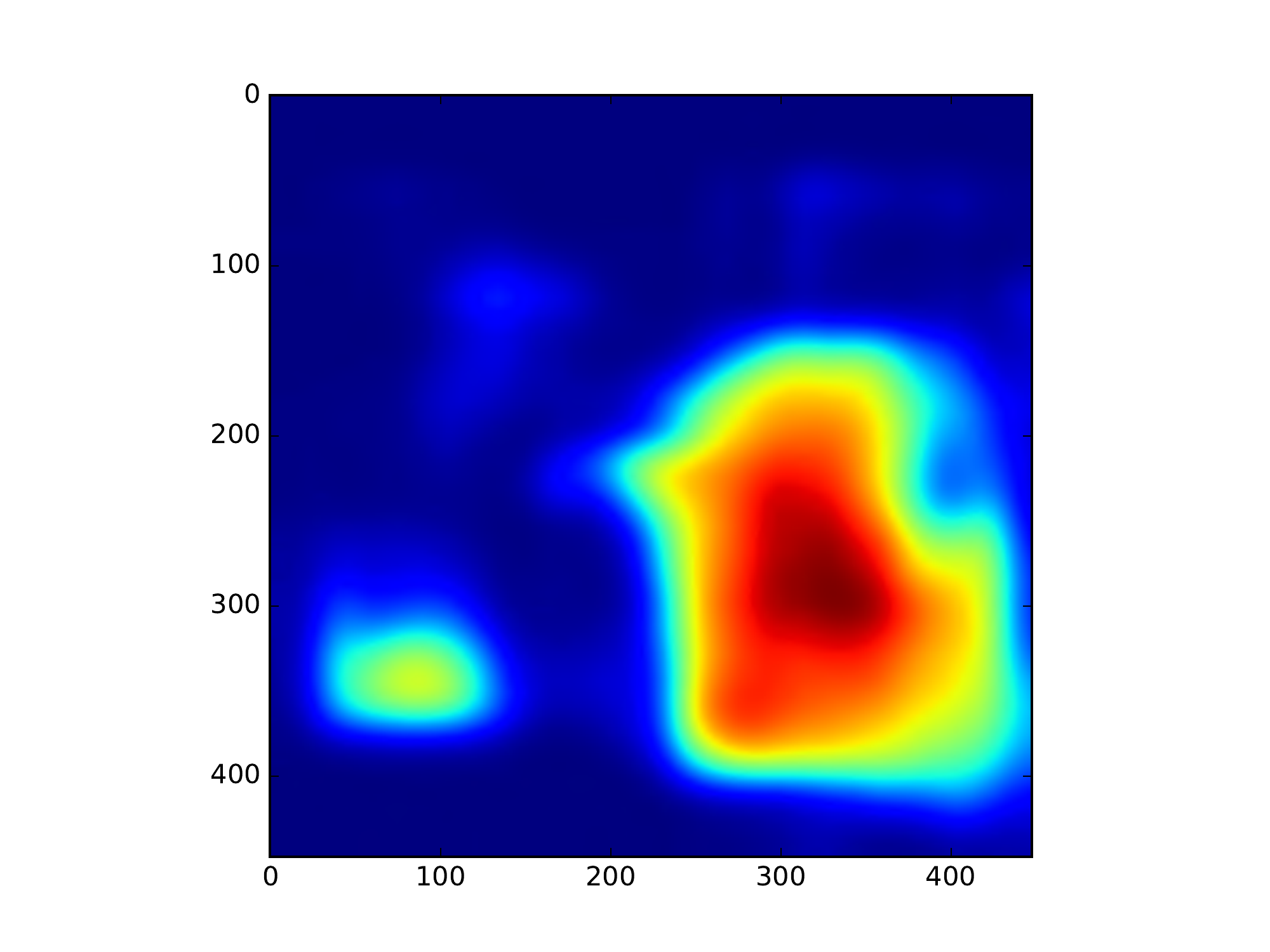}\end{minipage}
\\
\rotatebox[origin=c]{90}{{\footnotesize \textbf{RSD}}}& \hspace{-5mm}
\begin{minipage}{0.1\linewidth}\includegraphics[trim={4.5cm 2cm 4cm 2cm},clip,width=\linewidth]{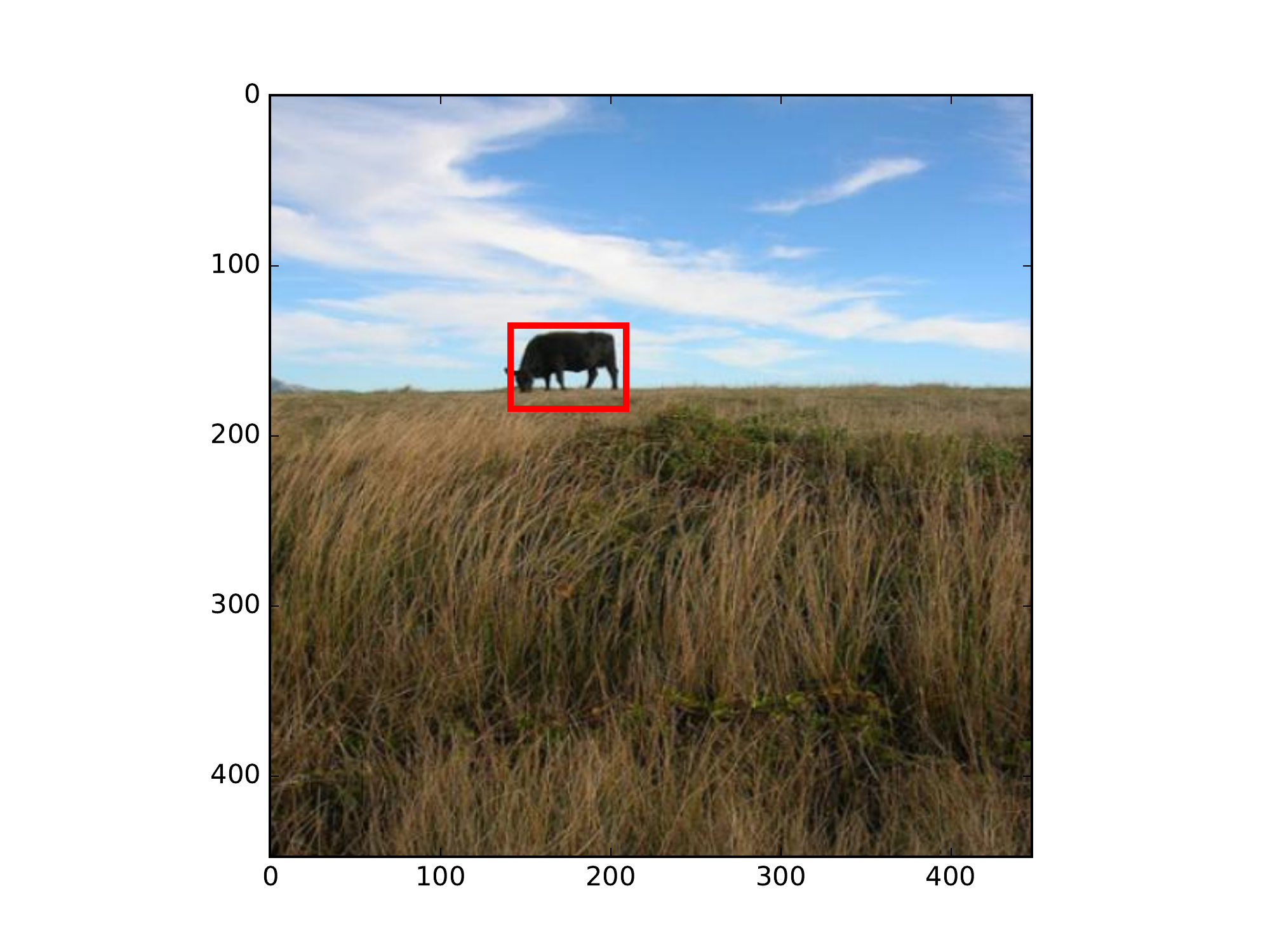}\end{minipage}&
\begin{minipage}{0.1\linewidth}\includegraphics[trim={4.5cm 2cm 4cm 2cm},clip,width=\linewidth]{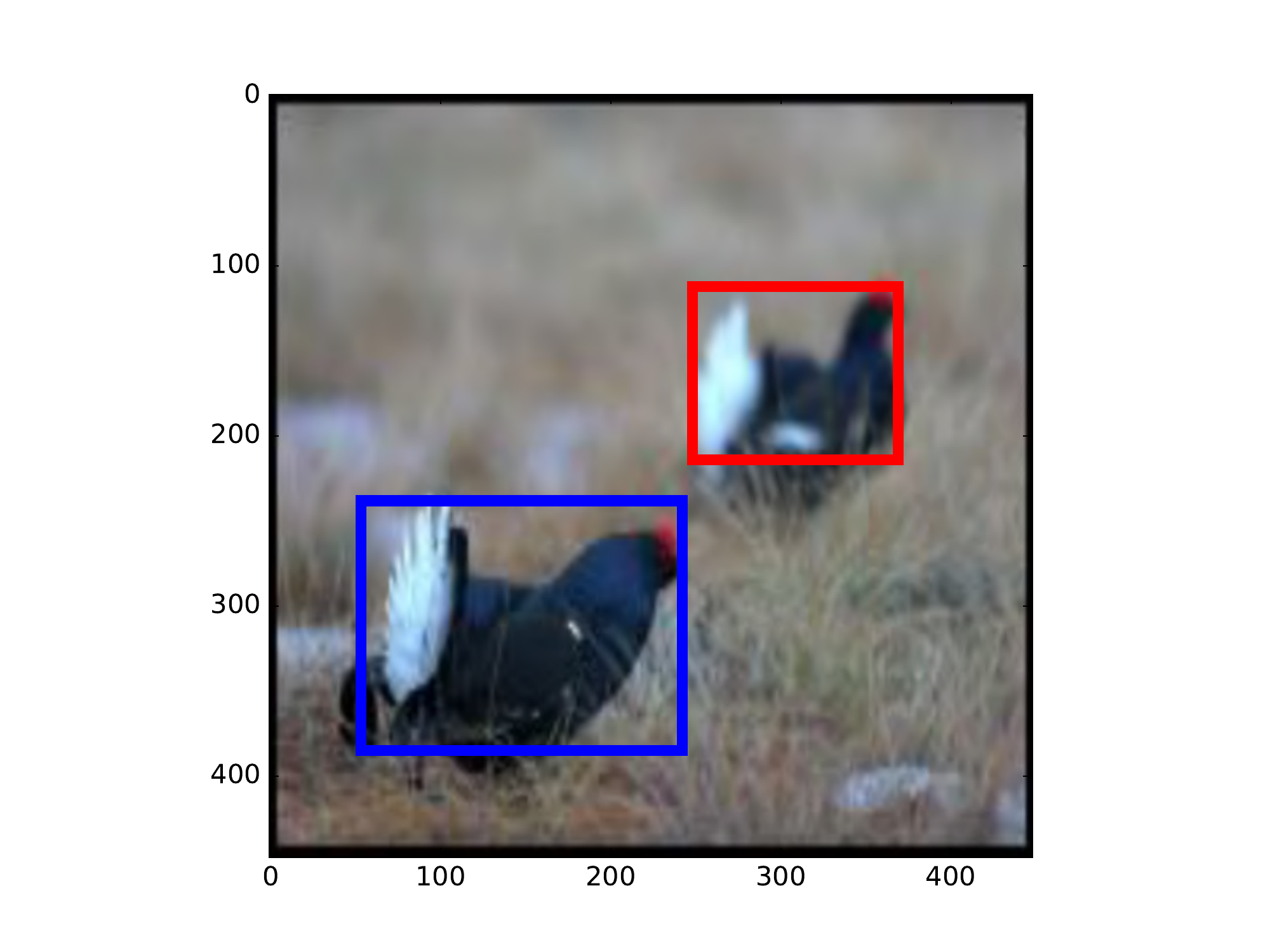}\end{minipage}&
\begin{minipage}{0.1\linewidth}\includegraphics[trim={4.5cm 2cm 4cm 2cm},clip,width=\linewidth]{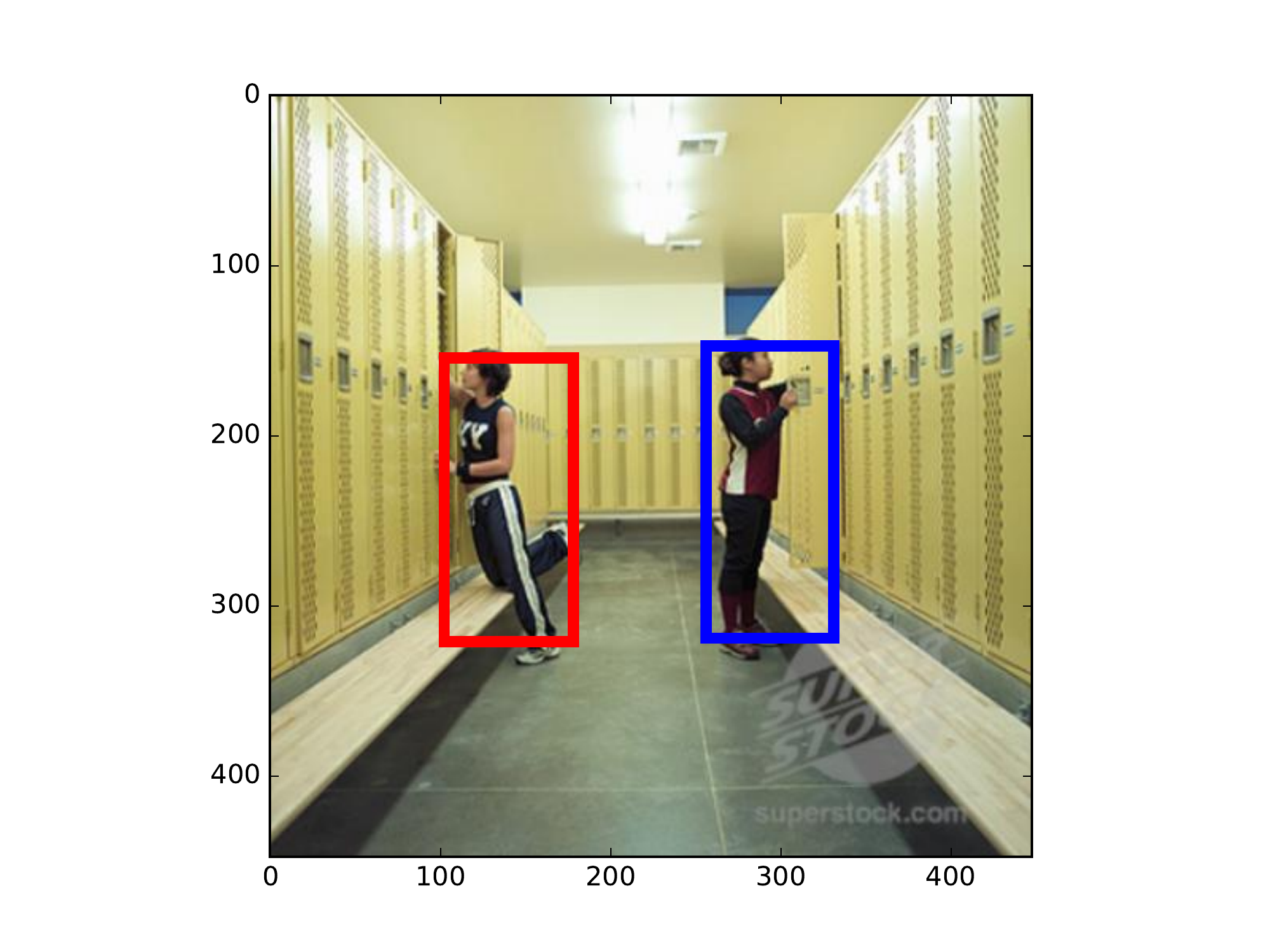}\end{minipage}&
\begin{minipage}{0.1\linewidth}\includegraphics[trim={4.5cm 2cm 4cm 2cm},clip,width=\linewidth]{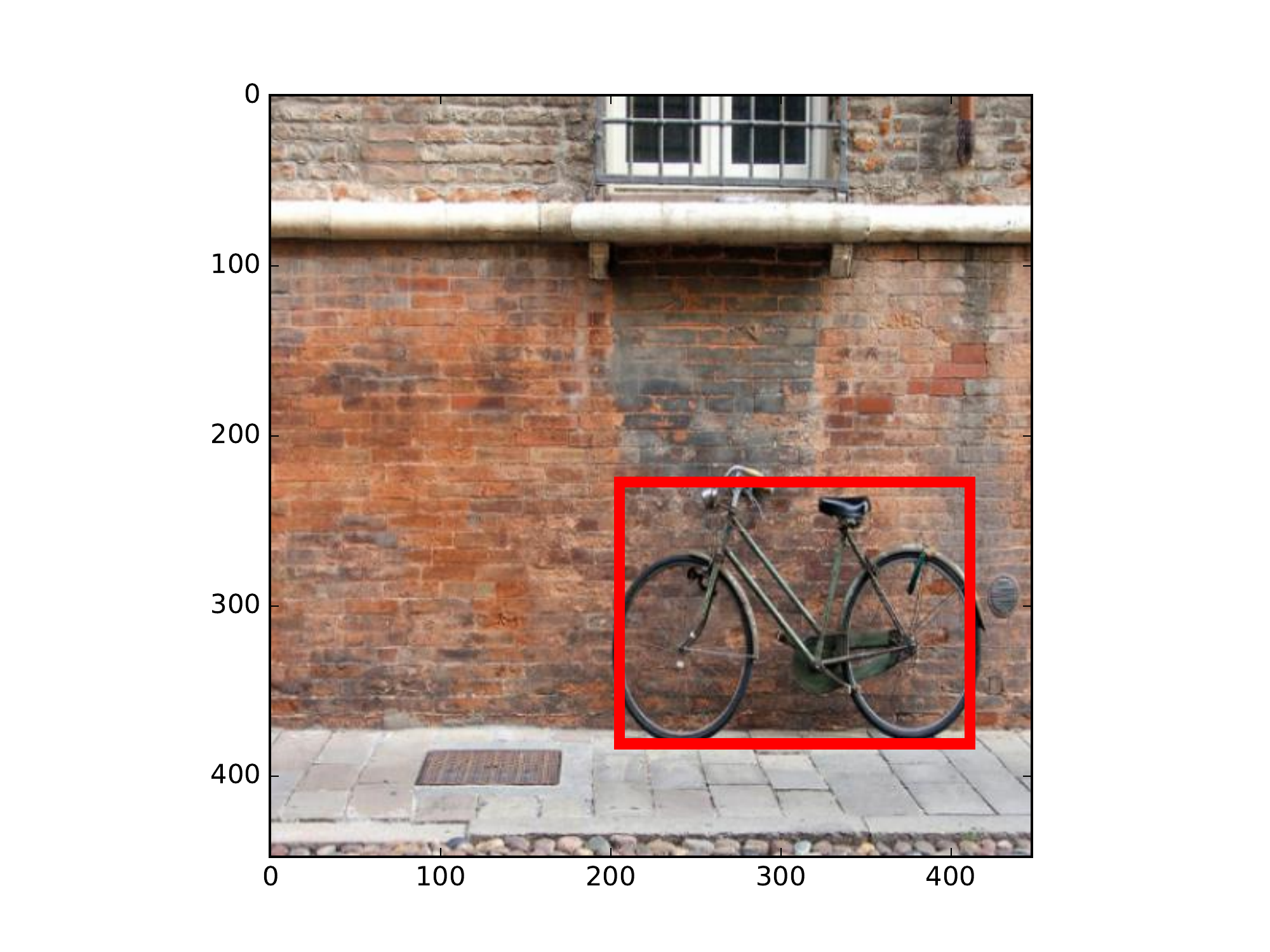}\end{minipage}&
\begin{minipage}{0.1\linewidth}\includegraphics[trim={4.5cm 2cm 4cm 2cm},clip,width=\linewidth]{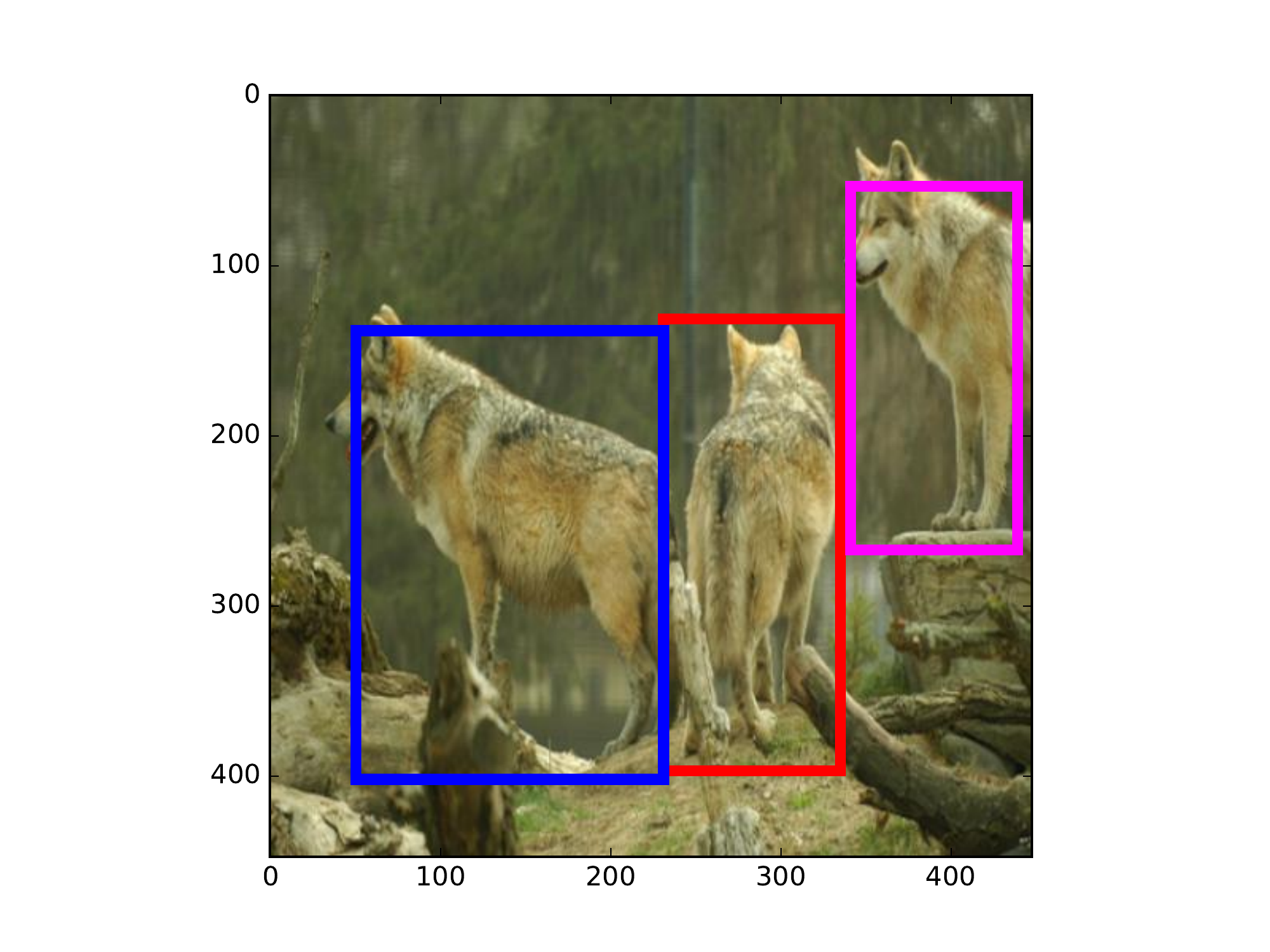}\end{minipage}&
\begin{minipage}{0.1\linewidth}\includegraphics[trim={4.5cm 2cm 4cm 2cm},clip,width=\linewidth]{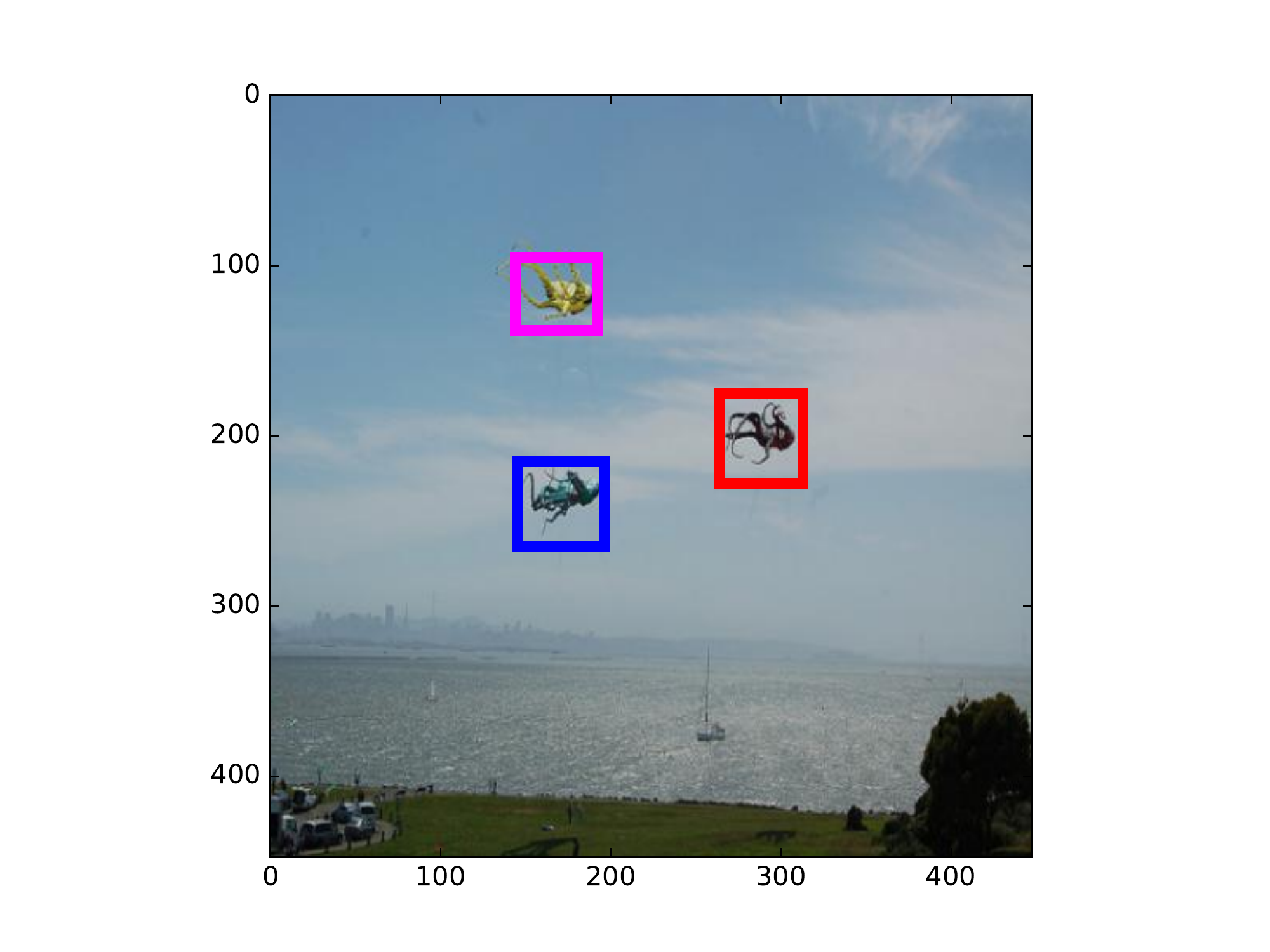}\end{minipage}&
\begin{minipage}{0.1\linewidth}\includegraphics[trim={4.5cm 2cm 4cm 2cm},clip,width=\linewidth]{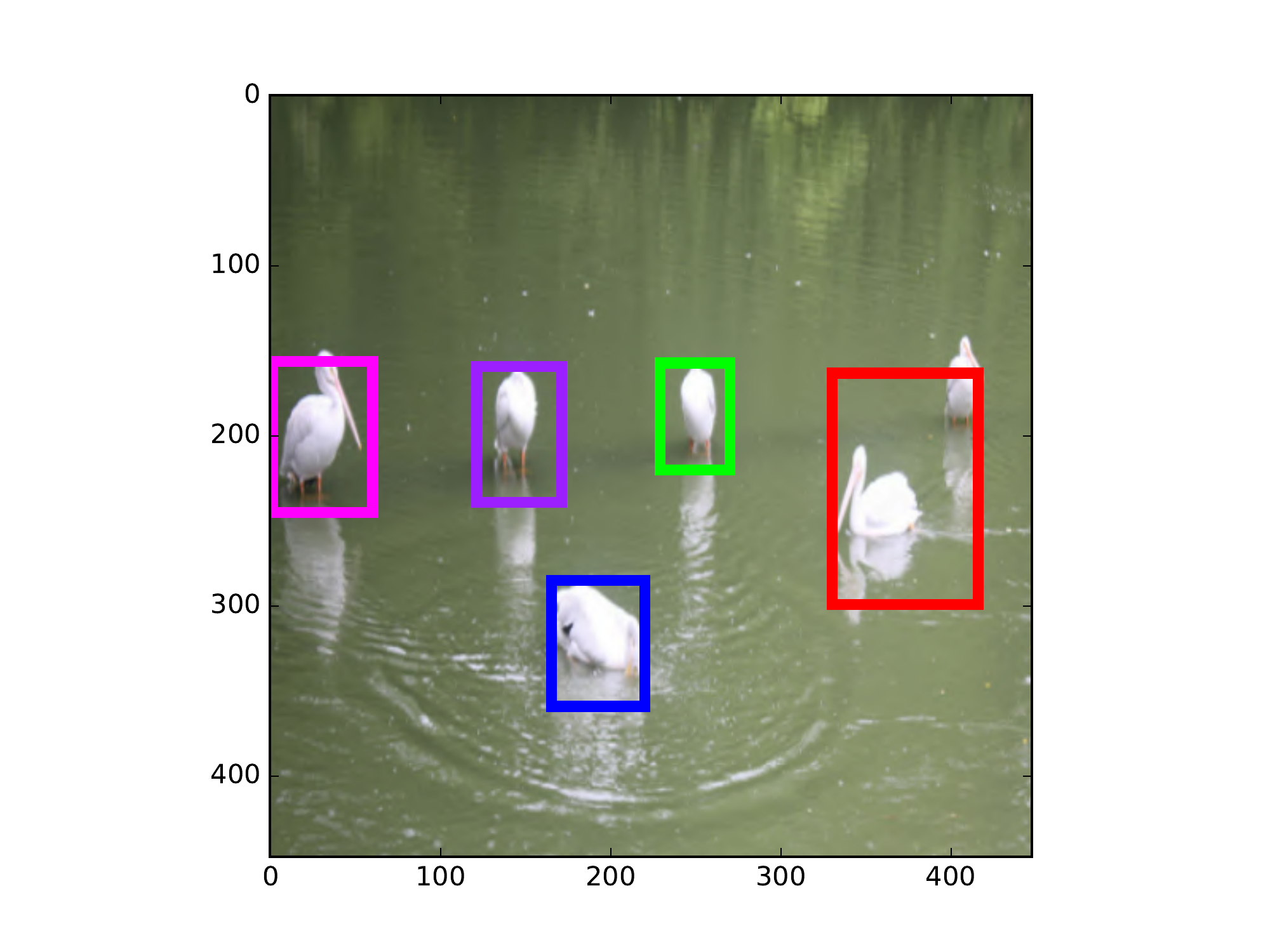}\end{minipage}&
\begin{minipage}{0.1\linewidth}\includegraphics[trim={4.5cm 2cm 4cm 2cm},clip,width=\linewidth]{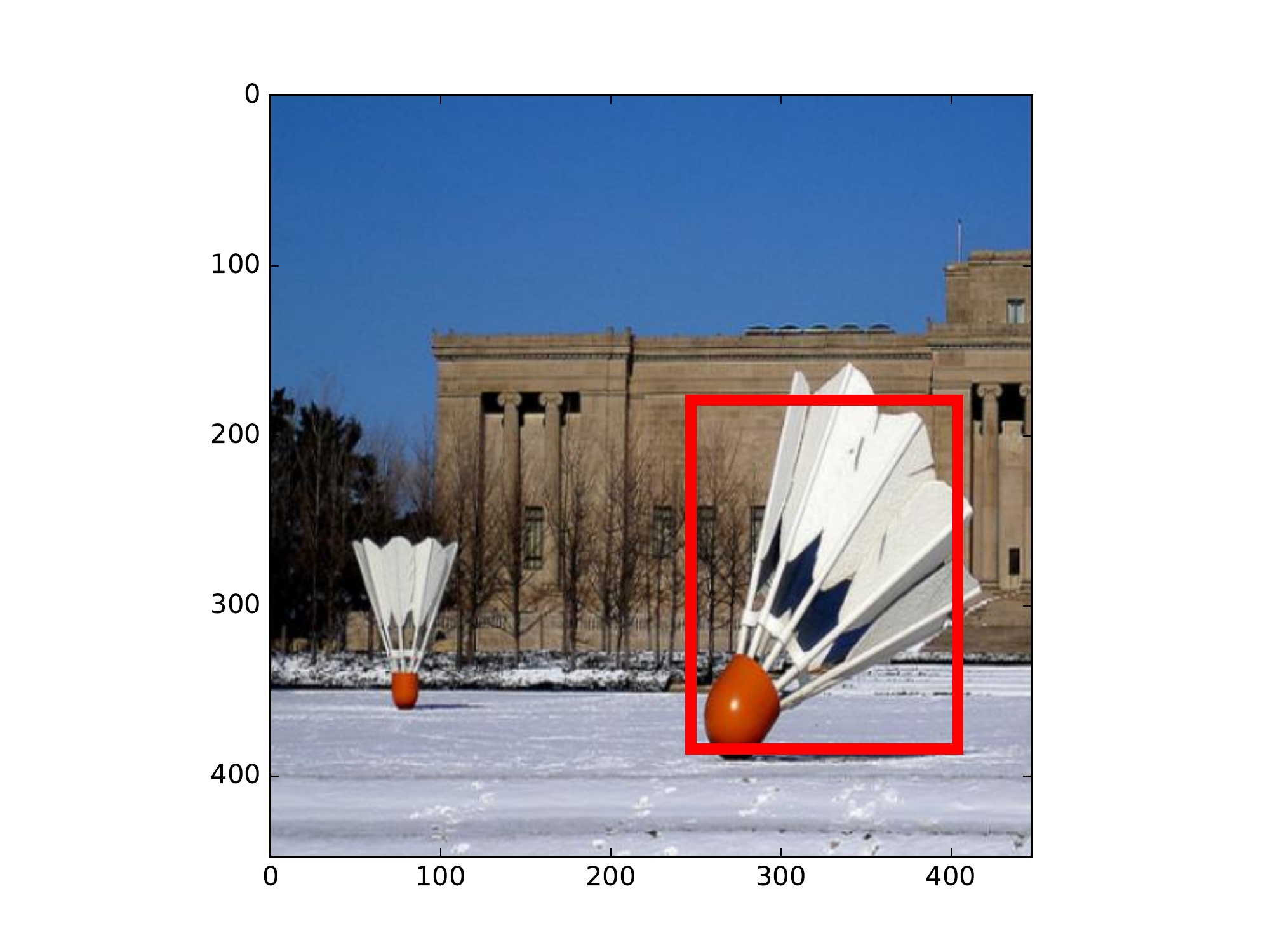}\end{minipage}
\\
\rotatebox[origin=c]{90}{{\footnotesize \textbf{SalCNN+MAP}}}& \hspace{-5mm}
\begin{minipage}{0.1\linewidth}\includegraphics[trim={4.5cm 2cm 4cm 2cm},clip,width=\linewidth]{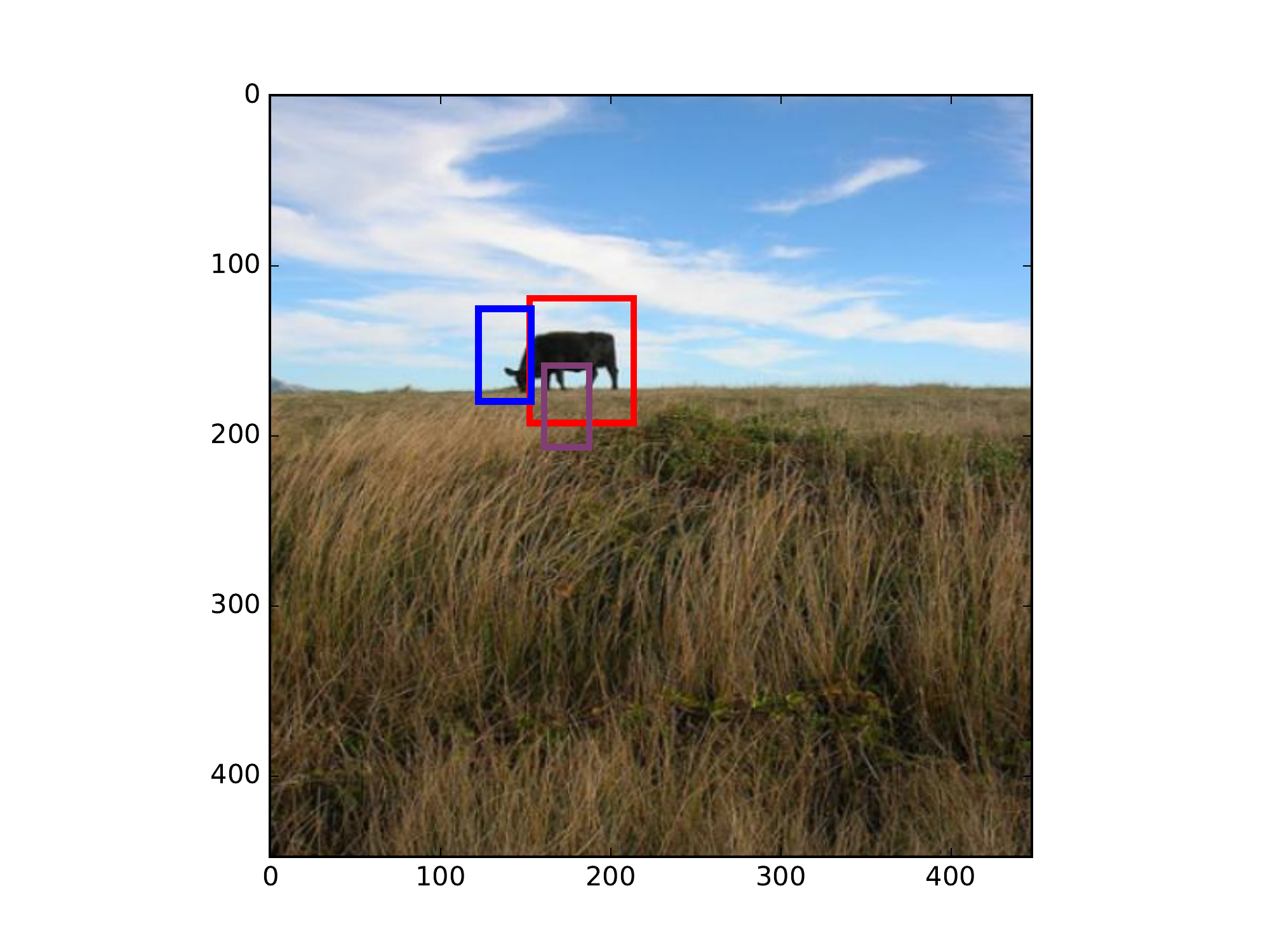}\end{minipage}&
\begin{minipage}{0.1\linewidth}\includegraphics[trim={4.5cm 2cm 4cm 2cm},clip,width=\linewidth]{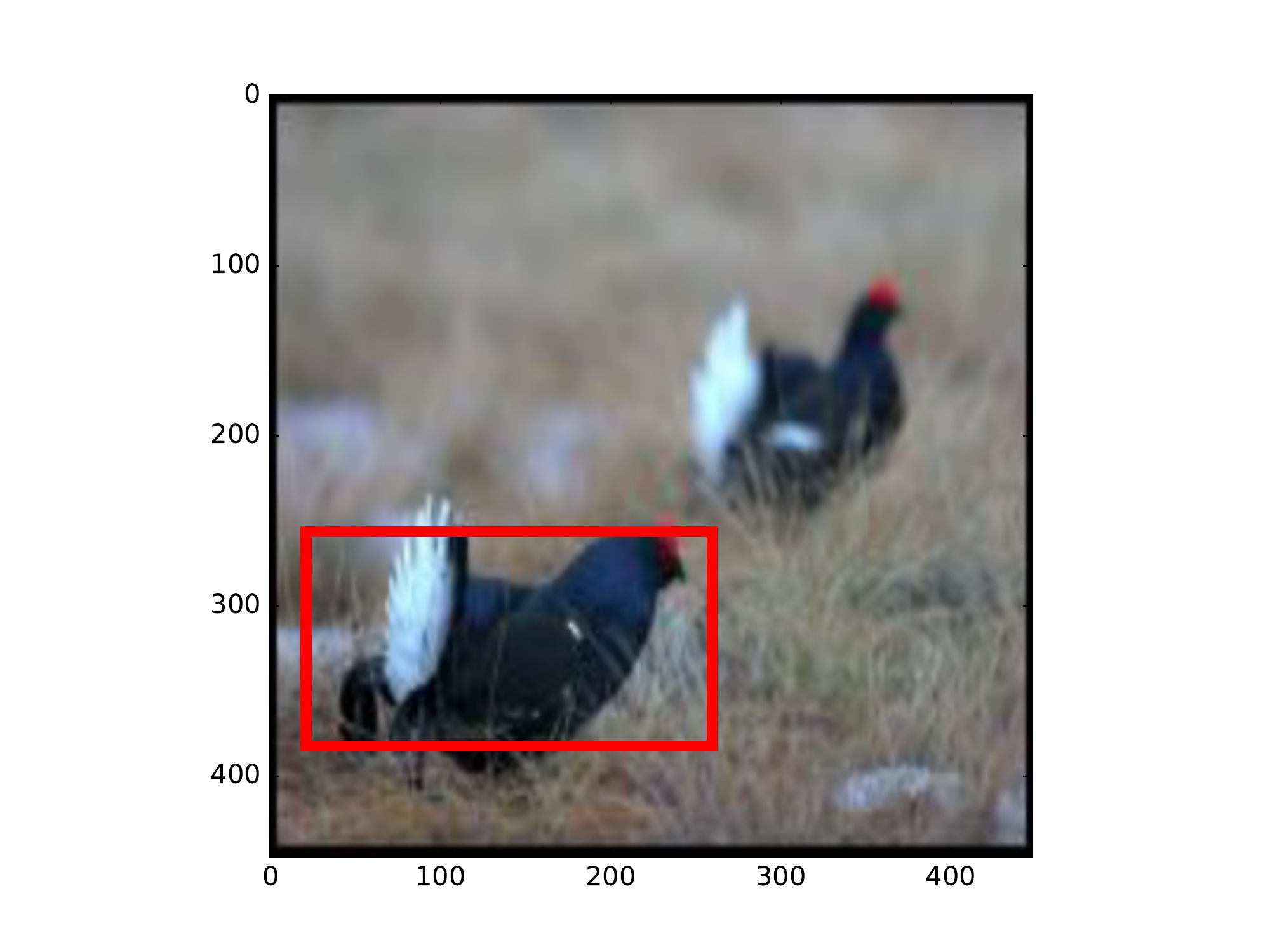}\end{minipage}&
\begin{minipage}{0.1\linewidth}\includegraphics[trim={4.5cm 2cm 4cm 2cm},clip,width=\linewidth]{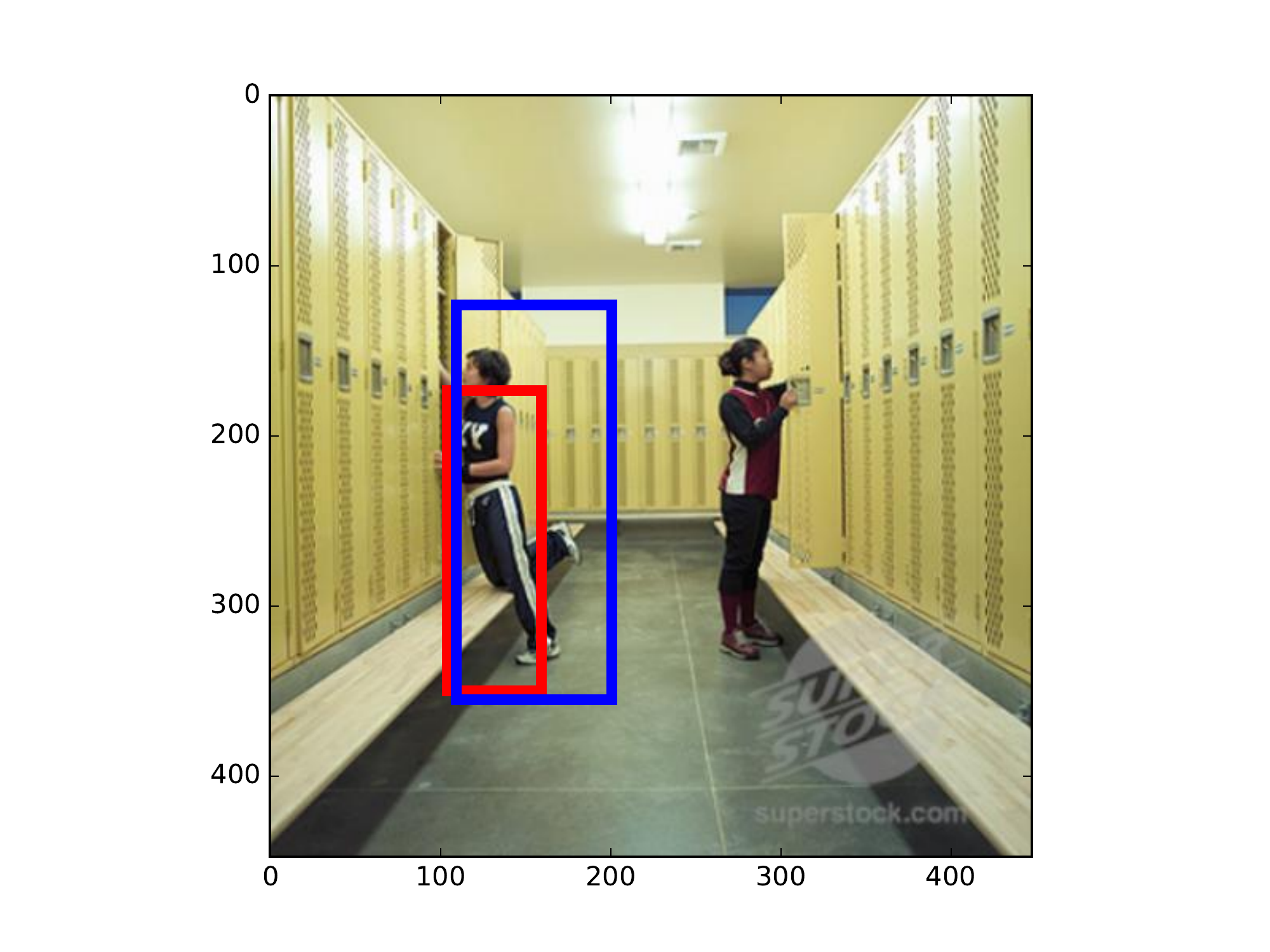}\end{minipage}&
\begin{minipage}{0.1\linewidth}\includegraphics[trim={4.5cm 2cm 4cm 2cm},clip,width=\linewidth]{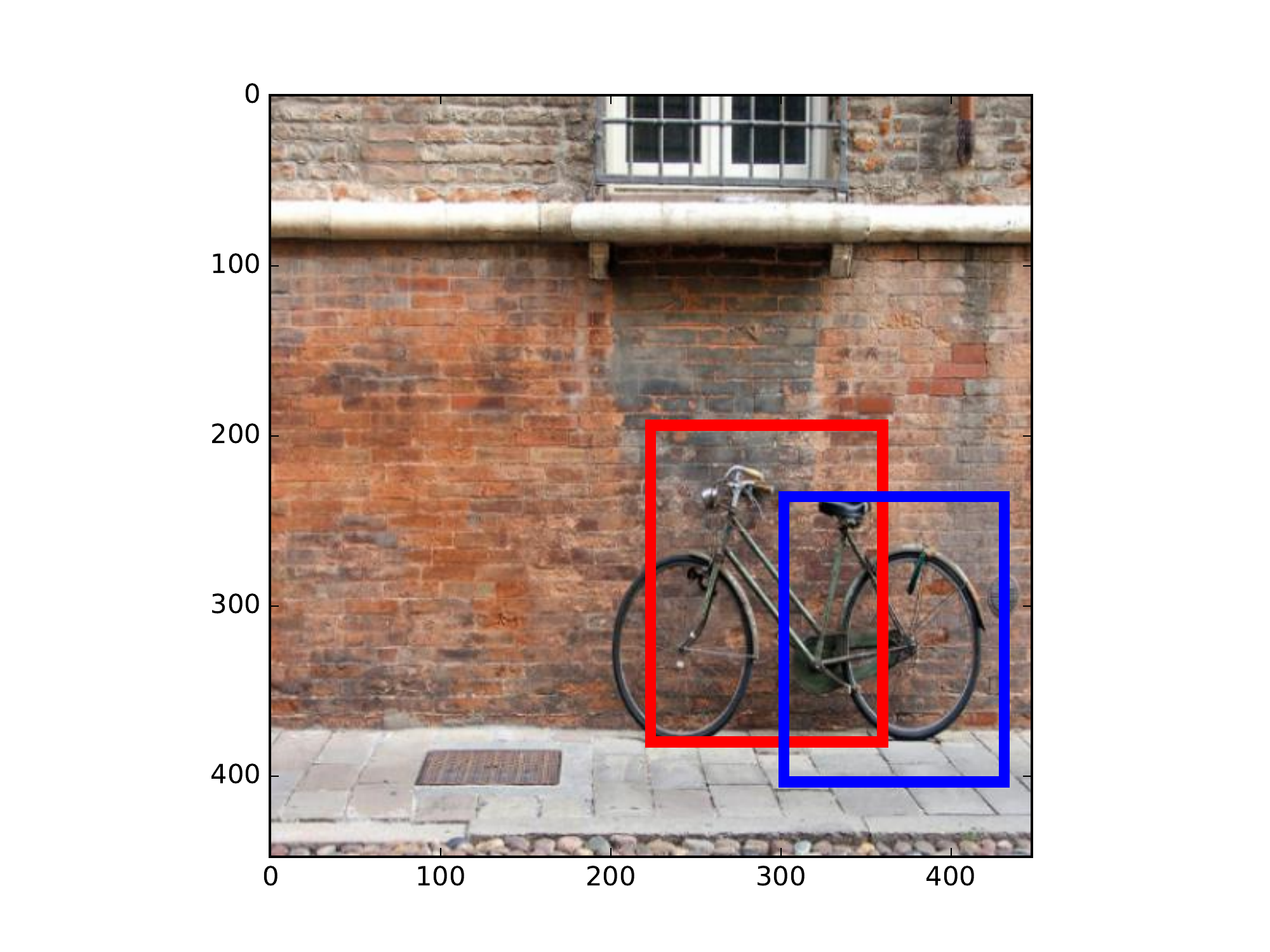}\end{minipage}&
\begin{minipage}{0.1\linewidth}\includegraphics[trim={4.5cm 2cm 4cm 2cm},clip,width=\linewidth]{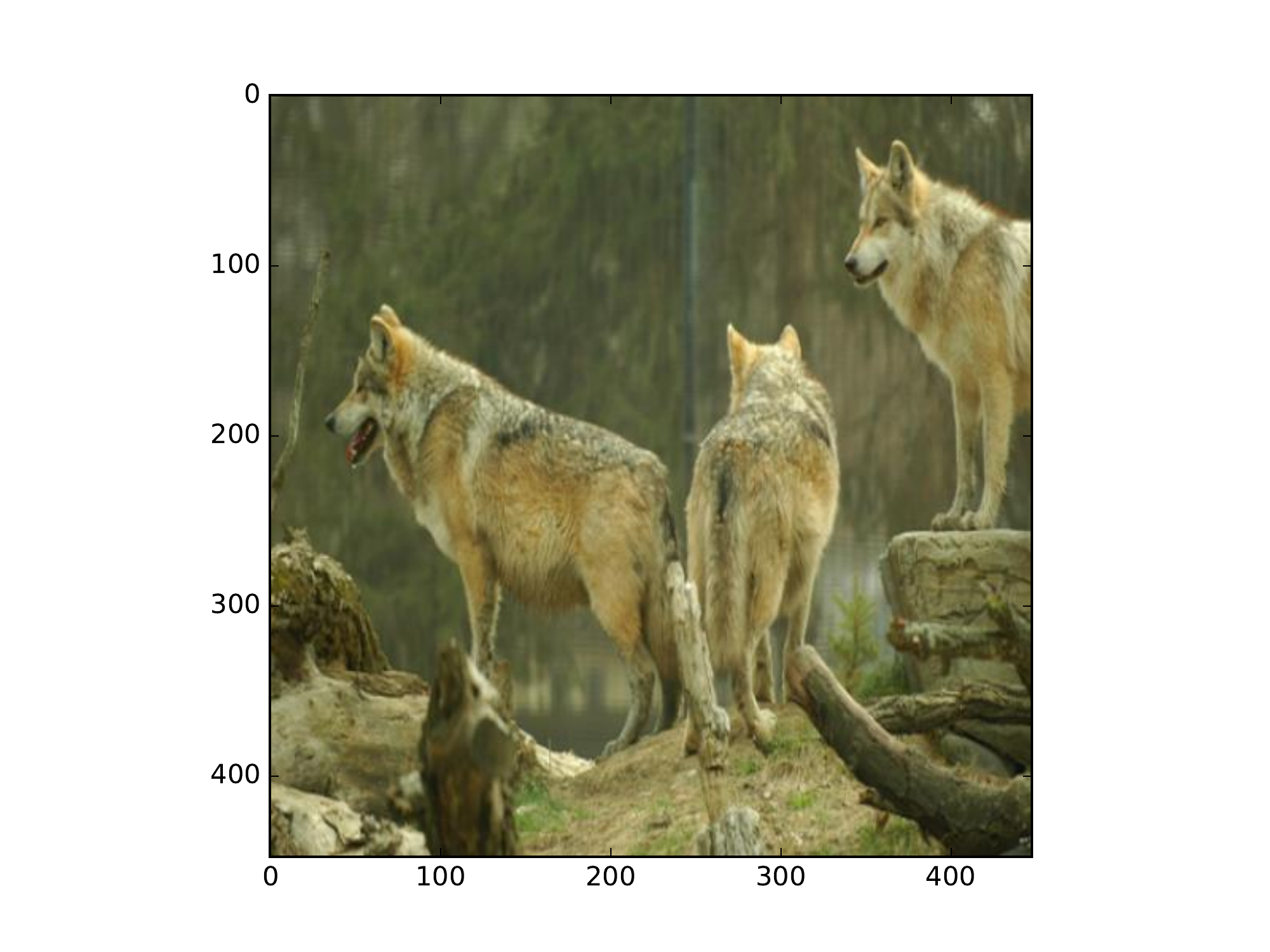}\end{minipage}&
\begin{minipage}{0.1\linewidth}\includegraphics[trim={4.5cm 2cm 4cm 2cm},clip,width=\linewidth]{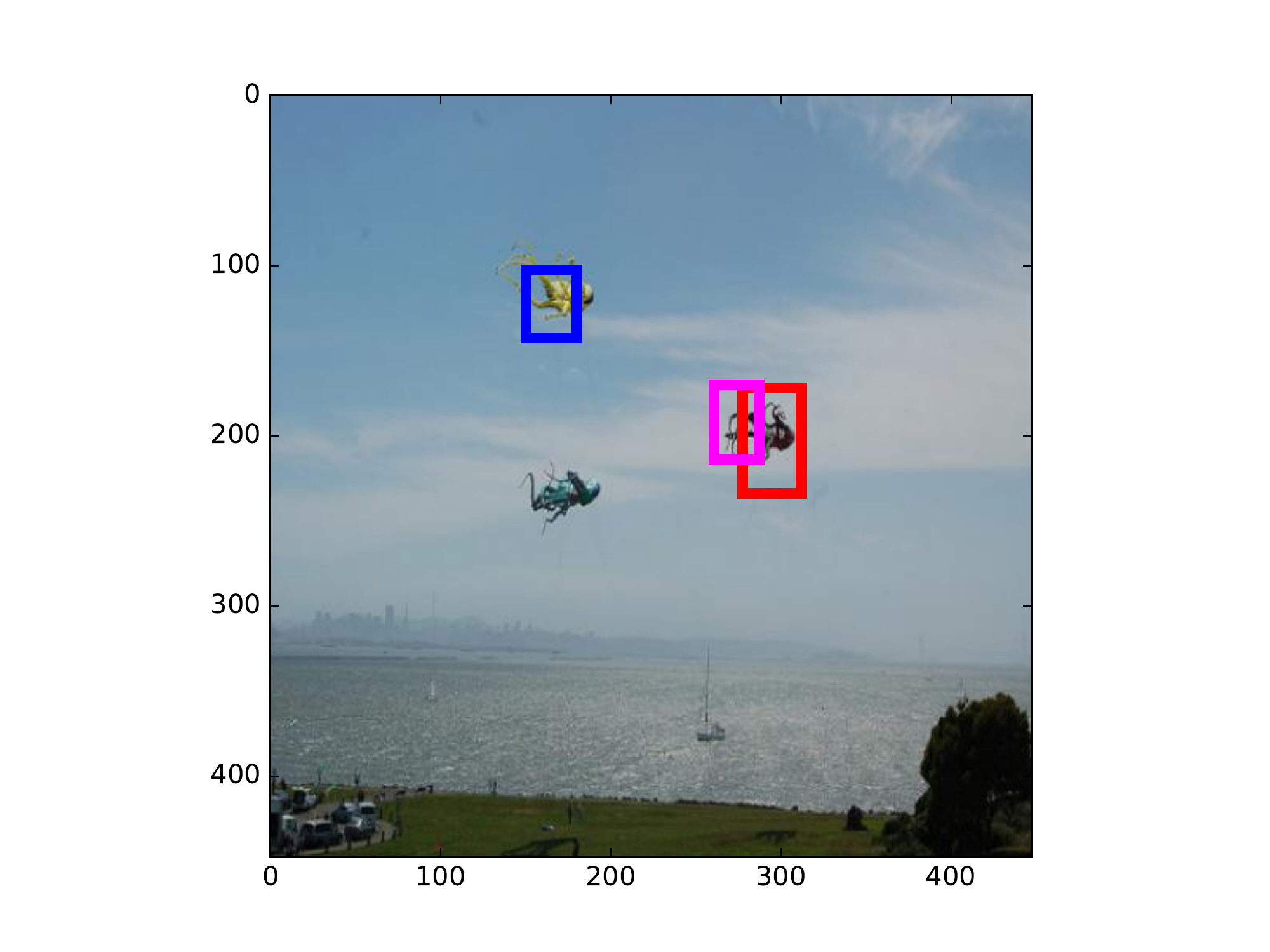}\end{minipage}&
\begin{minipage}{0.1\linewidth}\includegraphics[trim={4.5cm 2cm 4cm 2cm},clip,width=\linewidth]{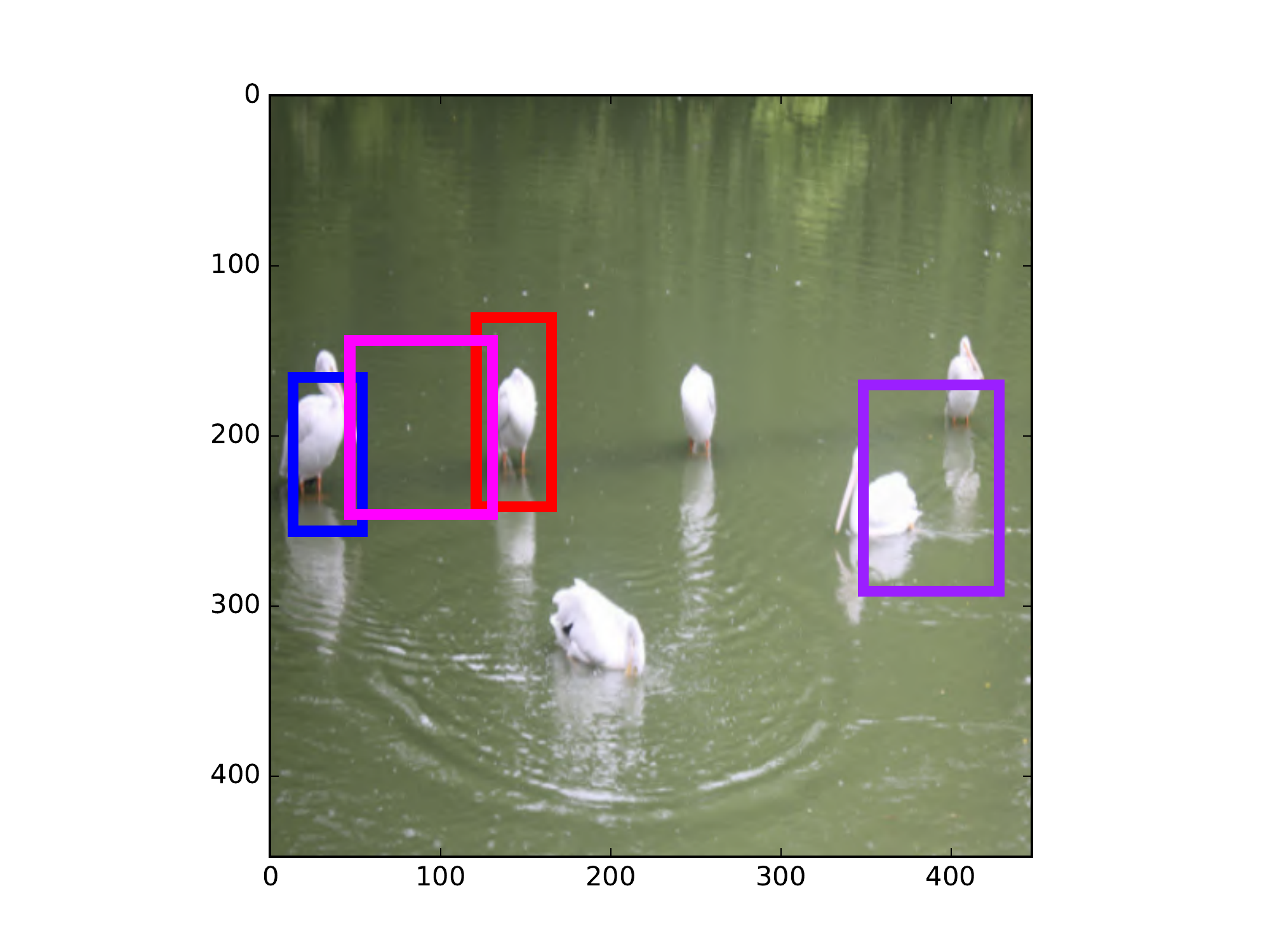}\end{minipage}&
\begin{minipage}{0.1\linewidth}\includegraphics[trim={4.5cm 2cm 4cm 2cm},clip,width=\linewidth]{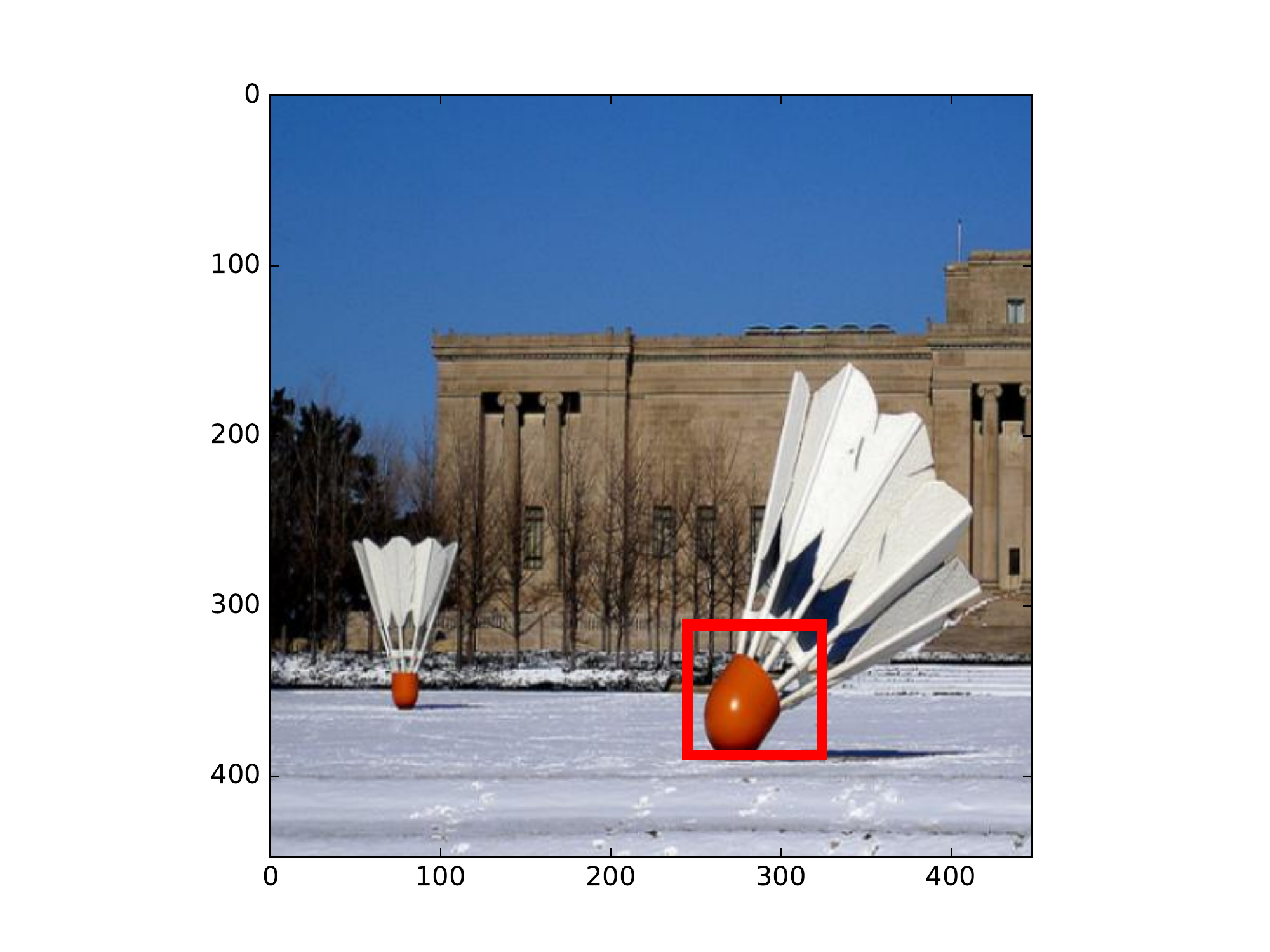}\end{minipage}
\end{tabular}
\caption{Qualitative results of our RSD-ResNet and SalCNN+MAP. Note that our method locates multiple salient object in the scene more accurately than SalCNN+MAP. The last two columns show hard examples where both SalCNN+MAP and ours cannot locate all the salient objects.}
\label{fig:qual}
\end{figure*}

{\flushleft \textbf{Localization.}} Although we do not use proposals and pruning stage like NMS, our straightforward bounding box generation algorithm generates good results. Moreover, bounding boxes generated by our method align with the ground-truth better compared to existing approaches, leading to the best precision and recall, as shown in Figure~\ref{fig:iou}.
In this experiment, we let other methods to pick their parameters to get the same recall as ours at IoU$=0.5$, and then change the IoU threshold to evaluate the performance change. 
Notably, if we have a more strict IoU criteria, such as 0.8, RSD still maintains a relatively high precision and recall, while the precision and recall of all the other methods greatly drop. At this IoU, even our fast RSD-VGG is able to outperform the state-of-the-art methods on all datasets by an average margin of around 10\% in terms of both precision and recall. The results clearly demonstrate that our network successfully predicts an accurate saliency map and easily generates only a few bounding boxes tightly enclosing the correct salient objects. Some qualitative results are presented in Figure~\ref{fig:qual}. Our RSD approach clearly outperforms SalCNN+MAP in generating better bounding boxes that more tightly enclose the ground-truth.

\begin{table*}[!h]
\renewcommand{\arraystretch}{0.95}
\scriptsize
  \centering
  \caption{The precision/recall of our RSD-ResNet and other methods on different datasets and objects of different sizes. The IoU threshold is $0.5$. $S$, $S_s$ and $S_l$ denote all objects, objects of small and large size.}
  \hspace{-6mm}
    \begin{tabular}
 {c@{\hspace{0.3em}}|@{\hspace{0.3em}}c@{\hspace{0.7em}}c@{\hspace{0.7em}}c@{\hspace{0.3em}}|@{\hspace{0.3em}}c@{\hspace{0.7em}}c@{\hspace{0.7em}}c@{\hspace{0.3em}}|@{\hspace{0.3em}}c@{\hspace{0.7em}}c@{\hspace{0.7em}}c@{\hspace{0.3em}}|@{\hspace{0.3em}}c@{\hspace{0.7em}}c@{\hspace{0.7em}}c@{\hspace{0.3em}}|@{\hspace{0.3em}}c@{\hspace{0.7em}}c@{\hspace{0.7em}}c@{\hspace{0.3em}}|@{\hspace{0.3em}}c@{\hspace{0.7em}}c@{\hspace{0.7em}}c@{\hspace{0.3em}}|@{\hspace{0.3em}}c@{\hspace{0.7em}}c@{\hspace{0.7em}}c@{\hspace{0.3em}}|@{\hspace{0.3em}}c@{\hspace{0.7em}}c@{\hspace{0.7em}}c@{\hspace{0.3em}}}
    \multirow{3}[3]{*}{{Method}} & \multicolumn{6}{c@{\hspace{0.3em}}|@{\hspace{0.3em}}}{MSO Dataset} & \multicolumn{6}{c@{\hspace{0.3em}}|@{\hspace{0.3em}}}{MSRA Dataset} & \multicolumn{6}{c@{\hspace{0.3em}}|@{\hspace{0.3em}}}{DUT-O Dataset} & \multicolumn{6}{c@{\hspace{0.3em}}}{PASCAL-S Dataset} \\
\cline{2-25}          & \multicolumn{3}{c@{\hspace{0.3em}}|@{\hspace{0.3em}}}{Precision} & \multicolumn{3}{c@{\hspace{0.3em}}|@{\hspace{0.3em}}}{Recall} & \multicolumn{3}{c@{\hspace{0.3em}}|@{\hspace{0.3em}}}{Precision} & \multicolumn{3}{c@{\hspace{0.3em}}|@{\hspace{0.3em}}}{Recall} & \multicolumn{3}{@{\hspace{0.3em}}c|@{\hspace{0.3em}}}{Precision} & \multicolumn{3}{c@{\hspace{0.3em}}|@{\hspace{0.3em}}}{Recall} & \multicolumn{3}{@{\hspace{0.3em}}c|@{\hspace{0.3em}}}{Precision} & \multicolumn{3}{c@{\hspace{0.3em}}}{Recall}\\
& $S$   & $S_s$ & $S_l$ & $S$   & $S_s$ & $S_l$ & $S$   & $S_s$ & $S_l$ & $S$   & $S_s$ & $S_l$ & $S$   & $S_s$ & $S_l$ & $S$   & $S_s$ & $S_l$ & $S$   & $S_s$ & $S_l$ & $S$   & $S_s$ & $S_l$ \\
    \hline
    SalCNN\cite{zhang2015SOD}+NMS   & 63.0 & 23.3 & 76.1 & 74.2  & 42.7 & 81.4 & 65.4 & 11.4 & 78.6 & 81.9 & \textbf{74.0} & 83.4 & 51.7 & 19.6 & 70.0 & 44.5 & 25.9 & 46.7 & 68.8  & 28.1 & 82.2 & 55.7 & 17.4 & 73.8 \\
    SalCNN+MMR   & 61.4 & 23.7 & 72.6  & 74.9 & 38.5 & 84.9 & 70.6 & 15.9 & 78.9 & 82.0 & 71.9 & 84.3 & 57.9 & 24.4 & 69.9 & 44.2 & 24.0 & 48.1 & 74.2 & 35.9 & 81.7 &  55.1 & 16.0 & 74.2 \\
    SalCNN+MAP   & 77.5 & 43.8 & 79.2 & 74.1 & 40.6 & 84.6 & 77.1 & 20.7 & 79.4  & 81.5  & 72.7 & 83.7 & 65.5 & 31.6 & 70.3 & 43.7 & 23.3 & \textbf{47.3} & 76.8 & 28 & 82.1 & 55.8 & 9.7& 77.3 \\
    RSD-ResNet  & \textbf{79.7} & \textbf{69.1} & \textbf{81.8} & \textbf{74.9} & \textbf{49.0} & \textbf{85.5} & \textbf{90.1} & \textbf{39.9} & \textbf{85.2} & \textbf{82.0} & 67.8  & \textbf{87.1} & \textbf{76.0} & \textbf{61.2} & \textbf{80.0} & \textbf{44.6} & \textbf{25.6} & 45.3 & \textbf{78.2} &\textbf{72.7} &\textbf{82.1} & \textbf{56.0} &\textbf{16.7} &\textbf{77.8}\\
    \hline
    \end{tabular}%
  \label{tab:prec_recall}%
\end{table*}%
\begin{table*}[!h]
  \centering
  \footnotesize
  \caption{The run-time speed (in fps) of our RSD and compared methods during inference. Our methods with suffix ``S'' are for single salient object detection, while the ones with suffix ``M'' are for multiple salient object detection.}
    \begin{tabular}{c|ccccccc}
    \hline
    {Method} & RSD-VGG-S & RSD-VGG-M & RSD-ResNet-S & RSD-ResNet-M & SalCNN+MAP   & SalCNN+NMS   & SalCNN+MMR \\
    \hline
    Speed & \textbf{120.48} & 113.63 & 19.05  & 18.65 & 10.64 & 10.72 & 10.71 \\
    \hline
    \end{tabular}%
  \label{tab:timing}%
\end{table*}
{\flushleft \textbf{Object size.}} The behavior of detection methods usually differs when dealing with small and large objects. To better understand how our method works compared to existing methods, we further analyze its performance with respect to different sizes of objects. Objects with an area larger than $200\times200$ pixels are counted as \emph{large} objects. For MSO and DUT-O datasets, the ground-truth boxes with an area less than $75\times75$ pixels are defined as \emph{small} objects. We increase this size to $125\times125$ pixels for the MSRA dataset to obtain a statistically reliable subset for performance estimation since the salient objects in this dataset are generally larger. 
We evaluate the precision and recall on small and large objects separately and show the results in Table~\ref{tab:prec_recall}.

Our RSD-ResNet clearly outperforms all the compared methods, achieving the best performance on the MSO dataset for both small and large objects. It also produces the best recall at the same precision for large objects on MSRA dataset and small objects on DUT-O dataset, indicating that it discovers objects of different sizes well under various conditions. At the same recall, our RSD-ResNet greatly improves the precision, especially for small objects that are difficult to locate by object proposal based approaches.
%

\subsection{Run-time efficiency}
By directly generating the saliency map through network forward without proposals, our approach is extremely efficient for salient object detection during inference. We compare the run-time speed of SalCNN~\cite{zhang2015SOD} and our approach in Table~\ref{tab:timing}. With ResNet, our approach achieves nearly 20 fps, while SalCNN only runs at 10 fps. With VGG16, our method achieves an impressive speed at 120 fps, $12\times$ faster than SalCNN, and readily applicable to real-time scenarios.
This experiment confirms that we successfully improve the run-time speed of the network by removing the bottleneck of proposal generation and refinement.

\section{Conclusion}

We have presented a real-time unconstrained salient object detection framework using deep convolutional neural networks, named RSD. 
By eliminating the steps of proposing and refining thousands of candidate boxes, our network learns to directly generates the exact number of salient objects. 
Our network performs saliency map prediction without pixel-level annotations, salient object detection without object proposals, and salient object subitizing simultaneously, all in a single pass within a unified framework.
Extensive experiments show that our RSD approach outperforms existing methods on various datasets for salient object detection and subitizing, and produces comparable results for salient foreground segmentation. In particular, our approach based on VGG16 network achieves more than 100 fps on average on GPU during inference time, which is $12\times$ faster than the state-of-the-art approach, while being more accurate.

{\small
\bibliographystyle{ieee}
\bibliography{egbib}
}

\end{document}